\documentclass{article}
\usepackage[utf8]{inputenc}
\usepackage[margin = 1in]{geometry}
\usepackage{authblk}

\usepackage{amsmath, amssymb, amsthm, amsfonts, mathtools, multirow, mathrsfs, tabu, booktabs, lipsum, url, subcaption, graphicx,xcolor, tabu,  authblk, hyperref, blindtext}

\usepackage[english]{babel}
\usepackage[shortlabels]{enumitem}
 \usepackage[ruled,vlined, linesnumbered]{algorithm2e}
\usepackage[capbesideposition=outside,capbesidesep=quad]{floatrow}
\usepackage[framemethod=tikz]{mdframed}
\usepackage{bbm}
\usepackage{tabularx}

\usepackage[sort&compress,numbers]{natbib}
 
\allowdisplaybreaks

\newcommand{\appropto}{\mathrel{\vcenter{
  \offinterlineskip\halign{\hfil$##$\cr
    \propto\cr\noalign{\kern2pt}\sim\cr\noalign{\kern-2pt}}}}}
    
\newcommand{\R}{\mathbb{R}} 

\DeclareMathOperator*{\argmin}{argmin} 
\DeclareMathOperator*{\argmax}{argmax} 

\newcommand{\ths}{\textsuperscript{th} } 

\usepackage[utf8]{inputenc}
\usepackage[autostyle=true, threshold=2]{csquotes}

\title{Unsupervised Diffusion and Volume Maximization-Based Clustering of Hyperspectral Images}
\author[1]{Sam L. Polk}
\author[2]{Kangning Cui}
\author[3,4]{Aland H. Y. Chan}
\author[3,4]{David A. Coomes}
\author[5]{\\ Robert J. Plemmons}
\author[1]{James M. Murphy\footnote{Corresponding Author: JM.Murphy@Tufts.edu}}
\affil[1]{ Department of Mathematics, Tufts University}
\affil[2]{ Department of Mathematics, City University of Hong Kong}
\affil[3]{ Conservation Research Institute, University of Cambridge}
\affil[4]{ Department of Plant Sciences, University of Cambridge}
\affil[5]{ Departments of Mathematics and Computer Science, Wake Forest University}
\date{}                     

\begin{document}

\maketitle
\begin{abstract}
Hyperspectral images taken from aircraft or satellites contain information from hundreds of spectral bands, within which lie latent lower-dimensional structures that can be exploited for classifying vegetation and other materials. A disadvantage of working with hyperspectral images is that, due to an inherent trade-off between spectral and spatial resolution,  they have a relatively coarse spatial scale, meaning that single pixels may correspond to spatial regions containing multiple materials. This article introduces the \emph{Diffusion and Volume maximization-based Image Clustering} (\emph{D-VIC}) algorithm for unsupervised material clustering to address this problem.  By directly incorporating pixel purity into its labeling procedure, D-VIC gives greater weight to pixels that correspond to a spatial region containing just a single material. D-VIC is shown to outperform comparable state-of-the-art methods in extensive experiments on a range of hyperspectral images, including land-use maps and highly mixed forest health surveys (in the context of ash dieback disease), implying that it is well-equipped for unsupervised material clustering of spectrally-mixed hyperspectral datasets.
\end{abstract}

\noindent \textbf{Index Terms}:  Hyperspectral Imaging, Clustering, Diffusion Geometry, Spectral Unmixing, Forest Health, Ash Dieback.

\section{Introduction}  \label{sec: intro}

Hyperspectral images (HSIs) are images of a scene or object that store spectral reflectance at a hundred or more spectral bands per pixel~\cite{eismann2012hyperspectral,ghamisi2017advanced, plaza2011recent}. HSI remote sensing data, which is generated continuously by airborne and space-borne sensors, has been used successfully for signal processing problems in fields including forensic medicine (e.g., age estimation of forensic traces~\cite{edelman2012hyperspectral}), conservation (e.g., species mapping in wetlands~\cite{adam2010multispectral, hirano2003hyperspectral}), and ecology (e.g., estimating water content in vegetation canopies~\cite{clevers2010estimating}). The high-dimensional characterization of a scene provided in remote sensing HSI data has motivated its use in \emph{material classification} problems~\cite{dalponte2008fusion}, wherein machine learning is used to separate pixels based on the constituent materials (including vegetation types, trees species, and plant health) within spatial regions ~\cite{ghamisi2017advanced, plaza2011recent, wang2022using}.

Though hyperspectral imagery has become an essential tool across many scientific domains, material classification using HSI data faces at least two key challenges. First, because of an inherent trade-off between spectral and spatial resolution, HSIs are generated at a coarse spatial resolution~\cite{jia2020status, price1997spectral, bioucas2013hyperspectral,  laparrcr2015spatial, miao2007endmember, pacheco2022challenging}. One would prefer an HSI with both a high spatial resolution (so that individual pixels correspond to spatial regions containing just one material) and a high spectral resolution (to enable capacity for material classification)~\cite{price1997spectral}. However, an increase in the spatial resolution of an HSI often comes at the cost of reducing the effective detection energy entering the recording spectrometer across each spectral band~\cite{jia2020status}. While this effect may at least partially be mitigated by increasing the aperture of the optical system underlying the spectrometer used to generate HSI data~\cite{jia2017high}, high-aperture instruments generally also have high volume and weight~\cite{jia2020status}. As such, HSI data is typically generated at a coarse spatial resolution (roughly 1~m from drone, 3-10~m from aircraft,  30~m from space). Thus, though some \emph{high-purity} pixels in an HSI may correspond to spatial regions containing predominantly just one material, other pixels are mixed: corresponding to spatial regions with many distinct materials~\cite{bioucas2013hyperspectral, miao2007endmember}.  A second challenge is that the generation of expert labels---often used for training supervised machine learning models---is generally impractical due to the large quantities of HSI data continuously produced by remote sensors~\cite{eismann2012hyperspectral}. To efficiently analyze unlabeled HSIs, one may use HSI clustering algorithms, which partition HSI pixels into groups of points sharing key commonalities~\cite{friedman2001elements}. These algorithms are \emph{unsupervised}; i.e., ground truth labels are not used to provide a partition of an HSI~\cite{friedman2001elements}.  Though clustering has become an important tool in the field of hyperspectral imagery \cite{murphy2018unsupervised, abdolali2021beyond, zhuang2016locality, kuang2012symmetric, wang2019scalable, camps2007semi, gao2014hyperspectral, wu2017semi, yang2018hyperspectral, nalepa2020unsupervised, gillis2014hierarchical, li2021self, sun2020deep, zhou2016novel}, HSI clustering algorithms that do not directly account for the fact that HSI pixels are often spectrally mixed may fail to extract meaningful latent cluster structure \cite{cui2021unsupervised, cui2022unsupervised}.

This article introduces the \emph{Diffusion and Volume maximization-based Image Clustering} (\emph{D-VIC}) algorithm for unsupervised material classification (i.e., \emph{material clustering}) of HSIs.  D-VIC is the first algorithm to simultaneously exploit the high-dimensional geometry~\cite{bachmann2005exploiting, abdolali2021beyond} and abundance structure~\cite{cui2021unsupervised, bioucas2013hyperspectral} observed in HSIs for the clustering problem. In its first stage, D-VIC locates cluster modes:  high-purity, high-empirical density pixels that are far in diffusion distance (a data-dependent distance metric~\cite{coifman2006diffusion}) from other high-purity, high-density pixels. These pixels serve as exemplars for all underlying material structure in the HSI. In its mode selection, D-VIC downweights high-density pixels that correspond to commonly co-occurring groups of materials. As such, D-VIC's exploitation of spectrally mixed structure in HSI data~\cite{jia2020status, price1997spectral, bioucas2013hyperspectral,  laparrcr2015spatial} enables the selection of modes that better represent material structure in the scene.  After detecting cluster modes, D-VIC propagates modal labels to non-modal pixels in order of decreasing density and pixel purity. Since pixel purity is also incorporated into D-VIC's non-modal labeling, D-VIC accounts for material abundance structure in the HSI during its entire labeling procedure. D-VIC is compared against classical and related state-of-the-art HSI clustering algorithms on three benchmark real HSI datasets and applied to the problem of unsupervised detection of a forest pathogen---ash dieback disease (\textit{Hymenoscyphus fraxineus})~\cite{baral2014hymenoscyphus, mckinney2014ash,stone2017application,  waser2014evaluating}---using real remote sensing HSI data. On each dataset,  D-VIC produces competitive unsupervised labelings and, moreover enjoys robustness to hyperparameter selection. Computationally, D-VIC scales quasilinearly in the size of the HSI, and its empirical runtime is competitive, suggesting it is well-suited to cluster large HSIs.

The rest of this article is structured as follows. Section \ref{sec: background} provides background on HSI clustering, diffusion geometry, and spectral unmixing. Section \ref{sec: D-VIC} motivates incorporating spectral unmixing into a nonlinear graph-based clustering framework and introduces D-VIC. Section \ref{sec: numerical experiments} demonstrates the efficacy of D-VIC through substantial experiments on three real HSI datasets.  Additionally, it is shown in Section \ref{sec: numerical experiments} that D-VIC may be used for an unsupervised ash dieback disease detection problem using remotely sensed HSI data collected over a forest in Great Britain~\cite{chan2021monitoring}.  We conclude and offer directions for future work in Section \ref{sec: conclusion}. Finally, in Appendix \ref{app: grid search}, we detail hyperparameter optimization.

\section{Background} \label{sec: background}

\subsection{Background on Unsupervised HSI Clustering} \label{sec: clustering}
\noindent
HSI clustering algorithms partition an HSI, denoted $X=~\{x_i\}_{i=1}^n\subset~\R^D$ (interpreted as a point cloud of HSI pixels' spectral signatures, with $n$   pixels and $D$  spectral bands) into $K$ \emph{clusters} of pixels. The partition, which we call a \emph{clustering} of $X$, may be encoded in a labeling vector $\mathcal{C}\in\{1,2,\dots, K\}^n$ such that $\mathcal{C}(x_i) = \mathcal{C}_i\in\{1,2,\dots,K\}$ is the label assigned to the pixel $x_i$. Ideally, pixels from any one cluster are in some sense ``related,'' and pixels from any two clusters are ``unrelated'' ~\cite{friedman2001elements, ng2002spectral,   maggioni2019learning, gillis2014hierarchical}. Clustering algorithms are \emph{unsupervised}, meaning that data points are labeled without the aid of any expert annotations or ground truth labels. This has motivated the development of algorithms explicitly built for material clustering using HSIs~\cite{murphy2018unsupervised, abdolali2021beyond, zhuang2016locality, kuang2012symmetric, wang2019scalable, camps2007semi, gao2014hyperspectral, wu2017semi, yang2018hyperspectral, nalepa2020unsupervised, gillis2014hierarchical, li2021self, sun2020deep, cahill2014schroedinger, zhou2016novel}. 

\begin{figure}[t]
    \centering
    \includegraphics[width = 0.32\textwidth]{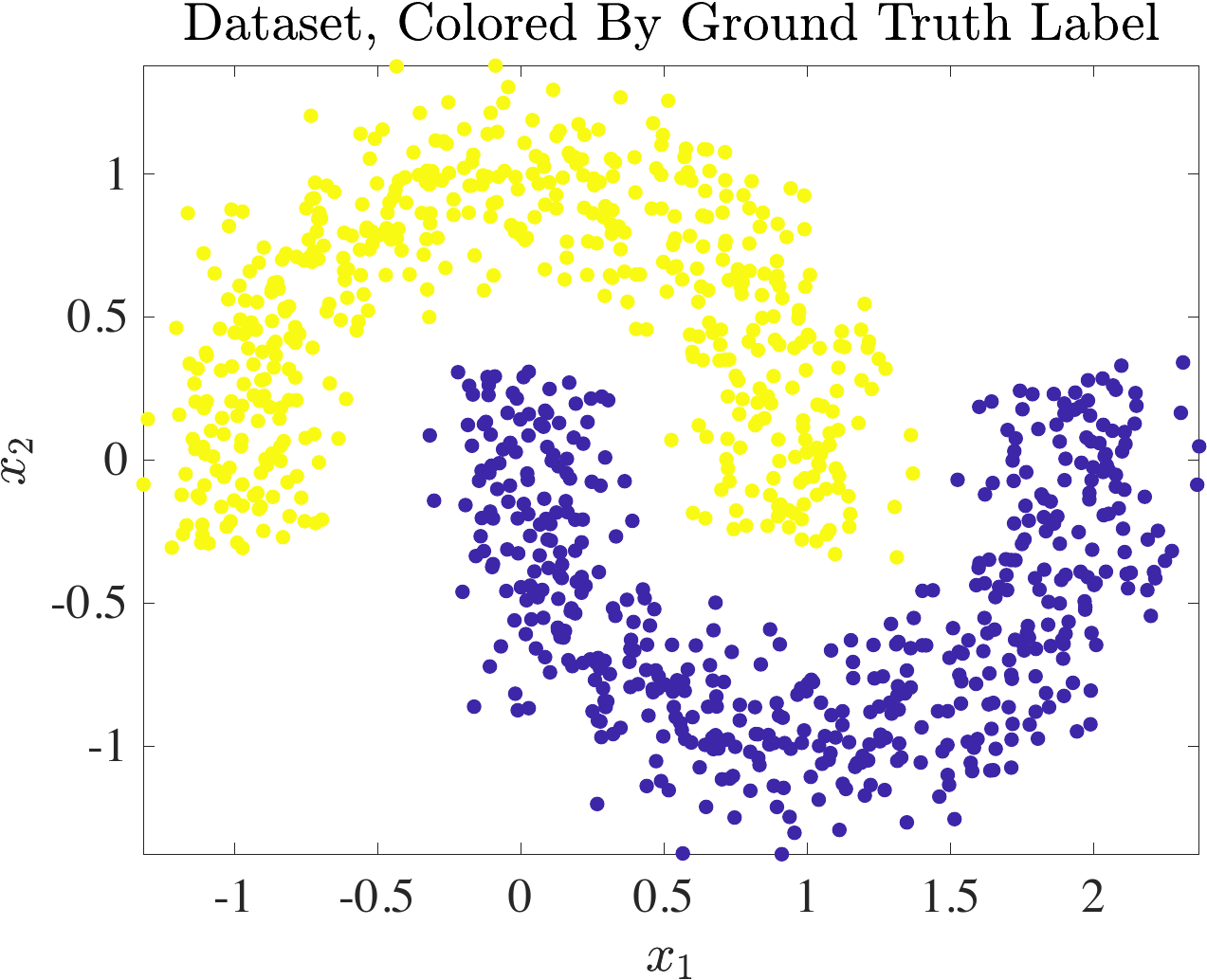}
    \includegraphics[width = 0.32\textwidth]{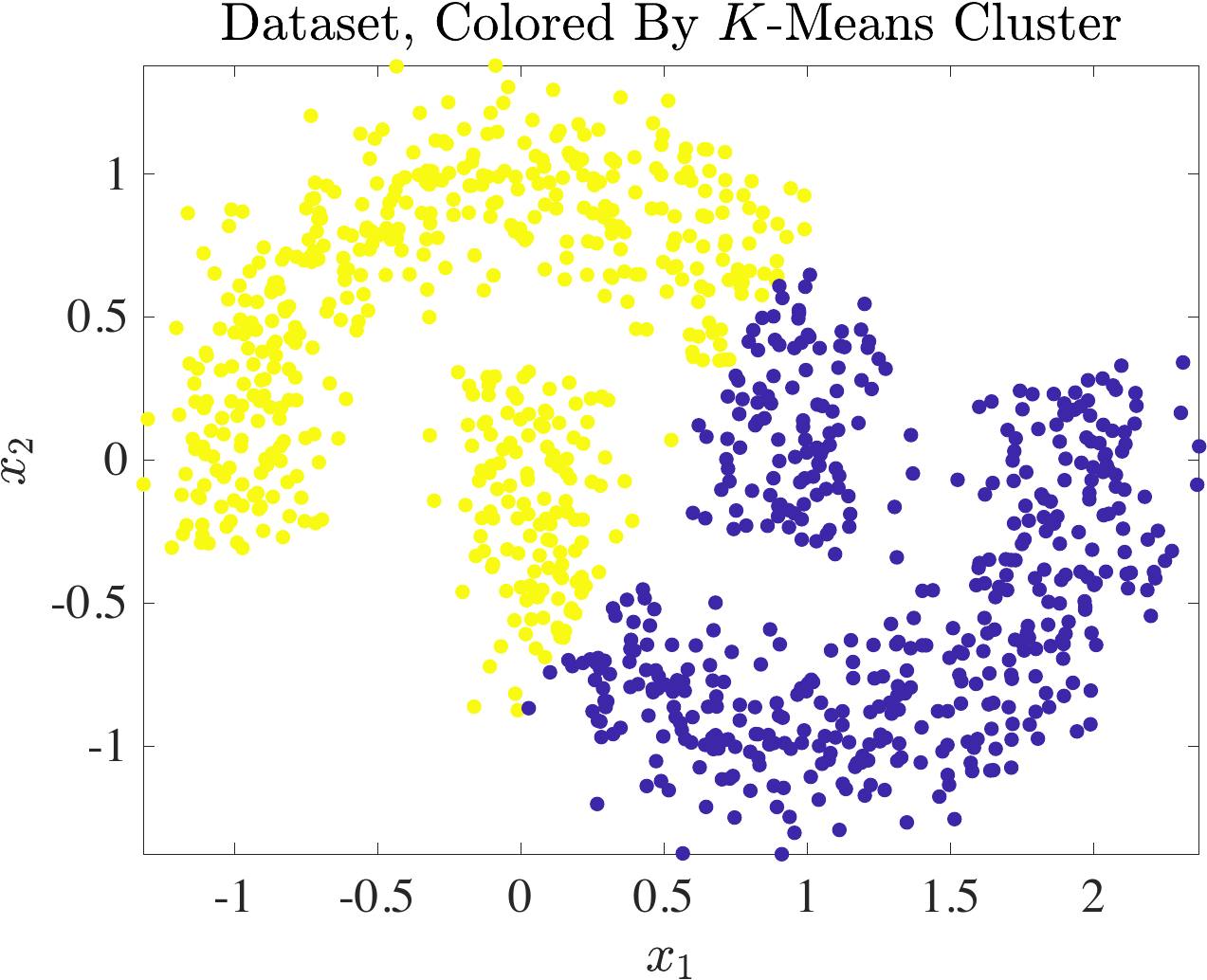}
    \includegraphics[width = 0.32\textwidth]{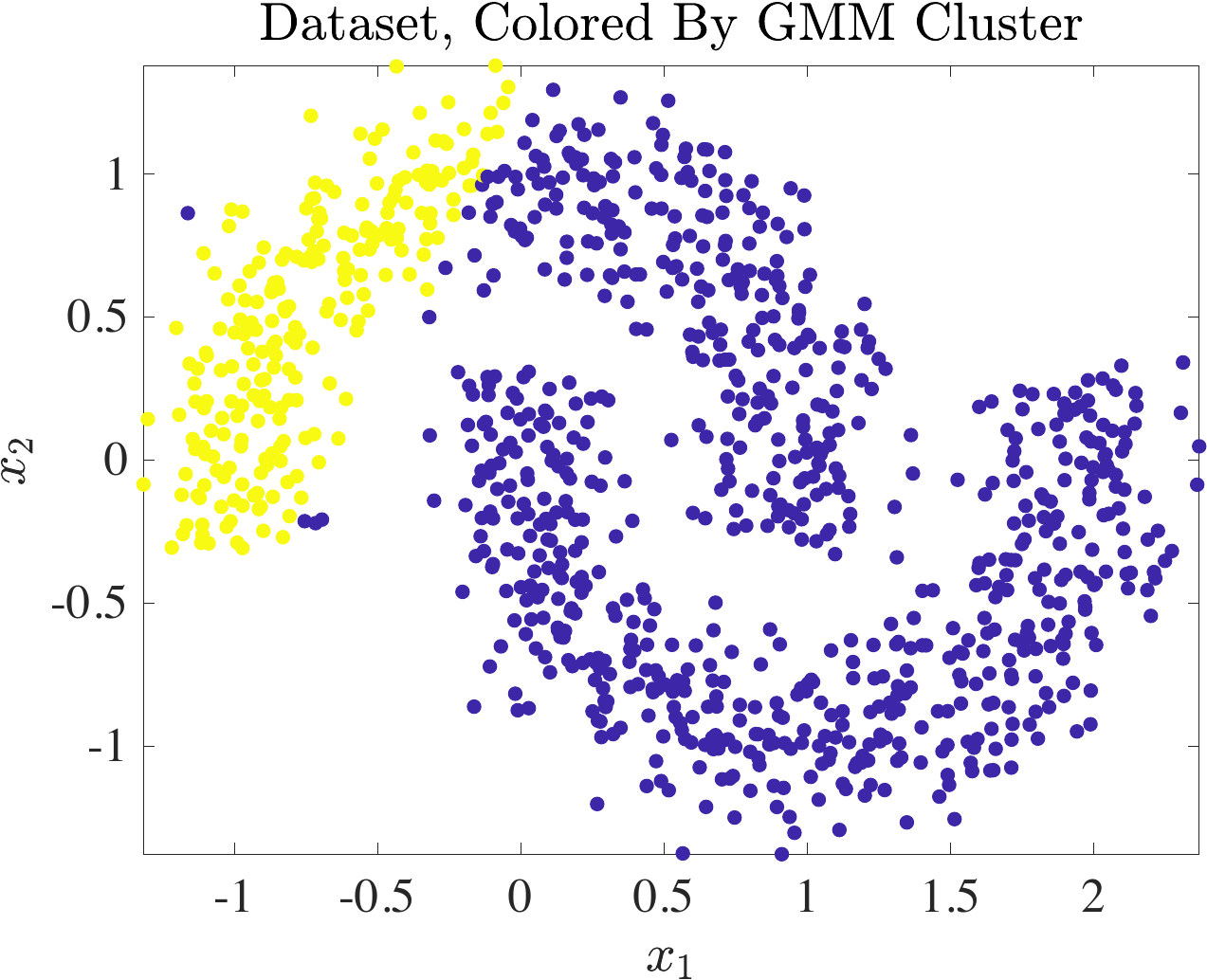}
    
    \hfill (a) \hspace{1.86in} (b) \hspace{1.86in} (c) \hspace{0.88in}

    \caption{Example of a toy dataset $(n=1000)$ with nonlinear cluster structure: two interleaving half-circles. The idealized clustering, visualized in (a), separates each half-circle. Due to the lack of a linear decision boundary, however, both $K$-Means (b) and GMM (c) were unable to extract latent cluster structure from this simple nonlinear dataset. Both algorithms were run with 100 replicates.}
    \label{fig:nonlinear}
\end{figure}

Though classical clustering algorithms such as $K$-Means and the Gaussian Mixture Model (GMM)~\cite{friedman2001elements} remain widely used in practice, these algorithms tend to perform poorly on HSIs for a number of reasons~\cite{murphy2018unsupervised, abdolali2021beyond}. First, algorithms that rely on Euclidean distances are prone to the ``curse of dimensionality'' on datasets like HSIs that have a high ambient dimension (i.e., the number of spectral bands is large)~\cite{theodoridis2006pattern}. Second, HSIs are often spectrally mixed~\cite{jia2020status, price1997spectral, bioucas2013hyperspectral,  laparrcr2015spatial}, and overlap may exist between clusters in Euclidean space~\cite{murphy2018unsupervised}. A final complication is that classical algorithms generally assume that latent clusters in a dataset are approximately ellipsoidal groups of points that are well-separated in Euclidean space~\cite{friedman2001elements}, but clusters in HSIs often exhibit nonlinear structure~\cite{bachmann2005exploiting}. A simple toy dataset, visualized in Fig. \ref{fig:nonlinear} \cite{ng2002spectral}, serves as an example of a dataset with nonlinear structure. This dataset lacks a linear decision boundary between its $K=2$ latent clusters, and classical algorithms ($K$-Means and GMM~\cite{friedman2001elements}) were unable to learn its latent nonlinear cluster structure. HSIs often contain clusters that can only be separated using a nonlinear decision boundary~\cite{murphy2018unsupervised}; thus, algorithms that rely solely on Euclidean distances are expected to perform poorly at material clustering on HSIs.

The limitations outlined above have motivated the application and development of nonlinear graph-based algorithms for HSI clustering~\cite{abdolali2021beyond, zhuang2016locality, kuang2012symmetric, zhu2017unsupervised, murphy2018unsupervised, wang2019scalable, camps2007semi, gao2014hyperspectral, ng2002spectral, cahill2014schroedinger, wang2021fast, zhou2016novel, bandyopadhyay2022tree}. Graph-based algorithms rely on data-generated graphs; pixels are represented as nodes in the graph, and edges encode pairwise similarity between them. Highly connected regions in the graph may then be summarized using a nonlinear coordinate transformation~\cite{tenenbaum2000global,roweis2000nonlinear, belkin2001laplacian, coifman2006diffusion}, as is described in more detail in Section \ref{sec: spectral graph theory}. Thus, a partition may be obtained by implementing a classical clustering algorithm on the dimension-reduced dataset.  Due to their reliance on a graph representation of an HSI, these algorithms tend to be robust to small perturbations in the data and noise. Moreover, theoretical guarantees exist for the successful recovery of latent cluster structure, even if boundaries between latent clusters are nonlinear~\cite{rohe2011spectral, murphy2021multiscale, maggioni2019learning}. Despite their exhibited successes, algorithms that rely solely on graph structure tend to perform poorly on datasets containing multimodal cluster structure~\cite{maggioni2019learning, murphy2021multiscale, nadler2007fundamental}; i.e., if a single cluster has multiple regions of high and low density. Importantly, this includes spectrally mixed HSIs, the classes of which often contain multiple co-occurring materials of varying abundances~\cite{jia2020status, price1997spectral, bioucas2013hyperspectral}. 

Deep neural networks and graph convolutional networks have recently become popular for material classification and clustering in HSIs because of their capacity for prediction on complex data sources ~\cite{wu2017semi, yang2018hyperspectral, nalepa2020unsupervised, li2021self, sun2020deep, dilokthanakul2016deep, min2018survey, bandyopadhyay2022tree}. While these algorithms tend to be highly accurate at material classification using real HSI data, many state-of-the-art deep models for HSI segmentation still rely in some part on training labels, whether via pre-training some or all of the network~\cite{dilokthanakul2016deep, min2018survey} or explicitly relying upon a small number of ground truth labels~\cite{tasissa2021deep, bandyopadhyay2022tree, wu2017semi, yang2018hyperspectral}, and/or pseudo-labels~\cite{li2021self, wu2017semi}. Moreover, even fully unsupervised ``deep clustering'' algorithms~\cite{nalepa2020unsupervised, sun2020deep} rely on deep neural networks, which have been shown to be prone to error from perturbations and noise~\cite{nguyen2015deep, szegedy2013intriguing} and whose success in unsupervised clustering is often due to data pre-processing steps rather than the unsupervised network learning meaningful features \cite{haeffele2020critique}.

\subsection{Background on Spectral Graph Theory} \label{sec: spectral graph theory}
\noindent
As overviewed in Section \ref{sec: clustering}, graph-based clustering algorithms learn latent, possibly nonlinear cluster structure from HSIs by treating pixels as nodes in an undirected, weighted graph, where connections between pixels are encoded in a weight matrix $\mathbf{W}\in\R^{n\times n}$~\cite{ ng2002spectral, maggioni2019learning, murphy2021multiscale, polk2021multiscale}. In large datasets like HSIs,  edges can be restricted to the first $N\ll n$ $\ell^2$-nearest neighbors (i.e., Euclidean distance nearest neighbors) and given unit weight.   In other words, $\mathbf{W}_{ij}=1$ if $x_{i}$ is one of the $N$ nearest neighbors of $x_{j}$ or vice versa, and $\mathbf{W}_{ij}=0$ otherwise.  Let  $\mathbf{P}=\mathbf{D}^{-1}\mathbf{W}$, where $\mathbf{D}\in\R^{n\times n}$ is the diagonal degree matrix defined by $\mathbf{D}_{ii} = \sum_{j=1}^n \mathbf{W}_{ij}$.  The matrix $\mathbf{P}\in\R^{n\times n}$ may be interpreted as the transition matrix for a Markov diffusion process on $X$ and has a unique stationary distribution $\pi\in\R^{1\times n}$ satisfying $\pi \mathbf{P} = \pi$~\cite{coifman2006diffusion, maggioni2019learning}.  Define $\{(\lambda_i, \psi_i)\}_{i=1}^n$ to be the (right) eigenvalue-eigenvector pairs of $\mathbf{P}$, sorted in non-increasing order so that $1 = \lambda_1 > |\lambda_2| > \dots > |\lambda_n|\geq 0$. The first $K$ eigenvectors of $\mathbf{P}$ often concentrate on the $K$ most coherent subgraphs in the graph underlying $\mathbf{P}$, making these vectors useful for clustering~\cite{ng2002spectral}.

\subsubsection{Background on Diffusion Geometry} \label{sec: diffusion}

Diffusion distances are a family of data-dependent distance metrics which enable comparisons between points in the context of the Markov diffusion process encoded in $\mathbf{P}$~\cite{  coifman2006diffusion}. Diffusion distances have been successfully used in a number of applications (e.g., in gene expression profiling~\cite{haghverdi2016diffusion, van2018recovering}, data visualization~\cite{zhao2014rotationally, moon2019visualizing}, and molecular dynamics analysis~\cite{rohrdanz2011determination, zheng2011polymer, chen2018molecular}). Moreover, diffusion distances have been shown to efficiently capture low-dimensional structure in HSI data, resulting in excellent clustering performance~\cite{murphy2018unsupervised, murphy2021multiscale}.  

Define $D_t(x_i,x_j) = \sqrt{\sum_{k=1}^n [(\mathbf{P}^t)_{ik}-(\mathbf{P}^t)_{jk}]^2/\pi_k}$ to be  the \emph{diffusion distance} at time $t\geq 0$ between pixels $x_i,x_j\in X$~\cite{coifman2005geometric, coifman2006diffusion, nadler2006diffusion}. Diffusion distances are a nonlinear data-dependent distance metric that have a natural connection to the clustering problem~\cite{maggioni2019learning, murphy2021multiscale}. To see this, note that $D_t(x_i,x_j)$ may be interpreted as the Euclidean distance between the $i$\ths and $j$\ths rows of $\mathbf{P}^t$, weighted according to $1/\pi$. If pixels from the same cluster share many high-weight paths of length $t$, but paths of length $t$ between any two pixels from different clusters are relatively low weight, then the $i$\ths and $j$\ths rows of $\mathbf{P}^t$ are expected to be nearly equal for pixels $x_i$ and $x_j$ from the same cluster and very different if these pixels come from different clusters. So, the diffusion distance between points from the same cluster is expected to be small, and the diffusion distance between points from different clusters is expected to be large~\cite{maggioni2019learning, murphy2021multiscale}.  Diffusion distances can be efficiently computed using the eigendecomposition of $\mathbf{P}$: $D_t(x_i,x_j) = \sqrt{\sum_{k=1}^n \lambda_k^{2t}[(\psi_i)_k - (\psi_j)_k ]^2}$~\cite{coifman2006diffusion, nadler2006diffusion, coifman2005geometric}. For $t$ sufficiently large so that $|\lambda_k|^{2t}\approx 0$ for $k>\ell$, the sum in diffusion distances can be truncated past the $\ell$\ths term, yielding an accurate and efficient approximation of diffusion distances. Importantly, the relationship between diffusion distances and the eigendecomposition of $\mathbf{P}$ indicates that diffusion distances may be interpreted as Euclidean distances after nonlinear dimensionality reduction via the following dimension-reduced mapping of the ambient space into $\R^\ell$: $x_i\rightarrow [\lambda_1^t(\psi_1)_i\; \lambda_2^t(\psi_2)_i\; \dots, \lambda_\ell^t(\psi_\ell)_i]$~\cite{coifman2006diffusion, coifman2005geometric, nadler2006diffusion}.

HSIs often encode well-defined latent multiscale cluster structure that can be learned by diffusion-based HSI clustering algorithms by varying the time parameter $t$ in diffusion distances~\cite{polk2021multiscale, murphy2021multiscale}. Indeed, smaller $t$ values generally enable the detection of fine-scale local cluster structure, while larger $t$ values enable the detection of coarse-scale global cluster structure. However, for algorithms that require $K$ as an input, $t$ must be tuned to correspond to the desired number of clusters. Thus, the choice of $t$ must be carefully considered when clustering a dataset using an algorithm that relies on diffusion distances~\cite{murphy2021multiscale, polk2021multiscale}.

\subsection{Background on Spectral Unmixing}
\noindent
Real-world HSI data is often generated at a coarse spatial resolution; thus, pixels may correspond to spatial regions containing  multiple materials~\cite{chan2011simplex, winter1999nfindr, manolakis2001hyperspectral}. To learn latent material structure from HSIs, \emph{spectral unmixing} algorithms decompose each pixel's spectrum into a linear combination of \emph{endmembers} that encode the spectral signature of materials in the scene.  The endmembers may be understood as ``pure'' material signatures.  When representing a pixel as a linear combination of these endmembers, the coefficients of the linear combination indicate the relative \emph{abundance} of materials within the spatial region corresponding to that pixel. Mathematically, a spectral unmixing algorithm learns $\mathbf{U}=(u_1\;u_2\;\dots\; u_m)^\top\in \R^{m\times D}$ (with rows encoding the spectral signatures of endmembers) and $\mathbf{A}\in \mathbb{R}^{n\times m}$ (with rows encoding abundances) such that $x_i \approx \sum_{j=1}^m \mathbf{A}_{ij}u_j $ for each $x_i\in X$~\cite{manolakis2001hyperspectral}. Usually, the entries of $\mathbf{A}$ are nonnegative and normalized so that $\sum_{j=1}^m\mathbf{A}_{ij}=1$ for each $i$; hence,  abundances are data-dependent features storing estimates for the relative frequency of materials in pixels. The \emph{purity} of $x_i\in X$, defined by $\eta(x_i) = \max_{1\leq j \leq m} \mathbf{A}_{ij}$~\cite{cui2021unsupervised}, will be large if the spatial region corresponding to $x_i$ is highly homogeneous (i.e., containing predominantly just one material) and small otherwise. As such, pixel purity and spectral unmixing may be used to aid in the unsupervised clustering of HSIs~\cite{cui2021unsupervised, gillis2014hierarchical, kuang2012symmetric}.

Spectral unmixing has become an important tool in hyperspectral imagery, prompting its usage in a number of applications (e.g., image reconstruction~\cite{zhao2013deblurring, berisha2015deblurring, wang2018compressed}, noise reduction~\cite{cerra2013noise, rasti2018noise, rasti2020hyperspectral}, spatial resolution enhancement~\cite{erturk2014spatial, bendoumi2014hyperspectral, kordi2017spatial}, supervised  material classification \cite{villa2010spectral, dopido2012quantitative, cui2021unsupervised}, change detection \cite{erturk2015informative, liu2016unsupervised, camalan2022change, li2022integrated}, and anomaly detection~\cite{qu2018hyperspectral,ma2018hyperspectral, zhou2016novel}). The importance of spectral unmixing in remote sensing has motivated the development of many algorithms for this task, which we broadly summarize here; see surveys ~\cite{somers2011endmember, quintano2012spectral, bioucas2012hyperspectral, bioucas2013hyperspectral, heylen2014review, borsoi2021spectral} for a more thorough overview. Geometric methods for spectral unmixing estimate endmembers by searching for points that form a simplex of minimal volume, subject to a constraint that at least some nearly pure pixels exist within the observed HSI pixels~\cite{chang2006new,neville1999automatic, boardman1995mapping, boardman1993automating, chan2009convex, abdolali2021beyond, bioucas2013hyperspectral, chan2011simplex, nascimento2005VCA, miao2007endmember, clasen2015spectral, heylen2011fully, hendrix2011new, iordache2011sparse}. For highly mixed HSIs that lack pure pixels, statistical methods may be used~\cite{berman2004ice, zare2007sparsity, dobigeon2009joint, moussaoui2006separation, themelis2011novel}. These methods typically treat spectral unmixing as a blind source separation problem, and though they are often successful at this task, statistical algorithms are usually more computationally expensive~\cite{bioucas2013hyperspectral}. 
Additionally, autoencoding methods learn latent spectral mixing structure by training neural networks that map pixel spectra to a lower-dimensional space that can be related to endmember and abundance matrices $\mathbf{U}$ and $\mathbf{A}$~\cite{palsson2020convolutional, su2019daen, palsson2018hyperspectral, qu2018udas, ozkan2018endnet, zhang2018hyperspectral, su2018stacked, khajehrayeni2020hyperspectral, feng2018hyperspectral}. 
Finally, while linear spectral unmixing is well-developed and widely used in practice, some nonlinear unmixing algorithms (including some relying on neural networks~\cite{guilfoyle2001quantitative, licciardi2011pixel, charles2011learning, su2019daen, zhang2018hyperspectral, wang2019nonlinear}) have been developed to account for nonlinear interactions between endmembers~\cite{guilfoyle2001quantitative, licciardi2011pixel, su2019daen, zhang2018hyperspectral, yokoya2013nonlinear, halimi2011nonlinear, chen2012nonlinear, heylen2015multilinear, heylen2010non, wang2019nonlinear}. Nevertheless, many of these algorithms typically require training data or hyperparameter inputs unlike many of the linear mixing models reviewed above~\cite{bioucas2013hyperspectral}. 

Spectral unmixing is relevant to our paper as a way to determine cluster modes in an unsupervised setting.  Below, we focus on two standard methods in unmixing but note that D-VIC is modular in this regard and other unmixing algorithms could be used.  

\subsubsection{Background on the HySime Algorithm}
 
Hyperspectral Signal Subspace Identification by Minimum Error (\emph{HySime}) 
is a standard algorithm for estimating the number of materials $m$ in $X$~\cite{bioucas2008HySime}. HySime assumes that each $x_i\in X$ is of the form  $x_i = y_i +\zeta_i$, where $y_i\in\R^D$ and $\zeta_i\in\R^D$ model the signal and noise associated with $x_i$, respectively. If signal vectors are linear mixtures of $m$ ground truth endmembers (i.e., $y_i = \sum_{j=1}^m \mathbf{A}_{ij}u_j$ for $1\leq i\leq n$), then the set $\{y_i\}_{i=1}^n$ lies on a $m$-dimensional subspace of $\R^D$.  With this motivation, HySime estimates the subspace dimension by balancing the error of projecting signal vectors  $\{y_i\}_{i=1}^n$ onto their first $m$ principal components with the amount of noise captured by those vectors' orthogonal complement. Though other algorithms exist for estimating the number of materials in a scene using HSI data, many of these alternatives rely on hyperparameter inputs to estimate $m$ or have large computational complexity~\cite{chang2018review}. For example, the ubiquitous virtual dimensionality---which relies on a Neyman-Pearson detection theory-based threshold to determine $m$~\cite{chang2004estimation}---has been shown to be highly sensitive to small perturbations in pixel spectra and hyperparameter inputs~\cite{bioucas2008HySime}. In contrast, HySime is hyperparameter-free and can compute a high-quality, numerically stable estimate using only $X$ in just $O(D^2n)$ operations. 

\subsubsection{Background on the AVMAX Algorithm}

Alternating Volume Maximization (\emph{AVMAX}) is a spectral unmixing algorithm, requiring $m$ as a parameter, that searches for vectors $\{u_i\}_{i=1}^m\subset\R^D$ that produce an $m$-simplex of maximal volume, subject to the constraint that each $u_i$ lies in the convex hull of the dataset after Principal Component Analysis (PCA) dimensionality reduction: projecting pixel spectra onto the span of the first $m-1$ principal components~\cite{chan2011simplex}. This dimensionality reduction step is motivated by the fact that any vector in the affine hull of the $m$ endmembers can always be expressed as $y=\mathbf{C}\alpha + \frac{1}{n}\sum_{x\in X} x$, where $\mathbf{C}\in\R^{D\times (m-1)}$ is related to the first $m-1$ principal components of $X$~\cite{chan2011simplex, chan2008convex}; see Lemma 1 in \cite{chan2011simplex} for details. 
Endmembers are optimized through multiple partial maximization procedures (i.e., keeping $m-1$ endmembers constant and optimizing for volume while varying the $m    \ths$ endmember) until convergence~\cite{chan2011simplex}. AVMAX has become popular for spectral unmixing because of its strong performance guarantees and the rigor behind its optimization framework~\cite{chan2011simplex, bioucas2013hyperspectral}. Indeed, in a noiseless, linearly mixed dataset containing the optimal endmember set, if each partial maximization problem in AVMAX converges to a unique solution, AVMAX is guaranteed to converge to the optimal endmember set~\cite{chan2011simplex}. Moreover, AVMAX can easily be modified to make it robust to random initialization; one can run multiple replicates of AVMAX in parallel and choose the endmember set with largest volume. Once endmembers are learned, abundances may be computed using a nonnegative least squares solver: $(\mathbf{A}_{i1}\;\mathbf{A}_{i2}\;\dots\;\mathbf{A}_{im}) = \argmin_{a\in[0,\infty)^m}\|x_i-\sum_{j=1}^m a_j u_j\|_2^2$ for each $x_i\in X$~\cite{bro1997fast}.

\section{Diffusion and Volume maximization-based Image Clustering} \label{sec: D-VIC}
\noindent
In spectrally mixed HSIs, any one pixel may correspond to a spatial region that contains many materials~\cite{bioucas2013hyperspectral, miao2007endmember}. Thus, even state-of-the-art algorithms for unsupervised material clustering may perform poorly on mixed HSIs, failing to recover clusterings that can be linked to materials within the scene. Algorithms that do not directly incorporate a spectral unmixing step into their labeling may assign clusters that correspond to groups of materials rather than clusters corresponding to individual materials. Thus, additional improvements are needed to develop algorithms suitable for material clustering on mixed HSIs.

\begin{algorithm}[t]
\SetAlgoLined
 \KwIn{ $X$ (HSI), $N$ (\# nearest neighbors),  $\sigma_0$ (KDE scale),   $t$ (diffusion time),   $K$ (\# clusters)}
\KwOut{$\mathcal{C}$ (clustering)}
Estimate $m$, the number of latent endmembers in $X$, using HySime~\cite{bioucas2008HySime}\;
Learn endmembers $\mathbf{U}\in\R^{m\times D}$  using AVMAX~\cite{chan2011simplex} and abundances $\mathbf{A}\in \R^{n\times m}$ using a nonnegative least squares solver~\cite{bro1997fast}\; 
For each $x\in X$, calculate pixel purity $\eta(x)$ and empirical density $p(x)$. Store $\zeta(x) = \frac{2\Bar{p}(x)\bar{\eta}(x)}{\bar{p}(x) + \bar{\eta}(x)}$, where $\bar{p}(x) = \frac{p(x)}{\max_{y\in X}p(y)}$ and $\bar{\eta}(x) = \frac{\eta(x)}{\max_{y\in X} \eta(y)}$\; 
Build $d_t(x)$ using $\zeta(x)$, where diffusion distances are computed from a KNN graph with $N$ edges per pixel\;  
Assign  $\mathcal{C}(x_{m_k}) = k$ for $1\leq k \leq K$, where  $\{x_{m_k}\}_{k=1}^K$ are the $K$ pixels maximizing $\mathcal{D}_t(x) = \zeta(x) d_t(x)$\; 
In order of non-increasing $\zeta(x)$, for each unlabeled $x\in X$, assign $x$ the label  $\mathcal{C}(x^*)$, where  $x^* = \argmin_{y\in X}\{D_t(x,y) |\;\zeta(y)\geq \zeta(x) \text{ and } \mathcal{C}(y)>0\}$.
 \caption{Diffusion and Volume maximization-based Image Clustering  \label{alg: D-VIC}}
\end{algorithm}

This section introduces the \emph{Diffusion and Volume maximization-based Image Clustering} (\emph{D-VIC}) algorithm (Algorithm \ref{alg: D-VIC}) for unsupervised material clustering of HSIs. To learn material abundances, D-VIC first performs a spectral unmixing step: decomposing the HSI by  learning the number of endmembers $m$ using HySime~\cite{bioucas2008HySime}, implementing AVMAX with that $m$-value to learn endmembers~\cite{chan2011simplex}, and calculating abundances and purity through a nonnegative least squares solver~\cite{bro1997fast}. 
As will  become clear in Section \ref{sec: numerical experiments}, this estimate for pixel purity resulted in high-quality material clustering with D-VIC. Nevertheless, the choice of algorithm used for spectral unmixing in D-VIC is quite modular and future work may consider applying other endmember extraction algorithms~\cite{chang2004estimation, bioucas2012hyperspectral, bioucas2013hyperspectral, heylen2014review, borsoi2021spectral} and/or abundance solvers that explicitly constrain estimates to sum to one~\cite{chen2013nonlinear, heinz2001fully}. 
D-VIC then estimates empirical density using a \emph{kernel density estimate} (\emph{KDE}) defined by $p(x) =~\frac{1}{Z}\sum_{y\in NN_N(x)} \exp(-\|x-y\|^2_2/\sigma_0^2)$, where $NN_N(x)$ is the set of $N$  $\ell^2$-nearest neighbors of $x$ in $X$, $\sigma_0>0$ is a \emph{KDE scale} controlling the interaction radius between points, and $Z$ is a constant normalizing $p(x)$ so that $\sum_{y\in X} p(y) = 1$. By construction, $p(x)$ will be large if the pixel $x$ is close to its $N$ $\ell^2$-nearest neighbors in $X$ and small otherwise~\cite{rodriguez2014clustering, murphy2018unsupervised, maggioni2019learning}. 

To locate pixels that are both high-density and indicative of an underlying material, D-VIC calculates  $\zeta(x) =~\frac{2\Bar{p}(x)\bar{\eta}(x)}{\bar{p}(x) + \bar{\eta}(x)}$, where $\bar{p}(x) = \frac{p(x)}{\max_{y\in X}p(y)}$ and $\bar{\eta}(x) =~\frac{\eta(x)}{\max_{y\in X} \eta(y)}$. Thus, $\zeta(x)$ returns the harmonic mean of $p(x)$ and $\eta(x)$, which are normalized so that density and purity are approximately at the same scale. By construction, $\zeta(x)\approx 1$ only at high-density, highly pure pixels $x$. In contrast, if a pixel $x$ is either low-density or low-purity, then $\zeta(x)$ will be small. Importantly, $\zeta(x)$ downweights mixed pixels that, though high-density, correspond to a spatial region containing many materials. Thus, points with large $\zeta$-values will correspond to pixels that are modal (due to their high empirical density) and representative of just one material in the scene (due to their high pixel purity).

D-VIC uses the following function to incorporate diffusion geometry into its procedure for selecting cluster modes: 
\begin{align*}
   d_t(x) = \begin{cases} \max_{y\in X} D_t(x,y) & x = \argmax_{y\in X} \zeta(y), \\ 
    \min_{y\in X}\{ D_t(x,y) | \zeta(y)\geq \zeta(x)\} & \text{otherwise}. 
    \end{cases}
\end{align*}
Thus, a pixel will have a large $d_t$-value if it is far in diffusion distance at time $t$ from its $D_t$-nearest neighbor of higher density and pixel purity. D-VIC assigns modal labels to the $K$ points maximizing  $\mathcal{D}_t(x) = d_t(x) \zeta(x)$, which are high-density, high-purity pixels far in diffusion distance at time $t$ from other high-density, high-purity pixels. 

\begin{figure}[t]
    \centering
    \includegraphics[width = 0.83\textwidth]{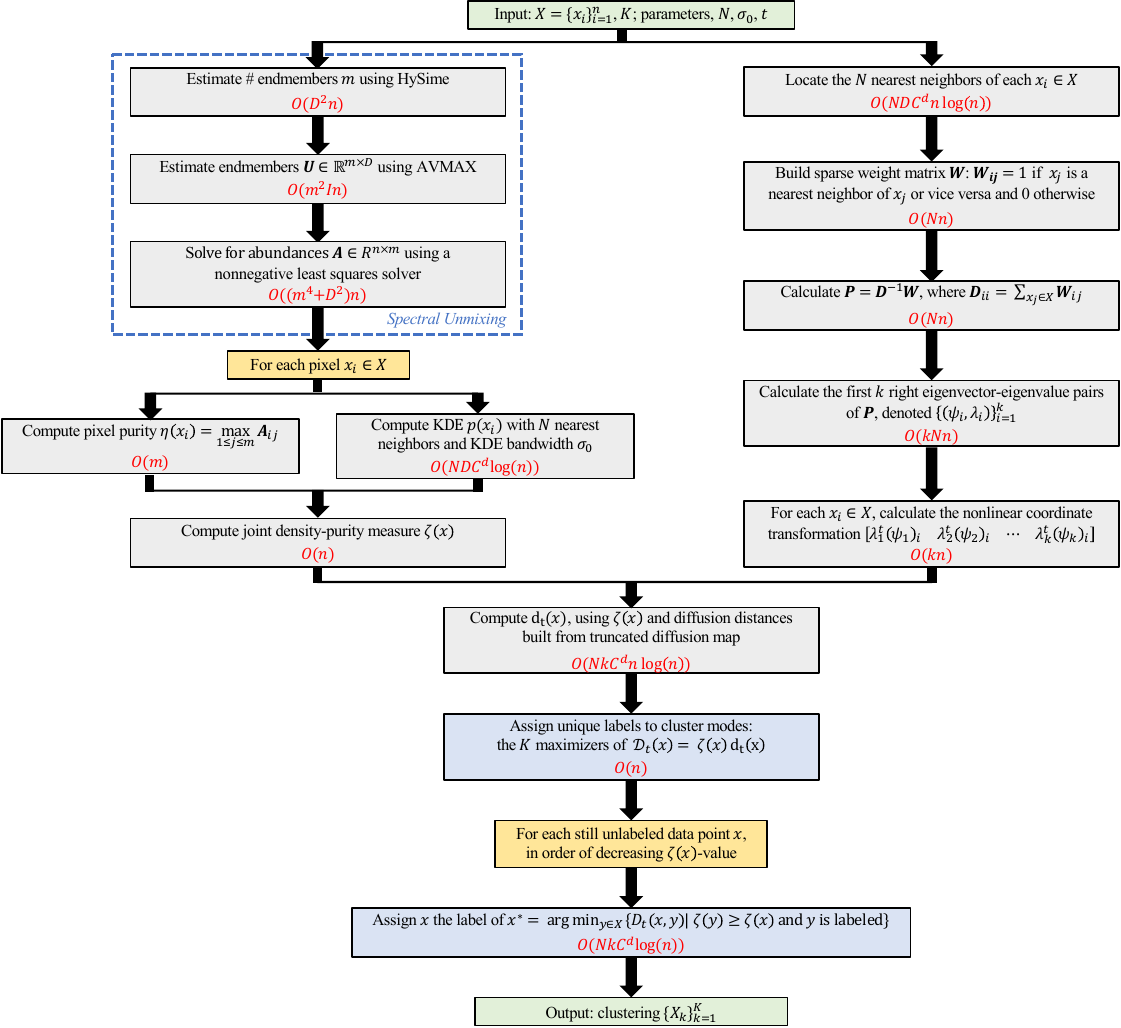}
    \caption{ Diagram of the D-VIC clustering algorithm. The computational complexity of each step is colored in red. The scaling of D-VIC depends on $n$ (no. pixels), $D$ (no. spectral bands), $m$ (no. endmembers), $I$ (no. AVMAX maximizations), $N$ (no. nearest neighbors), $d$ (doubling dimension of $X$~\cite{beygelzimer2006cover}), and $C$: a constant independent of all other parameters~\cite{beygelzimer2006cover}; see Section \ref{sec: CC} for details. Note that all steps are quasilinear with respect to $n$, implying that D-VIC scales well to large HSI datasets. We remark that the spectral unmixing step (indicated with a blue box) is quite modular and other approaches may be used in future work~\cite{bioucas2013hyperspectral, borsoi2021spectral, heylen2014review, somers2011endmember, chen2013nonlinear, heinz2001fully, chang2004estimation}.  
    } 
    \label{fig:schematic}
\end{figure}

After labeling cluster modes,  D-VIC labels non-modal points according to their $D_t$-nearest neighbor of higher $\zeta(x)$-value that is already labeled. Importantly, D-VIC downweights low-purity pixels through $\zeta(x)$ in its non-modal labeling. Thus, pixel purity is incorporated in all stages of the D-VIC algorithm through $\zeta(x)$. D-VIC is provided in Algorithm \ref{alg: D-VIC}, and a schematic is provided in Fig. \ref{fig:schematic}.

\subsection{Computational Complexity} \label{sec: CC}
\noindent

The computational complexity of the HySime algorithm is $O(D^2n)$ operations~\cite{bioucas2008HySime}, whereas the computational complexity of spectral unmixing using AVMAX and a standard nonnegative least squares solver~\cite{bro1997fast} is $O((D^2 + m^4 +m^2I) n )$ operations, where $I$ is the number of AVMAX partial maximizations~\cite{chan2011simplex}. We assume that nearest neighbor searches are performed using  \emph{cover trees}: an indexing data structure that enables logarithmic nearest neighbor searches~\cite{beygelzimer2006cover}. To see this, define the \emph{doubling dimension} of $X$ by $d = \log_2(c)$, where $c>0$ is the smallest value for which any ball $B_p(p,r) = \{q\in~X~|~\|p-~q\|~\leq~r\}$ can be covered by $c$ balls of radius $r/2$. If the spectral signatures of pixels in $X\subset \mathbb{R}^D$ have doubling dimension $d$, a search for the $N$ $\ell^2$-nearest neighbors of each HSI pixel using cover trees has computational complexity $O(NDC^dn\log(n))$, where $C$ is a constant independent of  $n$, $D$,  $N$,  and $d$.  Thus, if $\mathbf{W}$ is constructed using cover trees~\cite{beygelzimer2006cover} with $N$ nearest neighbors, and $O(1)$ eigenvectors of $\mathbf{P}$ are used to calculate diffusion distances, then the computational complexity of D-VIC is $O((D^2+m^4 + m^2I)n +~NDC^dn\log(n))$~\cite{maggioni2019learning, beygelzimer2006cover}. 

So long as the spatial dimensions of the scene captured by an HSI are not changed, we expect that $m$ (the expected number of materials within the scene) will be constant with respect to the number of samples $n$. Similarly, numerical simulations have shown that, if $m$ remains constant as the number of samples increases, then $I$ tends to grow only slightly~\cite{chan2011simplex}. If $m=O(1)$ and $I=O(\log(n))$ with respect to $n$, then the complexity of D-VIC reduces to $O(NDC^dn\log(n))$ (i.e., quasilinear in the image size).

\subsection{Comparison with Learning by Unsupervised Nonlinear Diffusion}

An important point of comparison for D-VIC is the \emph{Learning by Unsupervised Nonlinear Diffusion} (\emph{LUND}) algorithm~\cite{maggioni2019learning, murphy2018unsupervised}, which follows a similar procedure to D-VIC but crucially uses the KDE $p(x)$ in place of $\zeta(x)$.  To give some motivation for why we advocate for $\zeta$ instead of $p$ for material clustering, we remark that for any one cluster, there may be multiple reasonable choices for cluster modes: pixels that are exemplary of underlying cluster structure. In LUND, cluster modes are selected to be high-density pixels that are far in diffusion distance from other high-density pixels. However, not all high-density pixels necessarily correspond to underlying material structure. A maximizer of $p(x)$ could, for example, correspond to a spatial region containing a group of commonly co-occurring materials (rather than a single material). By weighting pixel purity and density equally, D-VIC avoids selecting such a pixel as cluster mode; thus, D-VIC modes will be both indicative of underlying material structure and modal, making these points better exemplars of underlying material structure than the modes selected by LUND.

 \begin{figure}[t] 
 \centering
        \includegraphics[height = 2.1in]{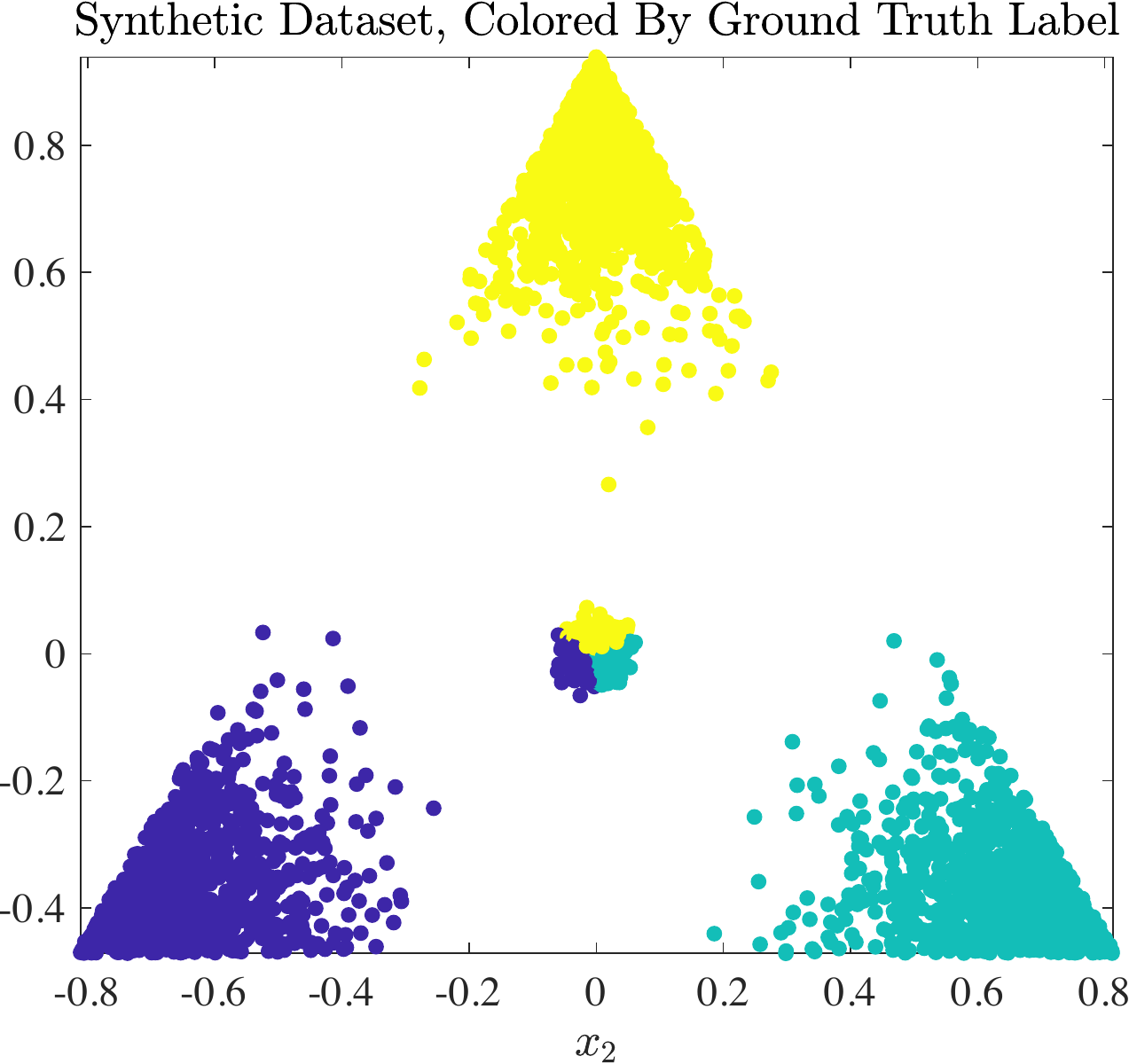}\hspace{0.25in} 
        \includegraphics[height = 2.1in]{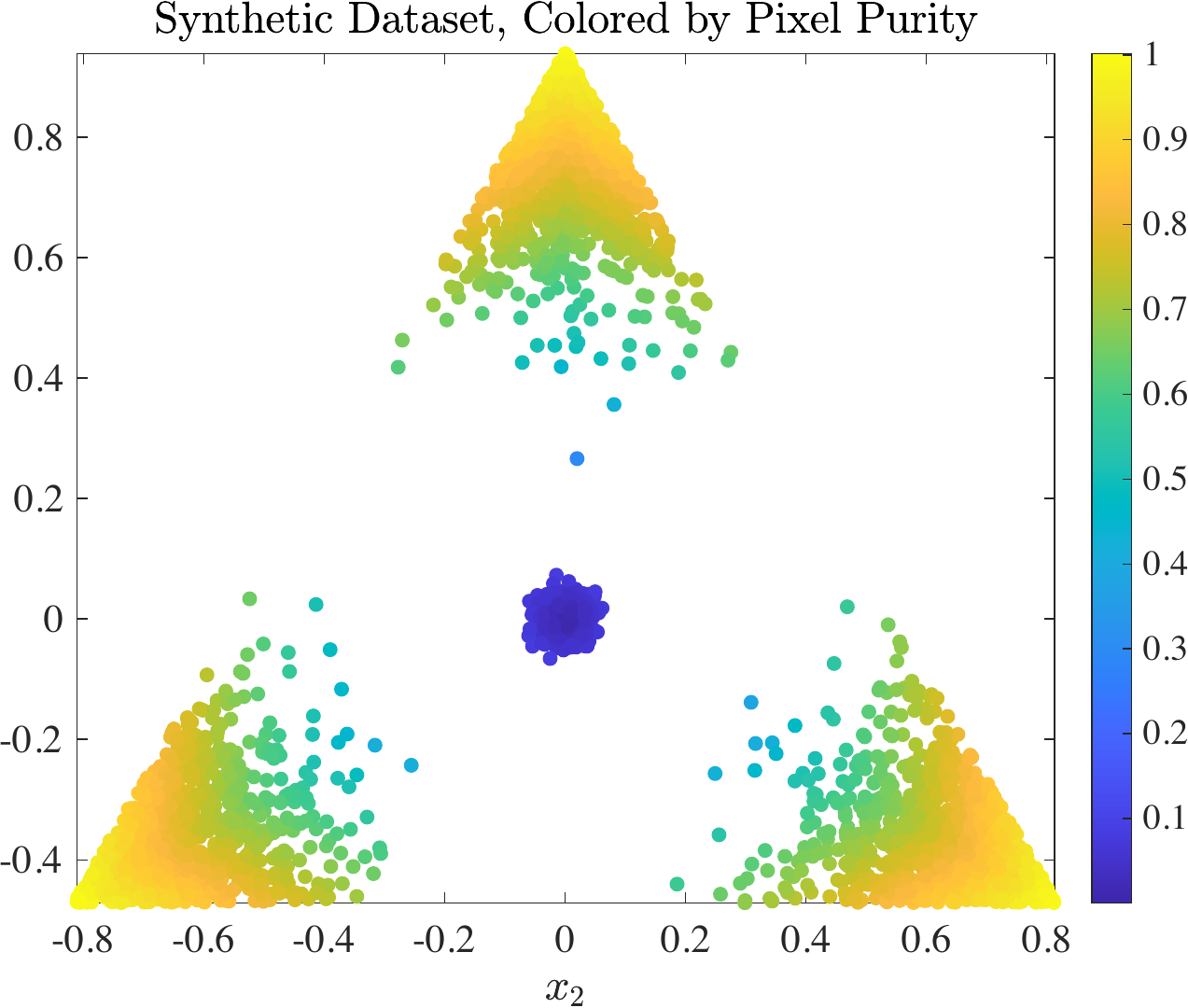}
    \caption{Ground truth labels and pixel purity of synthetic dataset sampled from a triangle in $\R^2$; the $K=3$ vertices of this triangle served as ground truth endmembers. Notice that empirical density maximizers near the origin are also lowest-purity data points. }
         \label{fig:syntheticGT} 
     \end{figure}
     
We demonstrate this key difference between LUND and D-VIC by implementing both algorithms on a simple dataset (visualized in Fig. \ref{fig:syntheticGT}) built to illustrate the idealized scenario where D-VIC outperforms LUND due to its incorporation of pixel purity. This dataset was generated by sampling $n=5000$ points from an equilateral triangle in $\R^2$ centered at the origin with edge length $\sqrt{2}$;  the $K=3$ vertices of this triangle served as ground truth endmembers. We sampled 1000 data points from a Gaussian distribution with a standard deviation of 0.175 centered at each endmember, keeping only the samples lying within in the convex hull of the ground truth endmember set. In addition, 2000 data points were sampled from a Gaussian distribution with zero-mean and a smaller standard deviation of 0.0175. As such, high-purity points indicative of latent material structure were also relatively low-density, and density maximizers were engineered so as not to be indicative of latent material structure. Each point was assigned a ground truth label corresponding to its highest-abundances ground truth endmember.

  \begin{figure}[t]
     \centering 
    \includegraphics[height = 2.1in]{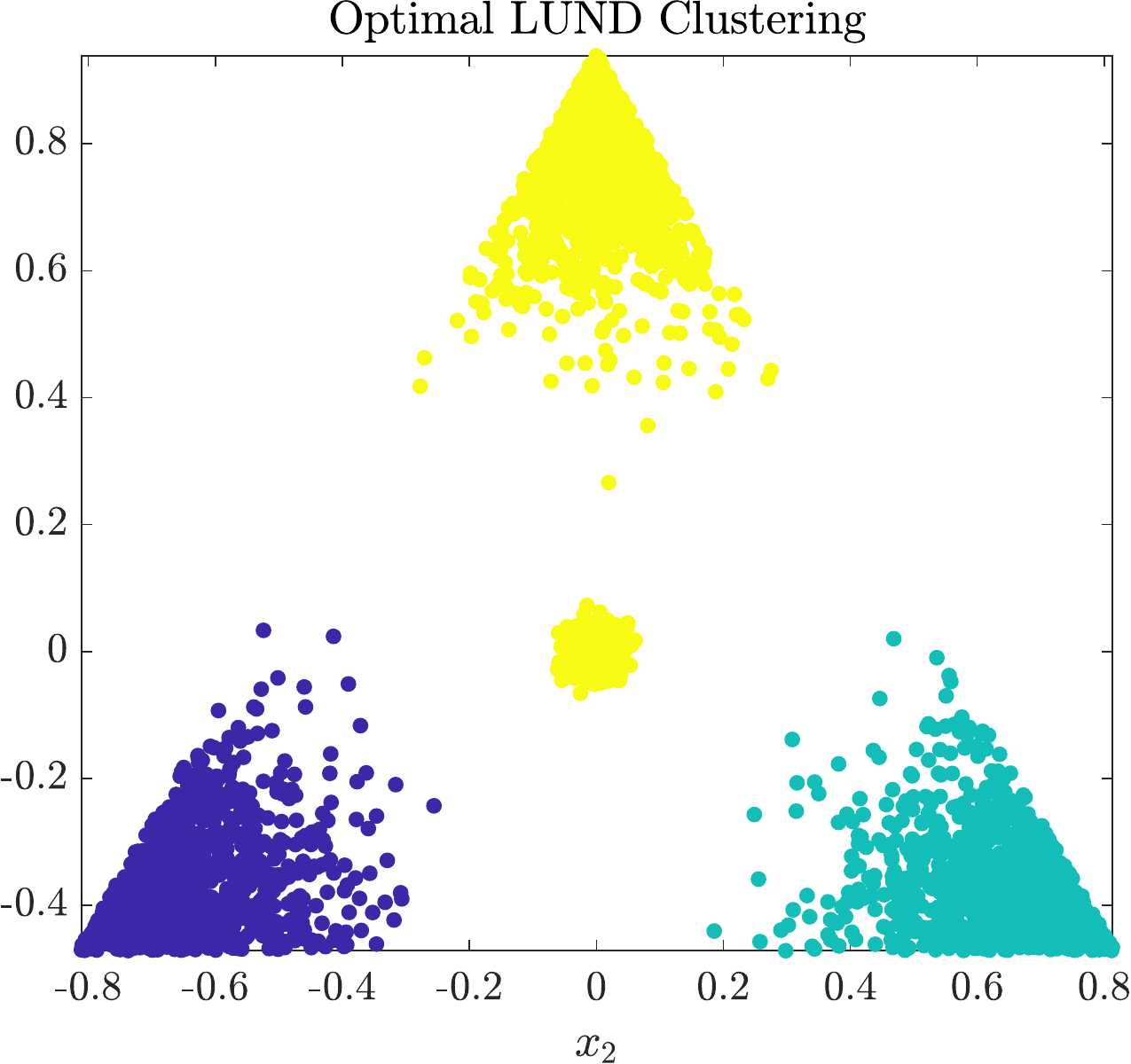}\hspace{0.25in} 
    \includegraphics[height = 2.1in]{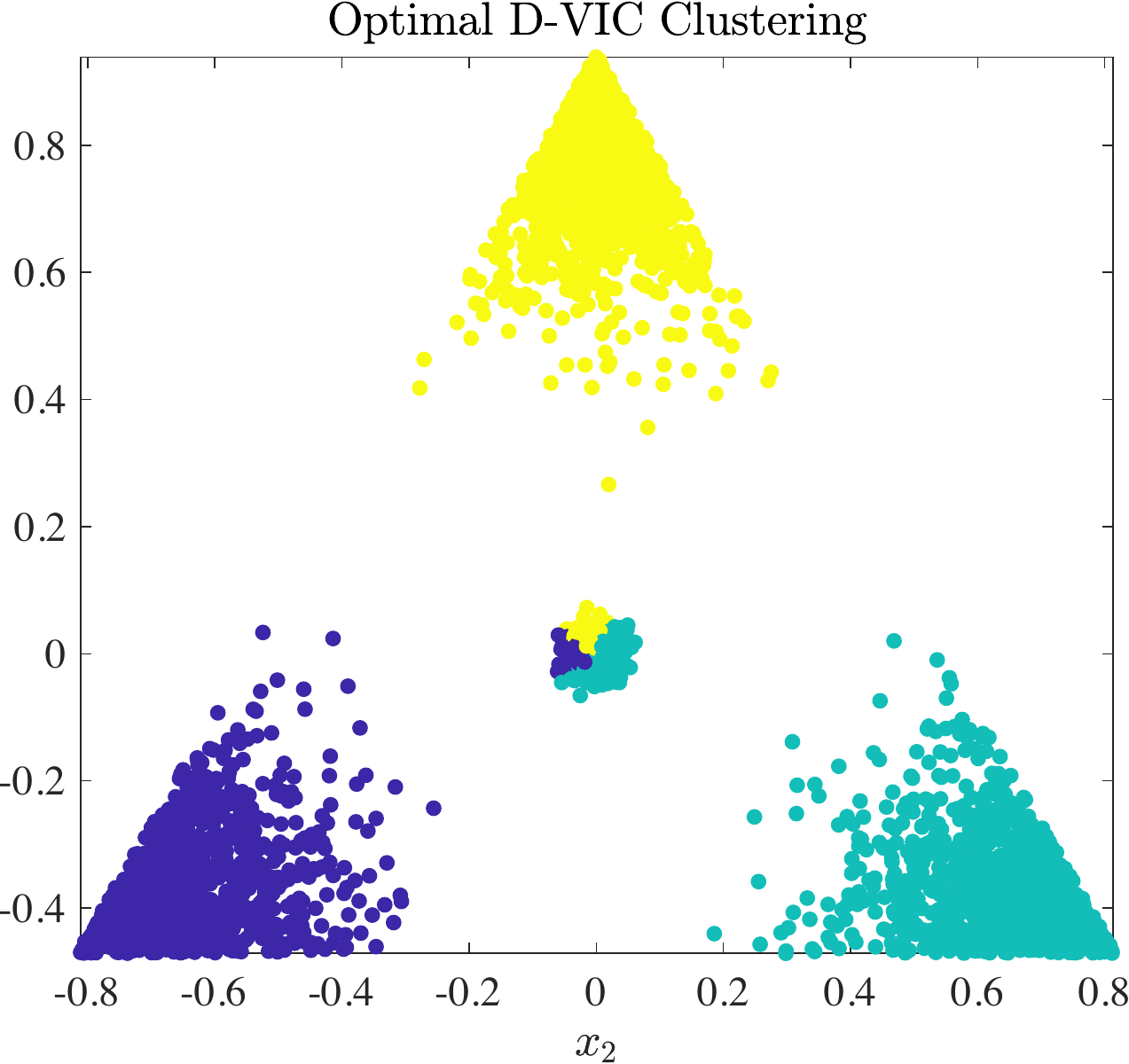} 
         \caption{Optimal LUND (OA = 0.739) and D-VIC (OA~=~0.905) clusterings of the synthetic dataset (Fig. \ref{fig:syntheticGT}). D-VIC explicitly incorporates data purity into its labeling procedure, resulting in better clustering performance than LUND in the high-density, low-purity region near the origin. 
     \label{fig:syntheticclusterings}}
 \end{figure}

For both LUND and D-VIC, \emph{overall accuracy} (\emph{OA}), defined to be the fraction of correctly labeled pixels, was optimized for across the same grid of relevant hyperparameter values (see Appendix  \ref{app: grid search}). The optimal clusterings and their corresponding OA values are provided in Fig. \ref{fig:syntheticclusterings}.  These results illustrate a fundamental limitation of relying solely on empirical density to select cluster modes in spectrally mixed HSI data. Because empirical density maximizers are not representative of underlying material structure in this synthetic dataset, LUND is unable to accurately cluster data points within the high-density, low-purity region near the origin, resulting in poor performance and an OA of 0.739. In contrast, D-VIC downweights high-density points that are not also high-purity and therefore selects points that are more representative of the dataset's underlying material structure as cluster modes. As a result, D-VIC correctly separates the high-density, low-purity region into three segments, yielding a substantially higher OA of 0.905: a difference of 0.166 when compared to LUND.  We note that both LUND and D-VIC are related to classical spectral graph clustering methods \cite{shi2000normalized, ng2002spectral, coifman2005geometric} in their use of a diffusion process on the graph to learn the intrinsic geometry in the high-dimensional data, but differ in their use of data density (LUND) and data purity (D-VIC) in identifying cluster modes as well as in their use of an iterative labeling scheme.

\section{Experiments and Discussion} \label{sec: numerical experiments}

This section contains a series of experiments indicating the efficacy of D-VIC. First, in Section \ref{sec: Benchmark Data Analysis}, classical and state-of-the-art clustering algorithms were implemented on three real, benchmark HSIs. D-VIC was compared against classical algorithms~\cite{friedman2001elements}: $K$-Means, $K$-Means applied to the first principal components of the HSI ($K$-Means+PCA), and GMM applied to the first principal components of the HSI (GMM+PCA). D-VIC was also compared against several state-of-the-art HSI clustering algorithms: Density Peak Clustering (DPC)~\cite{rodriguez2014clustering}, Spectral Clustering (SC)~\cite{ng2002spectral, shi2000normalized}, Symmetric Nonnegative Matrix Factorization (SymNMF)~\cite{kuang2012symmetric}, K-Nearest Neighbors Sparse Subspace Clustering (KNN-SSC)~\cite{abdolali2021beyond, zhuang2016locality}, Fast Self-Supervised Clustering (FSSC)~\cite{wang2021fast} and LUND~\cite{maggioni2019learning, murphy2018unsupervised}.  Our second set of experiments appears in Section \ref{sec: Madingley Data Analysis}, where D-VIC and other clustering algorithms were implemented on a remote sensing HSI generated over deciduous forest containing both healthy and dieback-infected ash trees in Madingley Village near Cambridge, United Kingdom~\cite{chan2021monitoring, polk2022unsupervised}. 

In all experiments, the number of clusters was set equal to the ground truth $K$.   Comparisons were made using OA and Cohen's $\kappa$ coefficient: $\kappa = \frac{p_{o}-p_e}{1-p_e}$, where $p_{o}$ is the relative observed agreement between a clustering and the ground truth labels and $p_e$ is the probability that a clustering agrees with the ground truth labels by chance \cite{cohen1960kappa}.  OA is a standard metric that, in some ways, captures the best sense of overall performance, as each pixel is considered of equal importance.  However, it is biased in favor of correctly labeling large clusters at the expense of small clusters and can be misleading when a dataset has many small clusters of importance.  To address this, we also consider $\kappa$; we note that in our experimental results, performance with respect to OA and $\kappa$ were highly correlated.  OA was optimized for across hyperparameters ranging a grid of relevant values for each algorithm (see Appendix \ref{app: grid search}). We report the median OA across 100 trials for K-Means, GMM, SymNMF, FSSC, and D-VIC to account for the stochasticity associated with random initial conditions. Diffusion distances were computed using only the first 10 eigenvectors of $\mathbf{P}$ in LUND and D-VIC. For D-VIC, AVMAX was run 100 times in parallel, and the endmember set that formed the largest-volume simplex was selected for later cluster analysis.

\subsection{Analysis of Benchmark HSI Datasets}\label{sec: Benchmark Data Analysis}
\noindent

 \begin{figure}[t]
 \captionsetup[subfigure]{justification=centering}
    \centering
     
    \vspace{0.05in}  
    
     \begin{subfigure}[t]{0.32\textwidth}
         \centering
        \includegraphics[height = 1.55in]{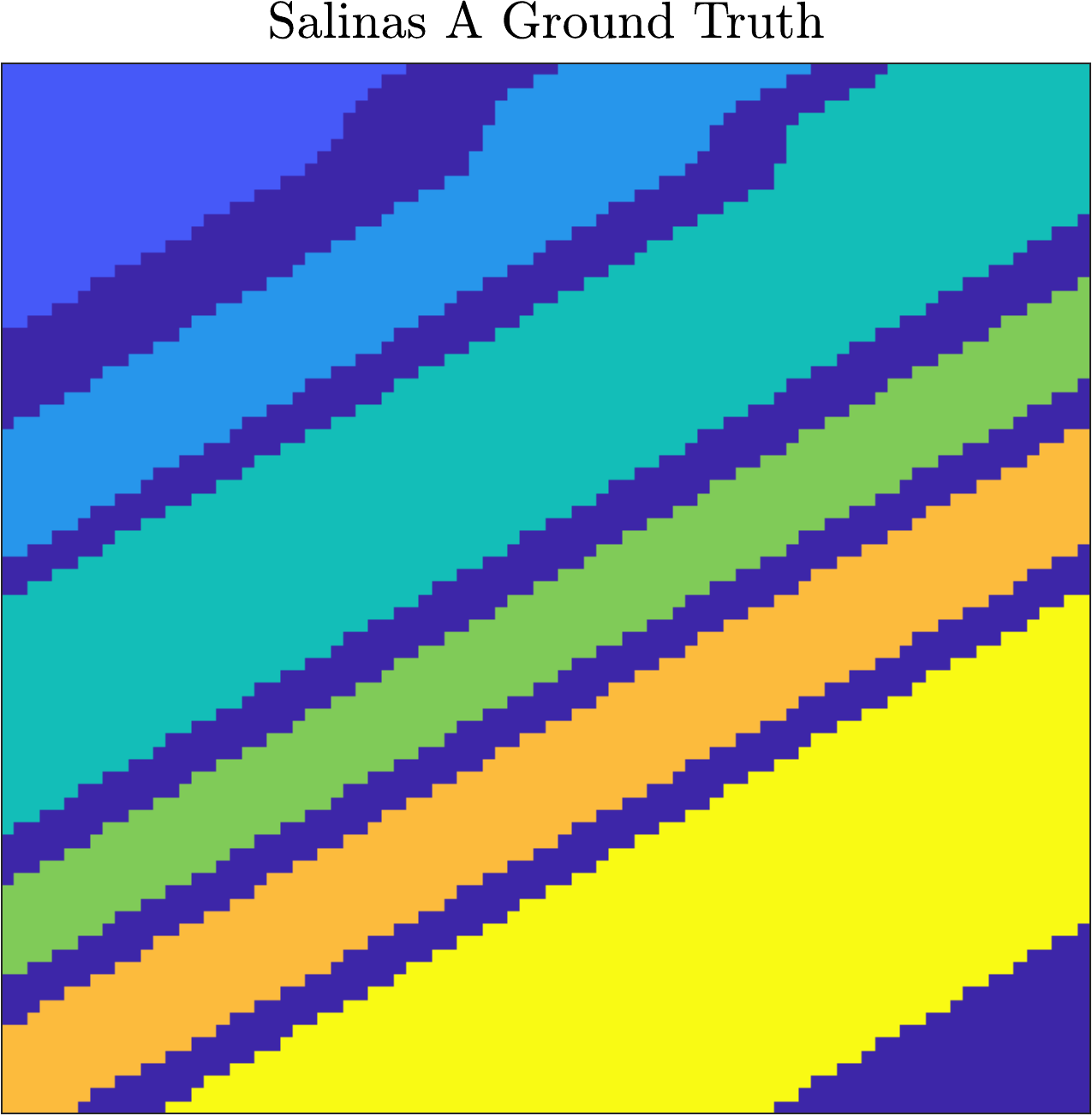} \hspace{0.14in}

        \vspace{0.1in}
        \includegraphics[height = 1.55in]{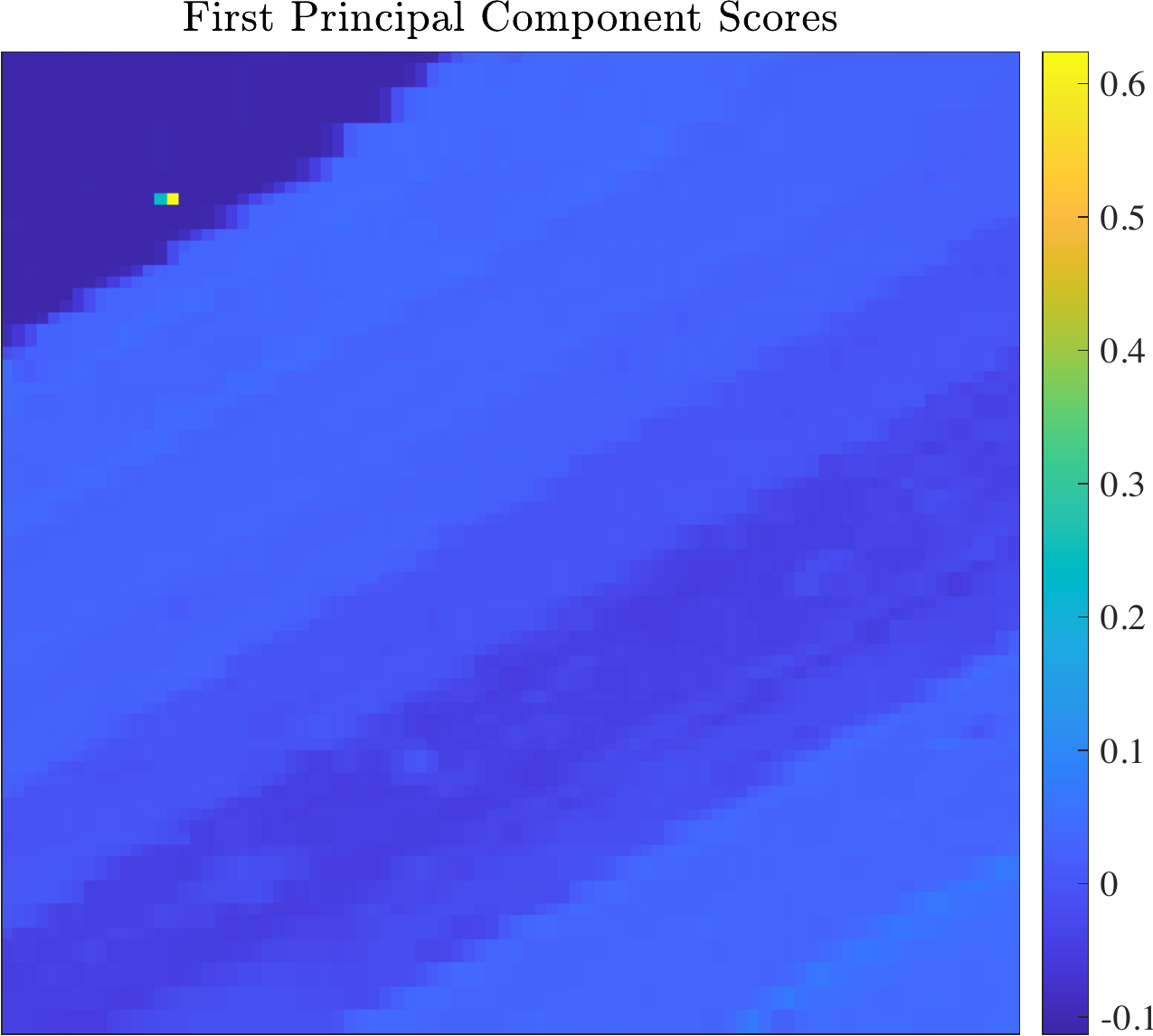}
             \caption{Salinas A}
         \label{fig:salinasAGT}
     \end{subfigure}
          \begin{subfigure}[t]{0.32\textwidth}
         \centering
        \includegraphics[height = 1.55in]{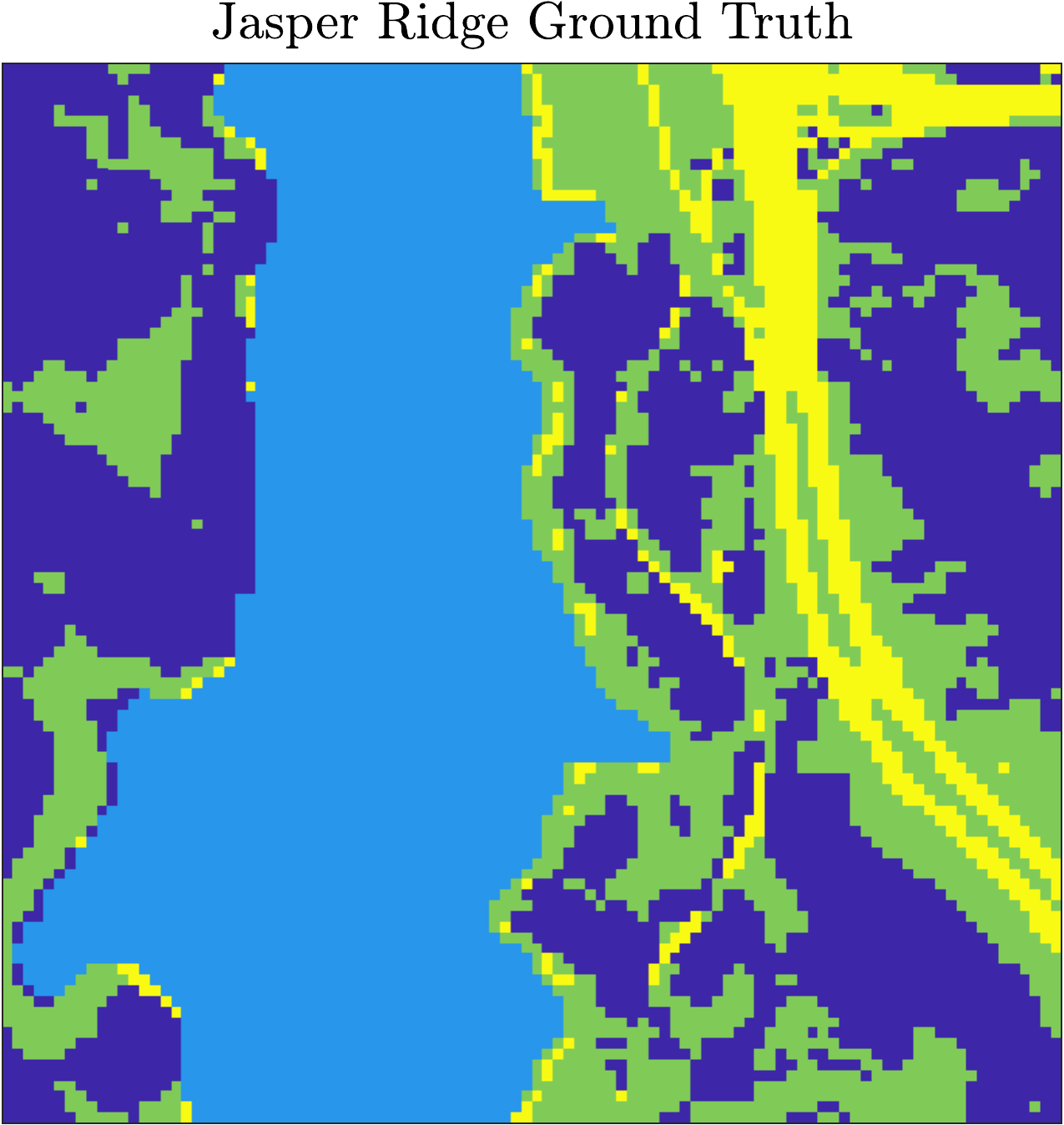} \hspace{0.14in}
        
        \vspace{0.1in}
        \includegraphics[height = 1.55in]{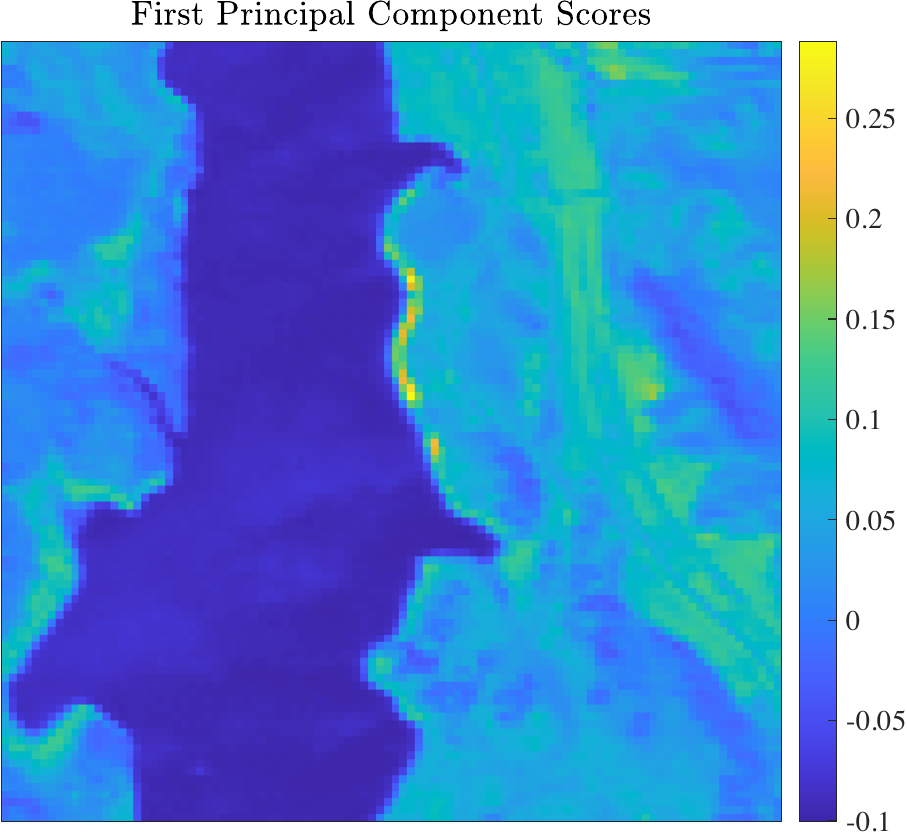}
             \caption{Jasper Ridge}
         \label{fig:jasperGT}
     \end{subfigure}
          \begin{subfigure}[t]{0.32\textwidth}
         \centering
        \includegraphics[height = 1.55in]{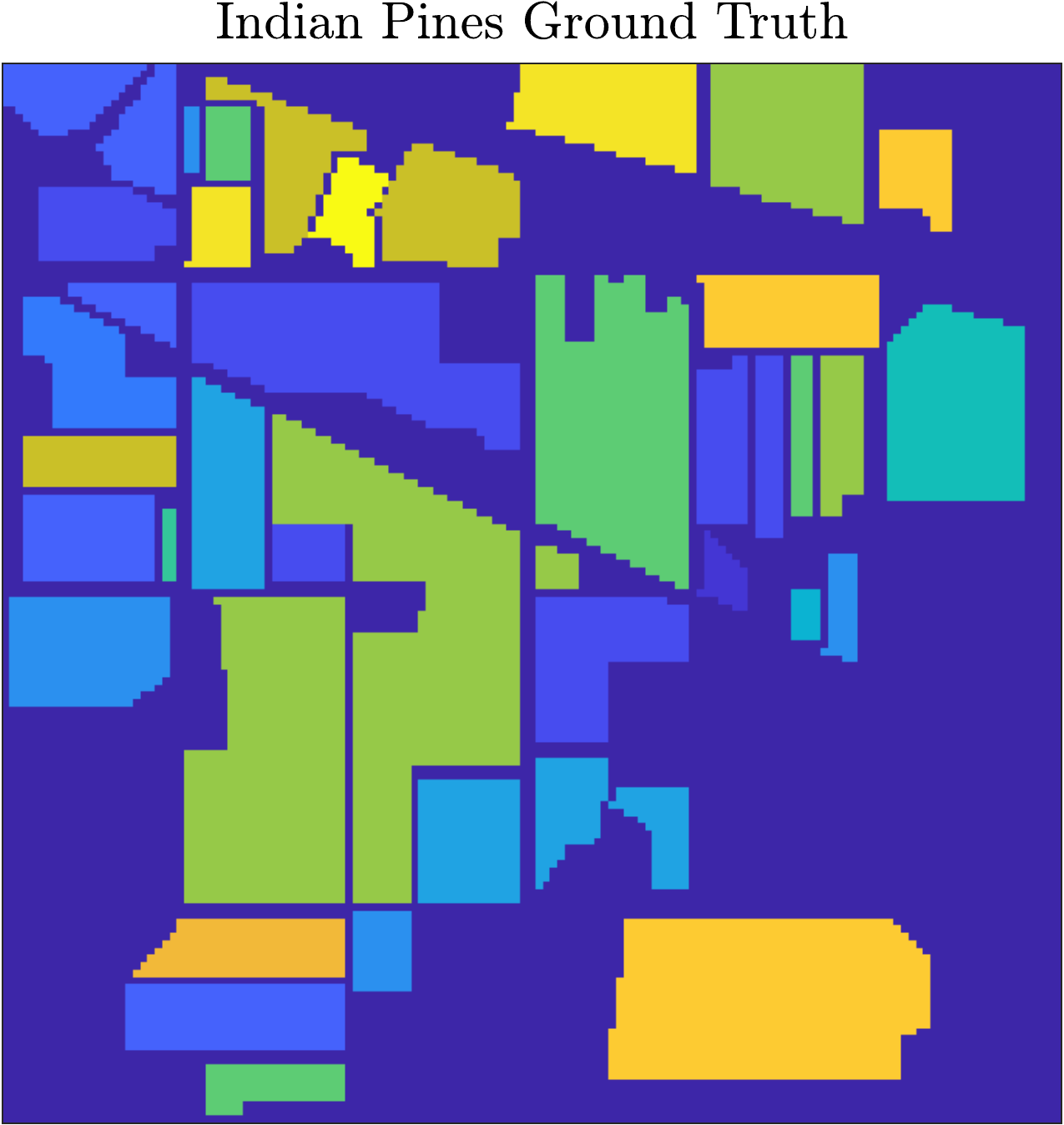} \hspace{0.14in}
        
        \vspace{0.1in}
        \includegraphics[height = 1.55in]{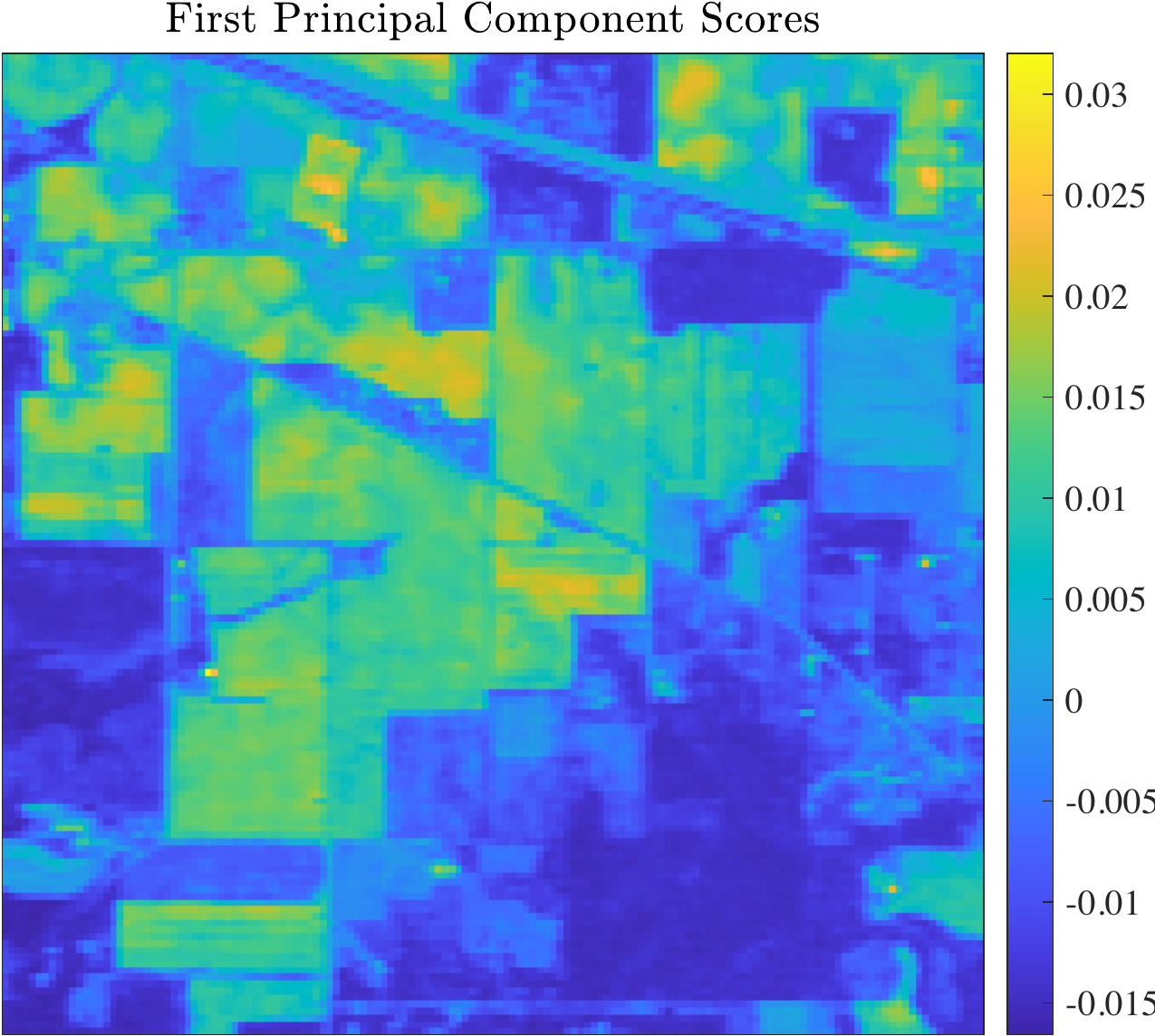}
             \caption{Indian Pines}
         \label{fig:indianPinesGT}
     \end{subfigure}

    \caption{Ground truth labels and first principal component scores for the real benchmark HSIs analyzed in this article. }
    \label{fig:benchmark summary}
\end{figure}  

\begin{table}[b]
    \centering
    \small
        \caption{ Summary of benchmark HSI datasets analyzed in Section \ref{sec: Benchmark Data Analysis}. }
    \label{tab: benchmark summary}
    \resizebox{1\textwidth}{!}{%
    \begin{tabular}{c|c|c|c|c|c|c}
    \toprule
        Dataset         &   Spatial Resolution  &   Spectral Range  &   Spatial Dimensions  &   Num. Pixels &      Num. Spectral Bands &    Num. Clusters   \\
        \midrule
        Salinas A       &   1.3~m               &   380-2500~nm     &   $83\times 86$       &   $n= 7138$   &      $D=224$             &    $K=6$           \\
        Jasper Ridge    &   5.0~m               &   380-2500~nm     &   $100\times100$      &   $n= 10000$  &      $D=224$             &    $K=4$           \\
        Indian Pines    &   20~m                &   400-2500~nm     &   $145\times145$      &   $n=21025$   &      $D=224$             &    $K=16$          \\
        \bottomrule
    \end{tabular}
}
\end{table}

To illustrate the efficacy of D-VIC, we analyzed three publicly available, real HSIs often used as benchmarks for new HSI clustering algorithms; see  Table \ref{tab: benchmark summary} and Fig. \ref{fig:benchmark summary}. Water absorption bands were discarded, and pixel reflectance spectra were standardized before analysis \cite{swinfield2020imaging}. We clustered entire images but discarded unlabeled pixels when comparing clusterings to the ground truth labels. Below, each benchmark HSI analyzed in this section is overviewed in detail; see Table \ref{tab: benchmark summary} for summary statistics on these benchmark HSIs.  \\

\noindent \textbf{Salinas A} (Fig. \ref{fig:salinasAGT}) was recorded by the Airborne Visible/Infrared Imaging Spectrometer (AVIRIS) sensor over farmland in Salinas Valley, California, USA in 1998 at a spatial resolution of 1.3~m. Spectral signatures, ranging in recorded wavelength from 380~nm to 2500~nm across 224 spectral bands, were recorded across $83\times 86$ pixels ($n=7138$). Gaussian noise (with mean 0 and standard deviation $= 10^{-7}$) was added to each pixel to differentiate two pixels with identical spectral signatures. The Salinas A scene  contains $K=6$ ground truth classes   corresponding to crop types.\\

\noindent \textbf{Jasper Ridge} (Fig. \ref{fig:jasperGT}) was recorded by the AVIRIS sensor over the Jasper Ridge Biological Preserve, California, USA in 1989 at a spatial resolution of 5~m. Spectral signatures, ranging in recorded wavelength from 380~nm to 2500~nm across 224 spectral bands, were recorded across spatial dimensions of $100\times 100$ pixels ($n=10000$). The Jasper Ridge scene  contains $K=4$ ground truth endmembers: road, soil, water, and trees. Ground truth labels were recovered by selecting the material of highest ground truth abundance for each pixel. \\

\noindent \textbf{Indian Pines} (Fig. \ref{fig:indianPinesGT}) was recorded by the AVIRIS sensor over farmland in northwest Indiana, USA in 1992 at a low spatial resolution of 20~m. Spectral signatures, ranging in recorded wavelength from 400~nm to 2500~nm across 224 spectral bands, were recorded across spatial dimensions of $145\times 145$ pixels ($n=21025$). The Indian Pines scene  contains $K=16$ ground truth classes (e.g., crop types and manufactured structures) as well as many unlabeled pixels.

\subsubsection{Discussion of Benchmark HSI Experiments} \label{sec: performance}

\noindent This section compares clusterings produced by D-VIC against those of related algorithms (Table \ref{tab: performance}). On each of the three benchmark HSIs analyzed, D-VIC produces a clustering closer to the ground truth labels than those produced by related algorithms. In the Indian Pines (Fig. \ref{fig:indianPinesGT}) scene, pixels from the same class exist in multiple segments of the image, and the size of ground truth clusters varies substantially across the $K=16$ classes. As such, though supervised and semi-supervised HSI classification algorithms may output highly-accurate classifications of the Indian Pines HSI~\cite{kotzagiannidis2021semi, qin2018spectral, hong2020graph, sun2021supervised, kavalerov20203}, this image is expected to be challenging for fully-unsupervised clustering algorithms that rely on no ground truth labels. Nevertheless, D-VIC achieves higher performance than all other algorithms on this challenging dataset. Notably, though all other algorithms (including state-of-the-art algorithms, such as LUND) achieve $\kappa$-statistics in the same narrow range of 0.271 to 0.316, D-VIC achieves a substantially higher $\kappa$-statistic of 0.350. As such, the incorporation of pixel purity in D-VIC enables superior detection even in this difficult setting.

\begin{figure}[t]
\centering
    \centering
    \includegraphics[height = 1.45in]{figures/groundtruth/SalinasACorrectedGT-cropped.pdf} \hspace{0.02in}
    \includegraphics[height = 1.45in]{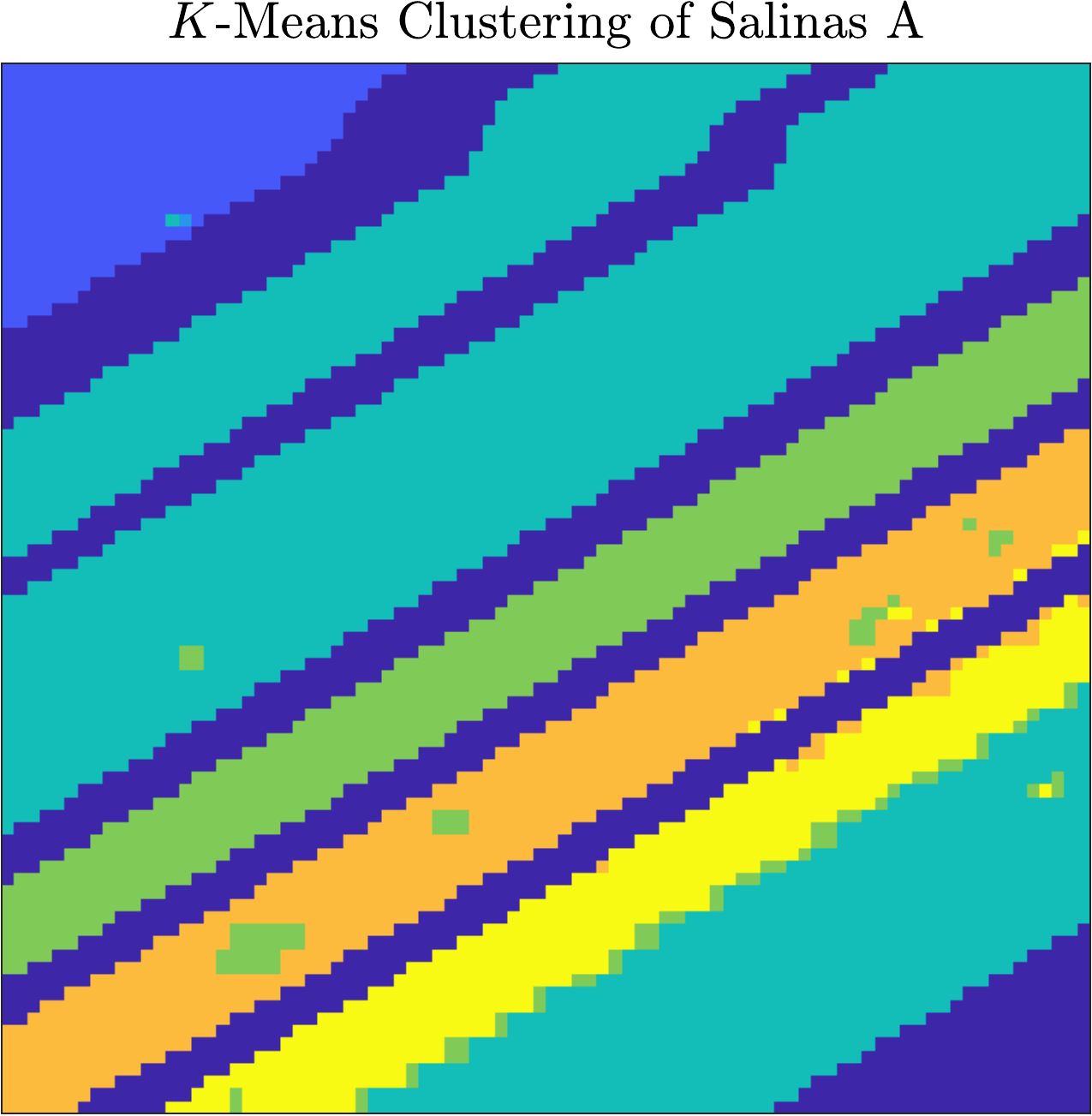} \hspace{0.02in}
    \includegraphics[height = 1.45in]{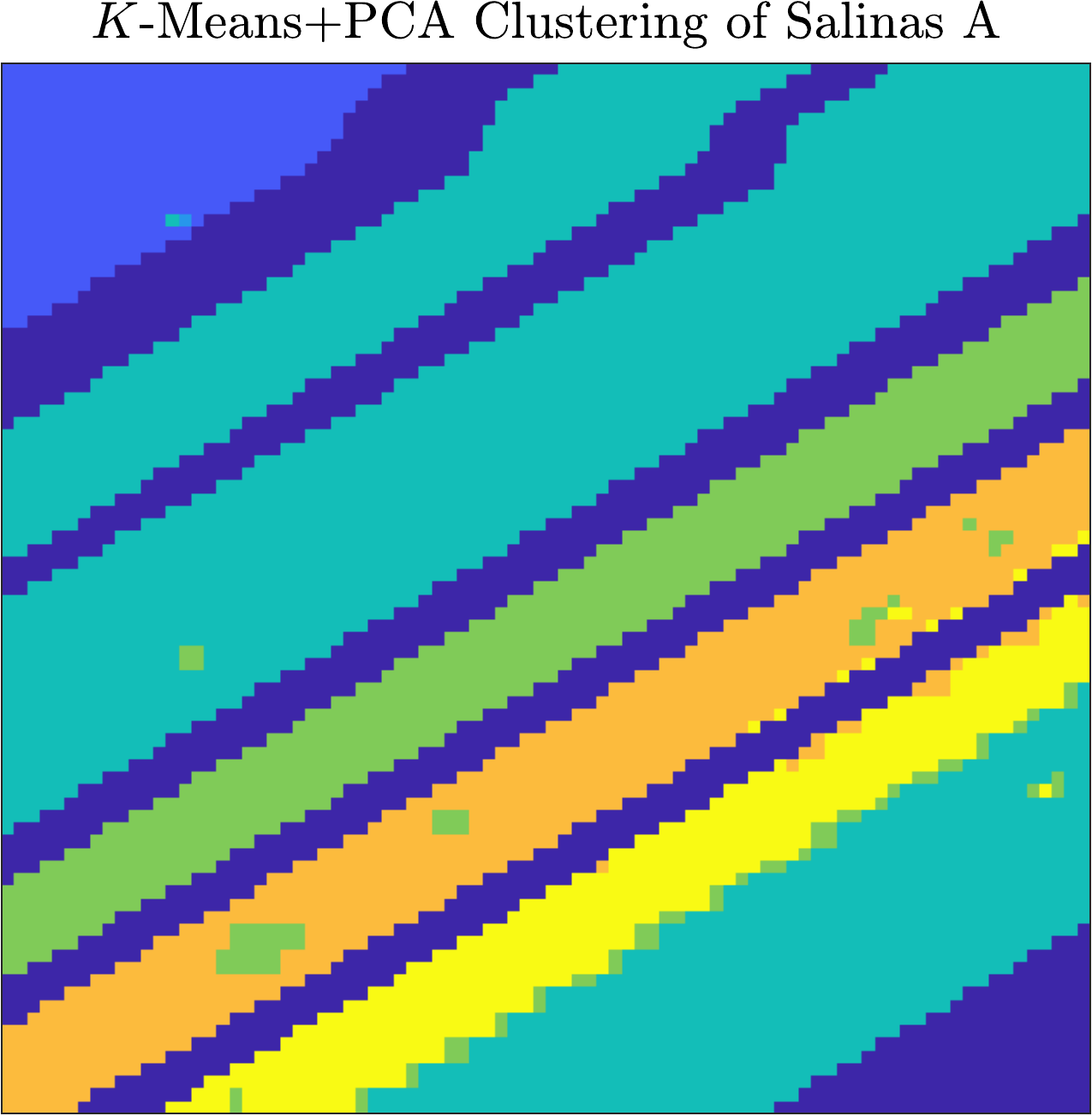}
    \hspace{0.02in}
    \includegraphics[height = 1.45in]{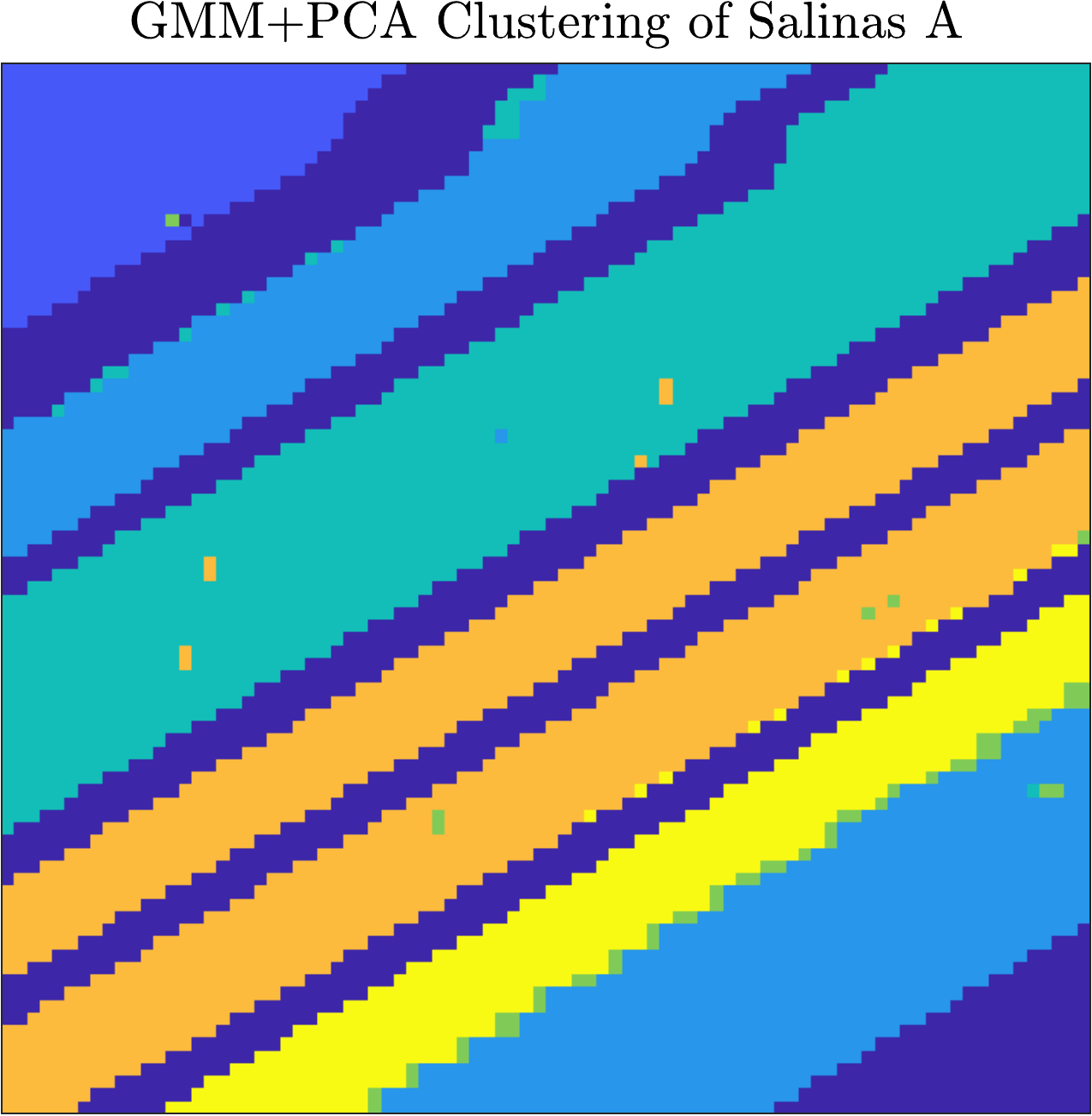}

    \vspace{0.1in}
        \includegraphics[height = 1.45in]{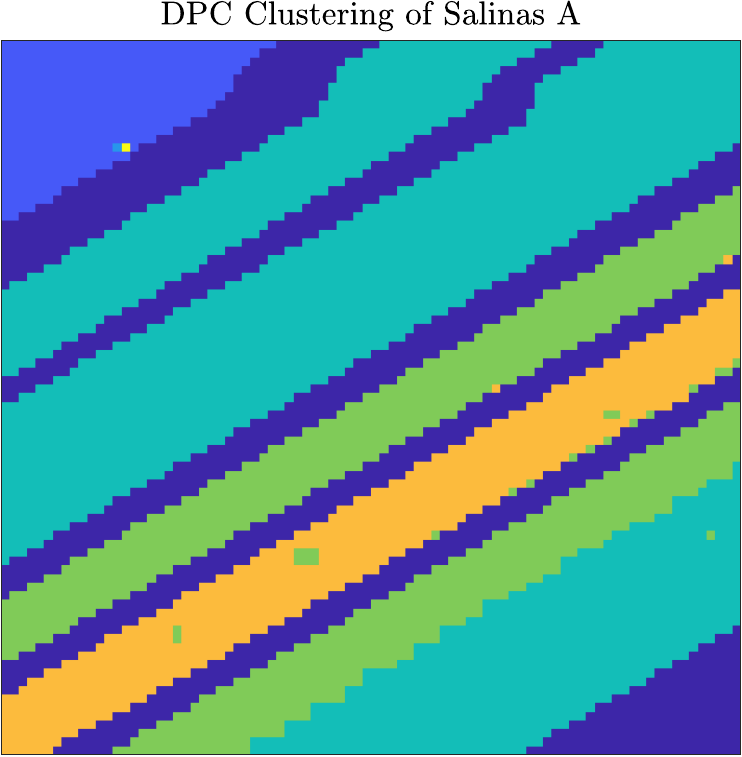}
\hspace{0.02in}
    \includegraphics[height = 1.45in]{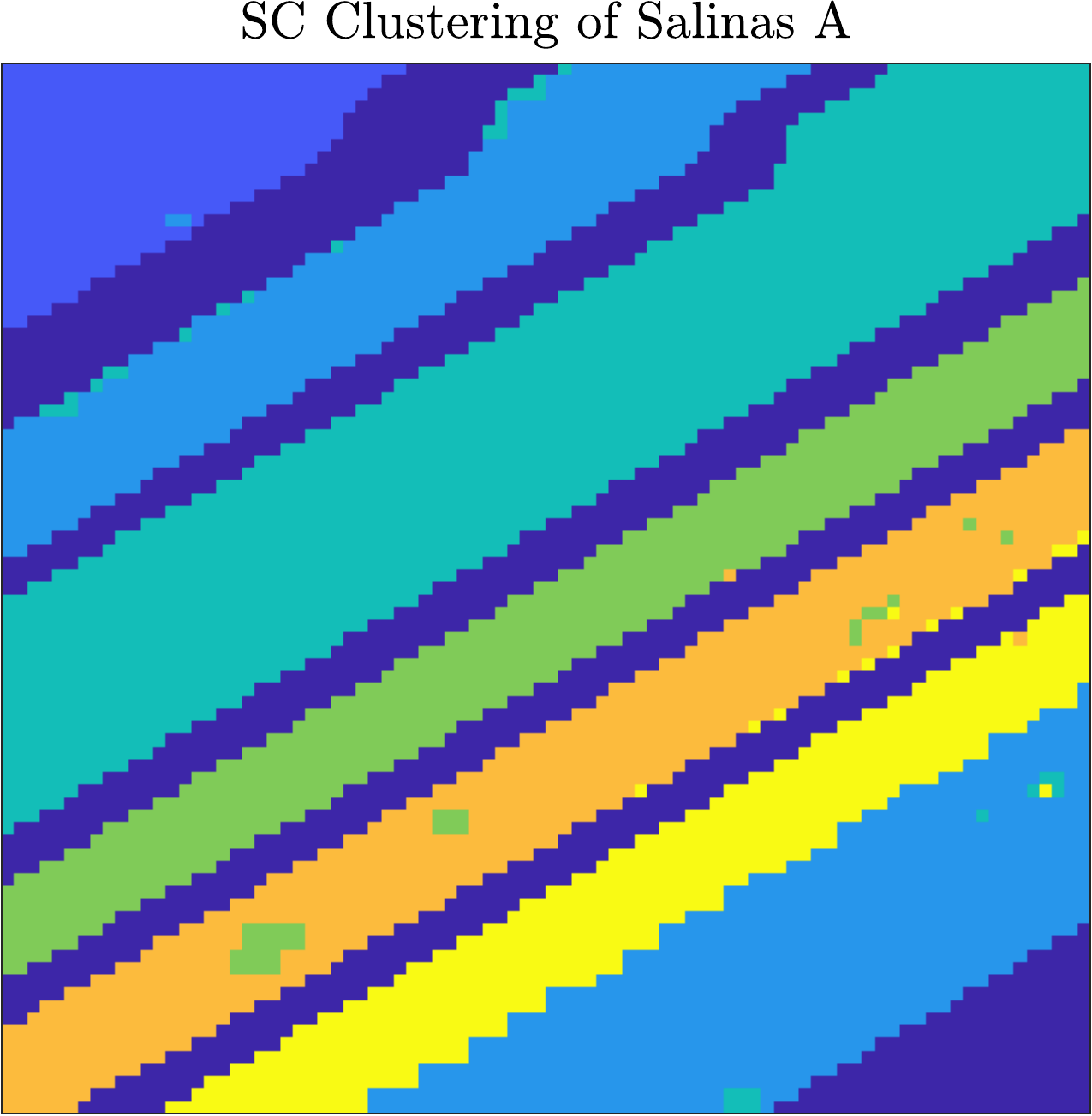}
    \hspace{0.02in}
    \includegraphics[height = 1.45in]{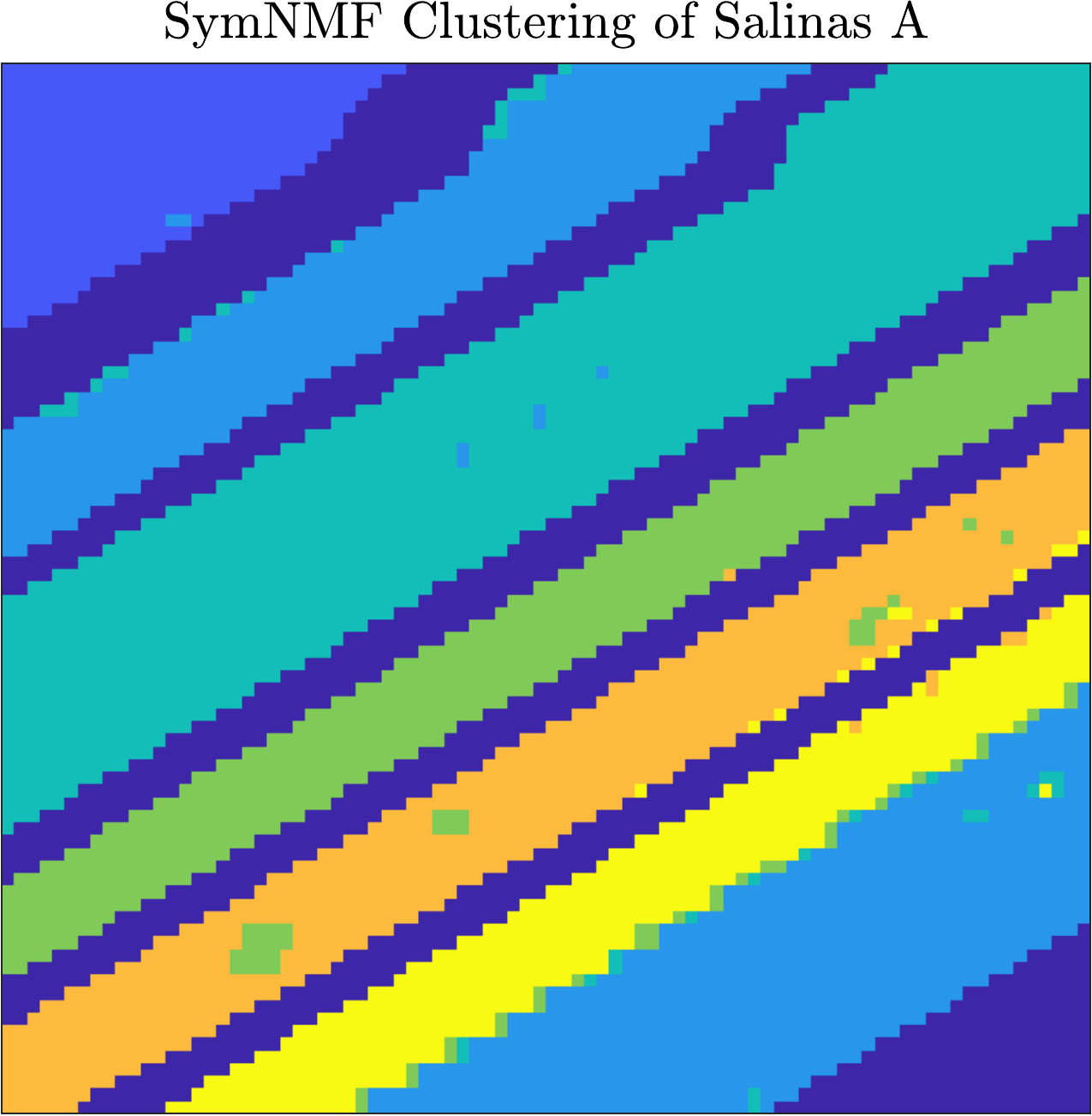}
    \hspace{0.02in}
    \includegraphics[height = 1.45in]{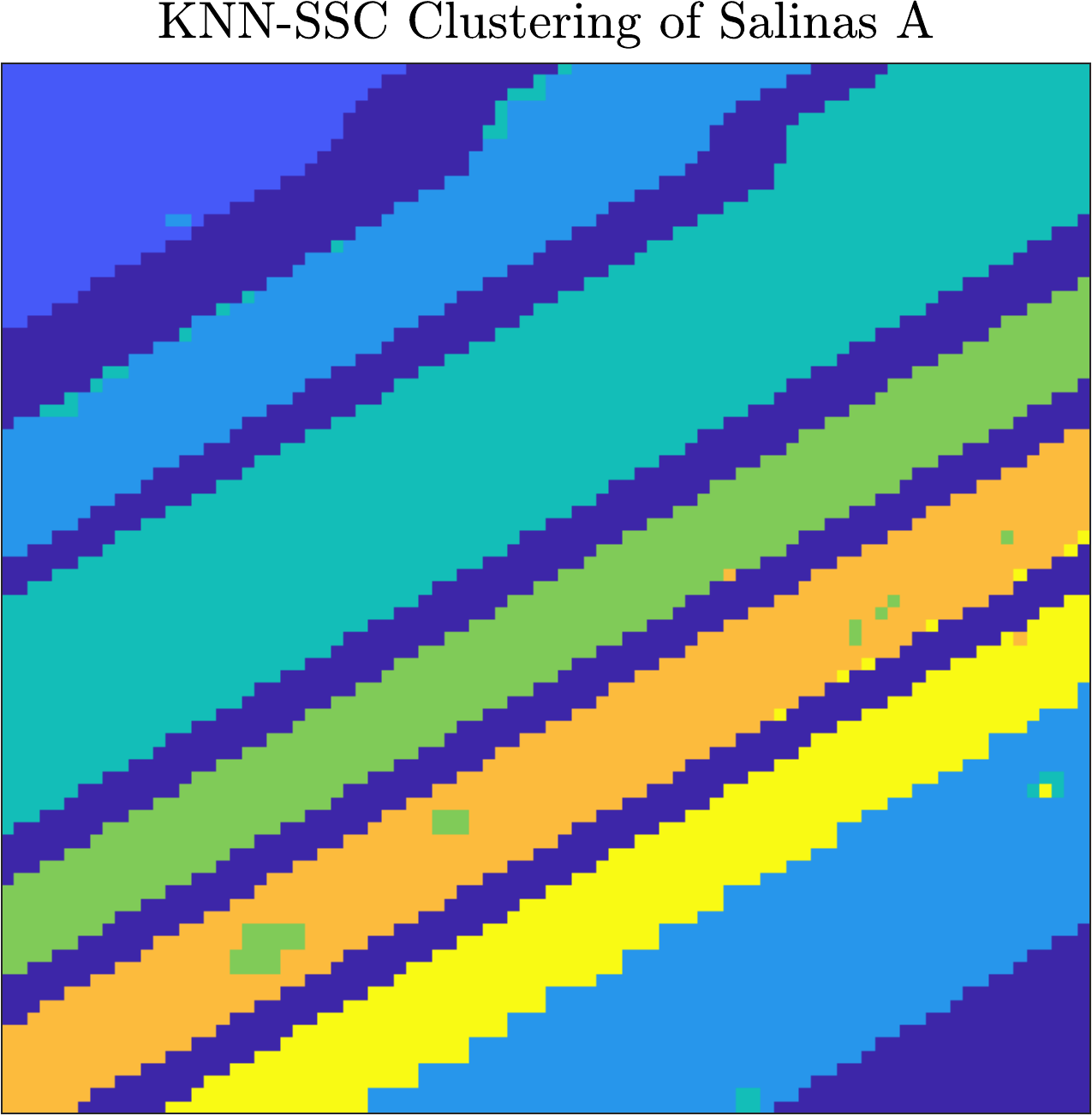}

    \vspace{0.1in}
    \includegraphics[height = 1.45in]{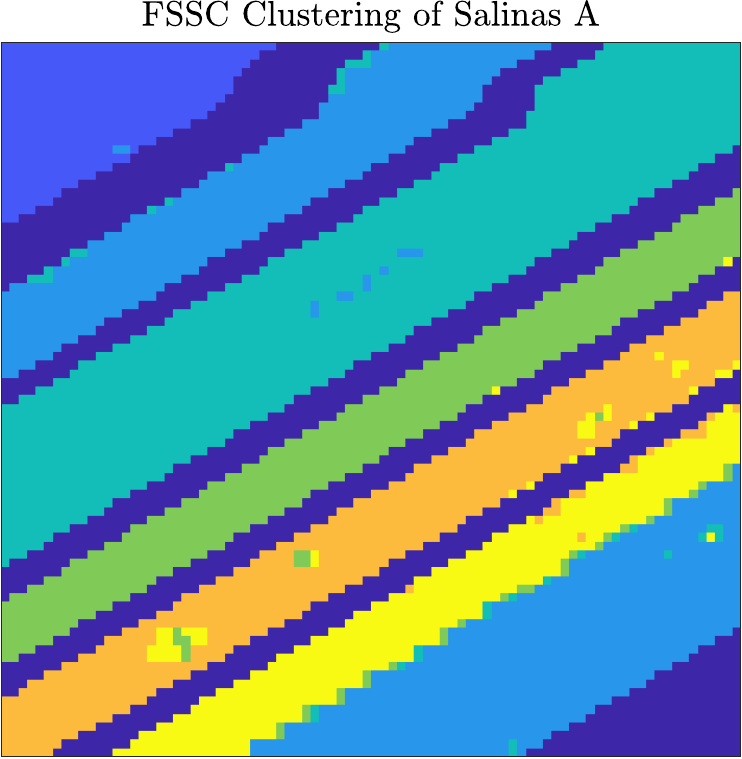}
        \hspace{0.02in}
    \includegraphics[height = 1.45in]{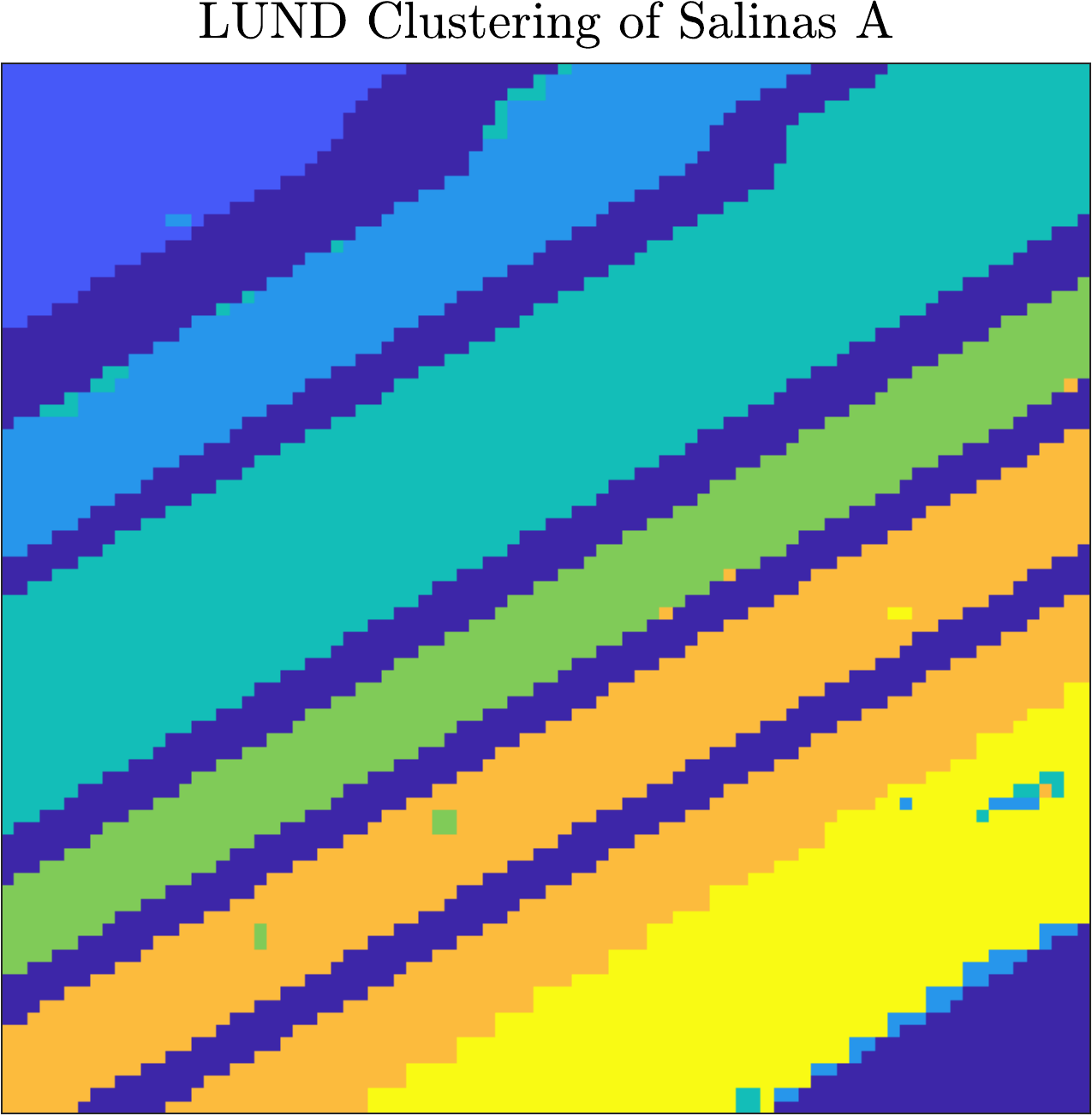}
    \hspace{0.02in}
    \includegraphics[height = 1.45in]{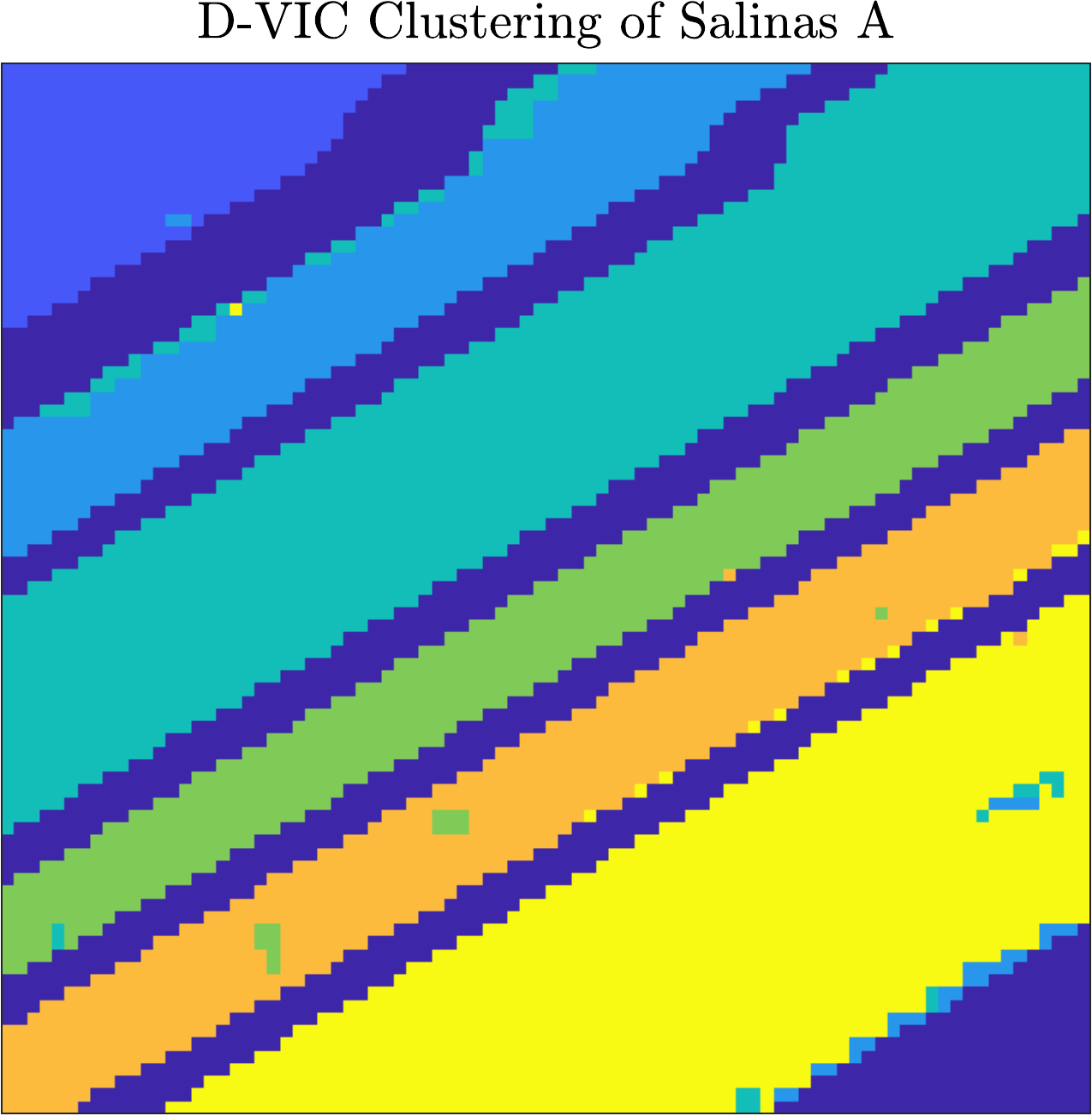}
    
\caption{Comparison of clusterings produced by D-VIC and related algorithms on the Salinas A HSI. Unlike any comparison method,    D-VIC correctly groups pixels corresponding to 8-week maturity romaine (indicated in yellow), resulting in near-perfect recovery of the ground truth labels. \label{fig:SalinasA} } 

\end{figure}

\begin{table}[t]
\small
\caption{\label{tab: performance} 
Performances of D-VIC and related algorithms on benchmark HSIs. 
Highest and second-highest performances are bolded and underlined, respectively.  D-VIC offers substantially higher performance on all datasets evaluated. 
}
\begin{tabular}{p{2.5cm}|cc|cc|cc}
		\toprule
          &
          \multicolumn{2}{c|}{Salinas A} & 
          \multicolumn{2}{c|}{Jasper Ridge} & 
          \multicolumn{2}{c}{Indian Pines}  \\
                    & OA & $\kappa$    & OA & $\kappa$ & OA & $\kappa$    \\ 
                    \midrule
                $K$-Means       &   0.764    &0.703&    0.784 &      0.703&    0.383  &  0.315\\        
                $K$-Means+PCA   &   0.764    &0.703&    0.785 &      0.703&    0.382  &  \underline{0.316}\\
                GMM+PCA         &   0.611   &   0.512&	    0.789 &      0.701&    0.364  &  0.292\\
                DPC             &   0.629     & 0.529   &   0.809   &   0.727   &   \underline{0.410}   &   0.271   \\
                SC              &   0.834    &0.797&    0.760 &      0.670&    0.382  &  0.314\\  
                SymNMF          &   0.828    &0.791&    0.662 &      0.542&    0.365  &  0.304\\  
                KNN-SSC         &   0.844   & 0.809&    0.726 &      0.629&    0.371  &  0.308\\  
                FSSC            &   0.830   & 0.793&    0.780 &      0.691 &    0.396  &  0.281\\
                LUND            &    \underline{0.887} &   \underline{0.860}&   \underline{0.815} &      \underline{0.737}&    0.404  &  0.312\\  
                D-VIC           &   \textbf{0.976}&   \textbf{ 0.970}&    \textbf{0.865}&    \textbf{0.805}&    \textbf{0.445}&    \textbf{0.350}\\ 
                \bottomrule
        \end{tabular}

\end{table}

\begin{figure}[t]
\centering
    \centering
    \includegraphics[height = 1.45in]{figures/groundtruth/JasperRidgeGT-cropped.pdf} \hspace{0.02in}
    \includegraphics[height = 1.45in]{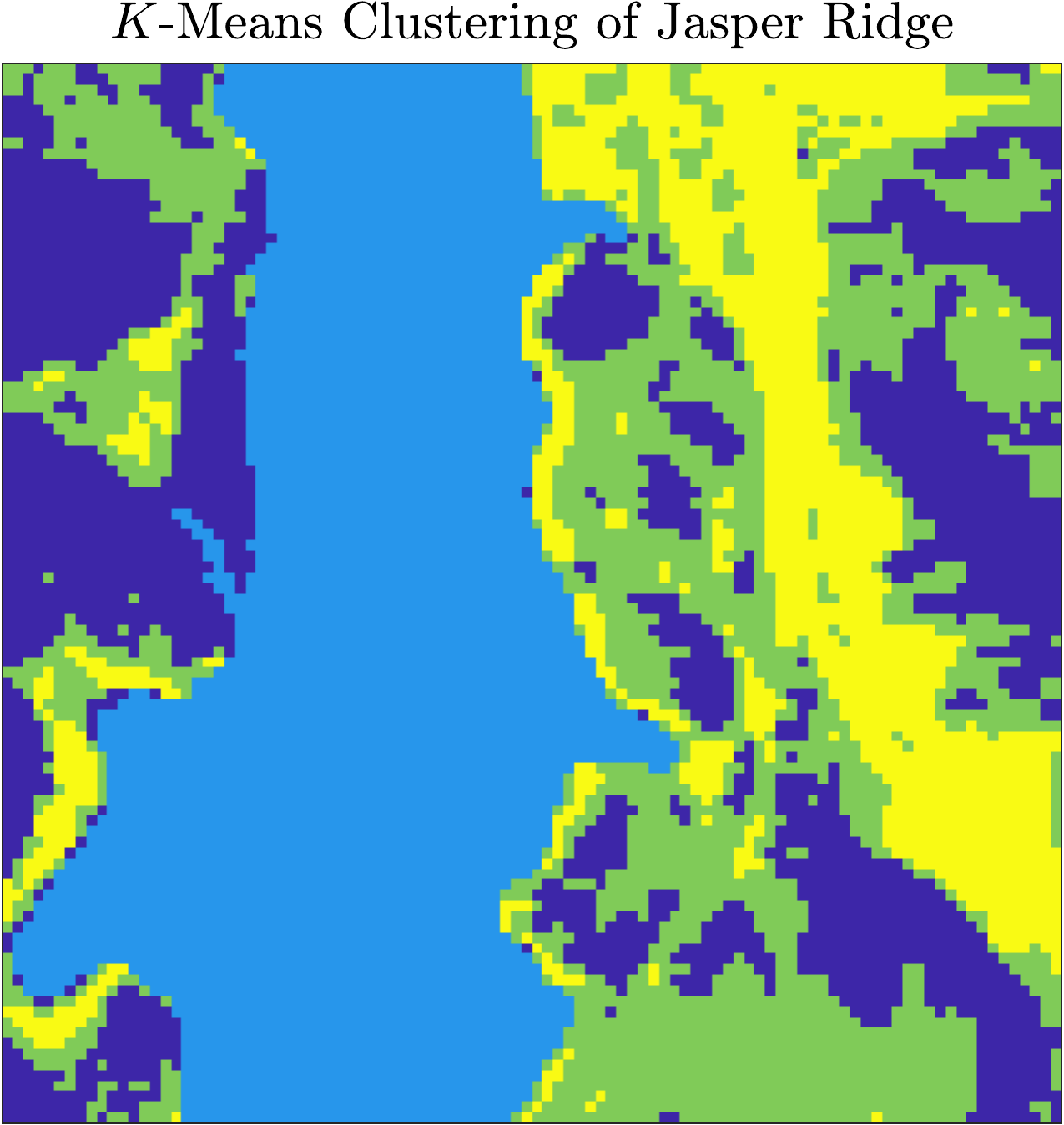} \hspace{0.02in}
    \includegraphics[height = 1.45in]{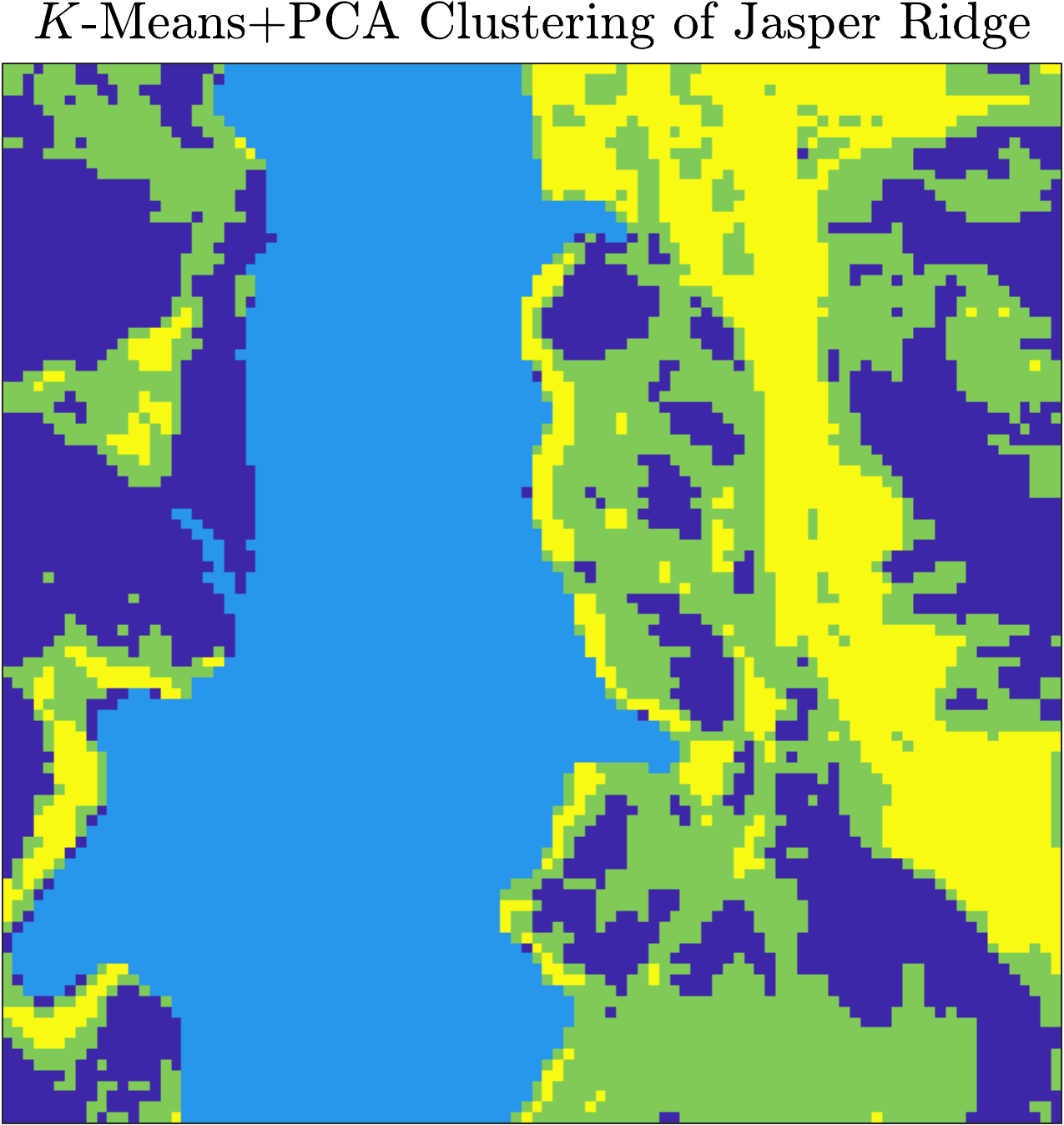}
    \hspace{0.02in}
    \includegraphics[height = 1.45in]{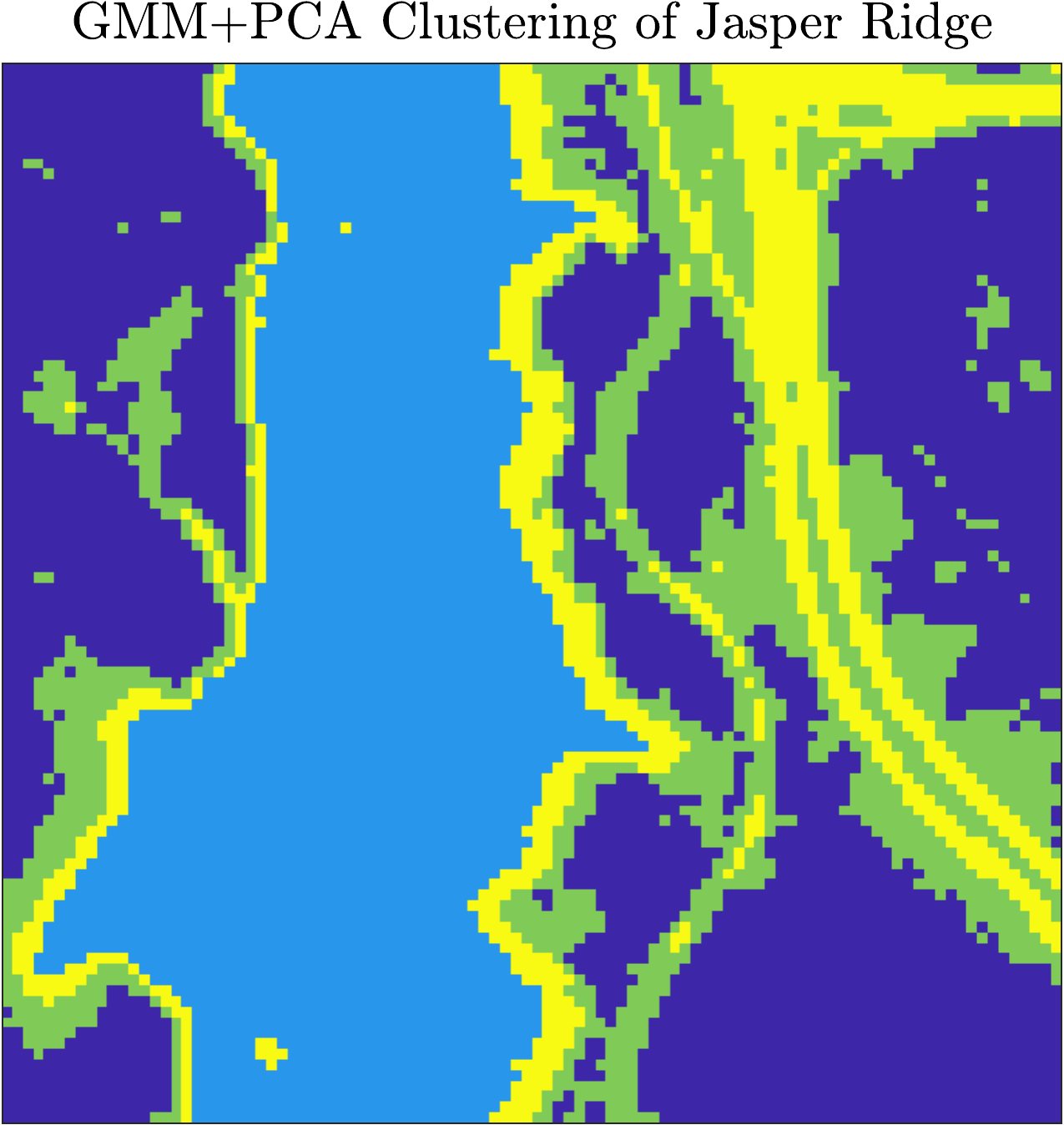}
    \vspace{0.1in}
    
    \includegraphics[height = 1.45in]{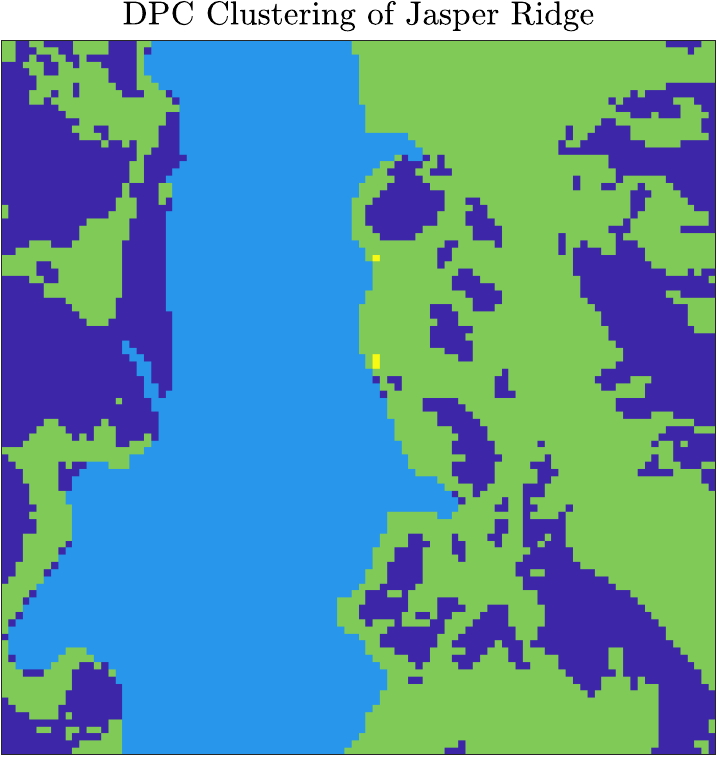}
    \hspace{0.02in}
    \includegraphics[height = 1.45in]{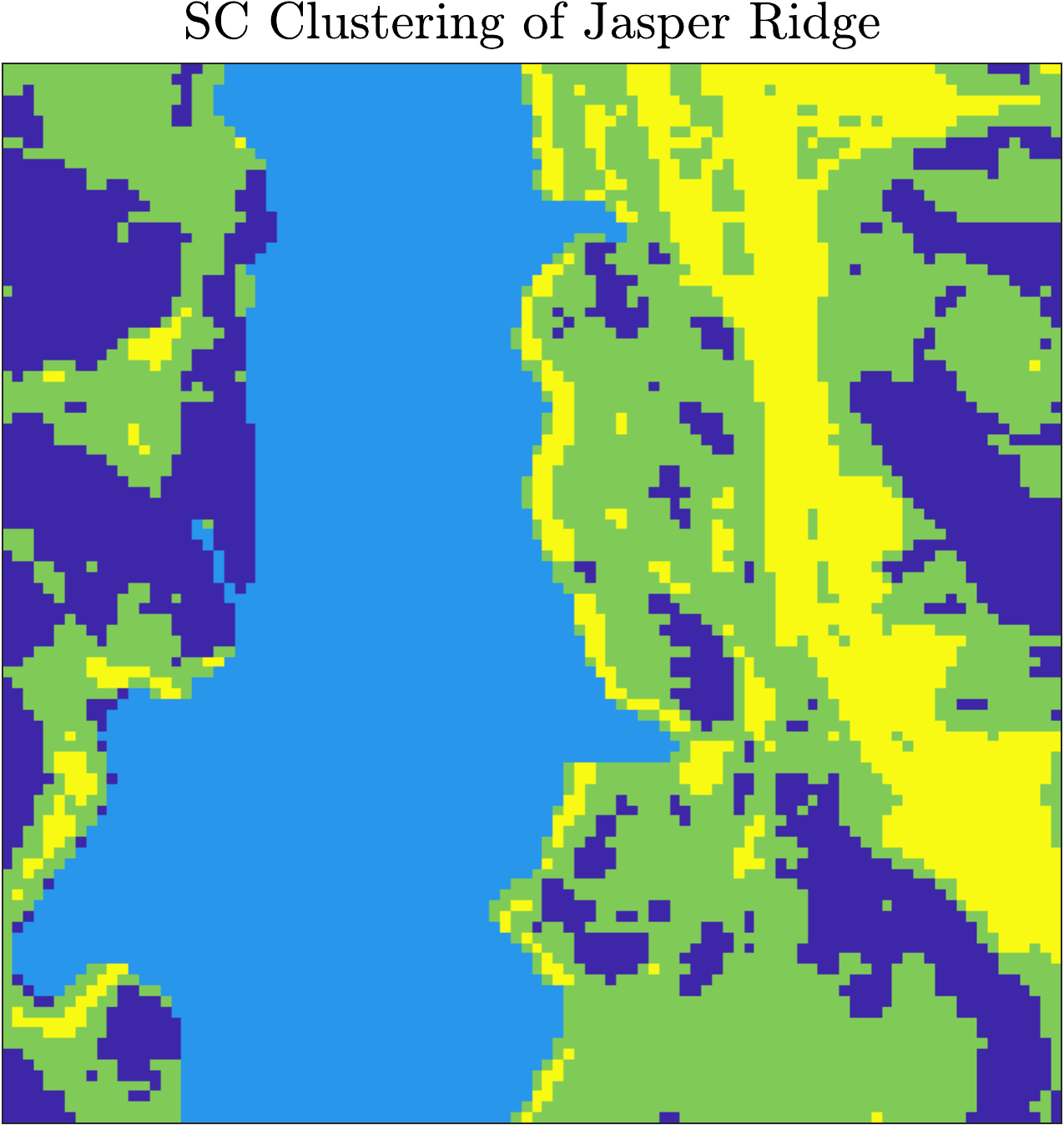}
    \hspace{0.02in}
    \includegraphics[height = 1.45in]{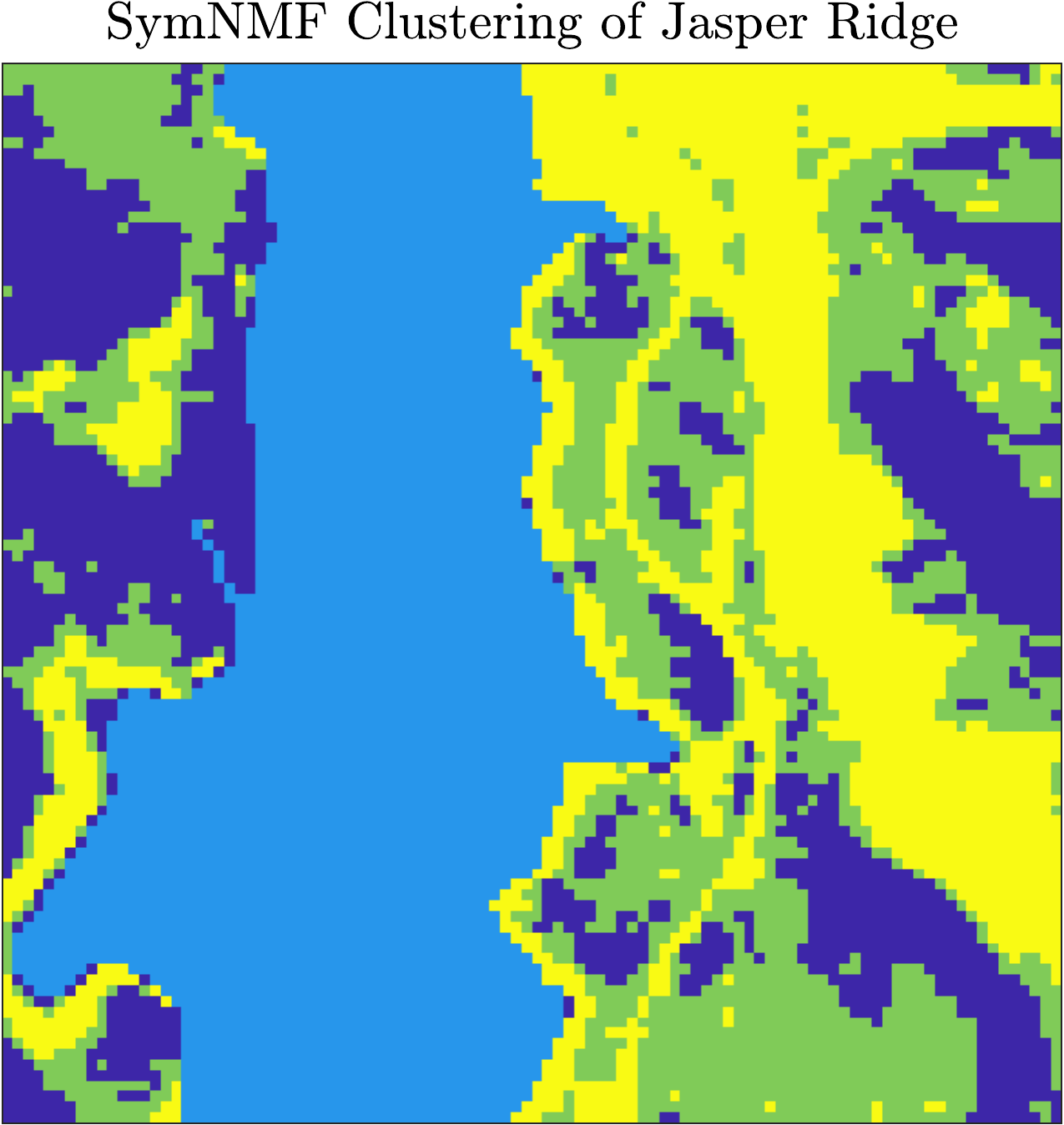}
    \hspace{0.02in}
    \includegraphics[height = 1.45in]{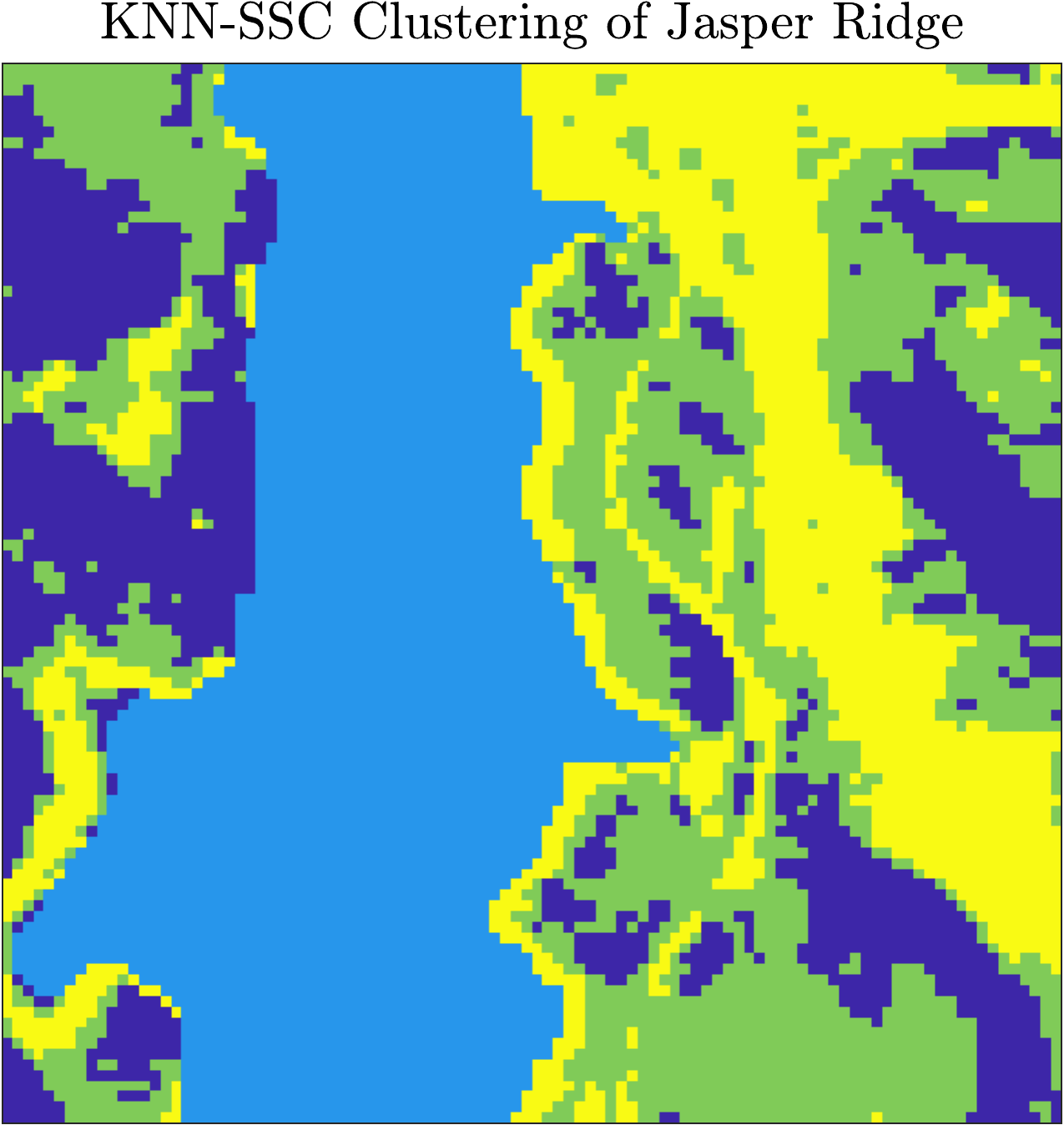}
    
    \vspace{0.1in}
    \includegraphics[height = 1.45in]{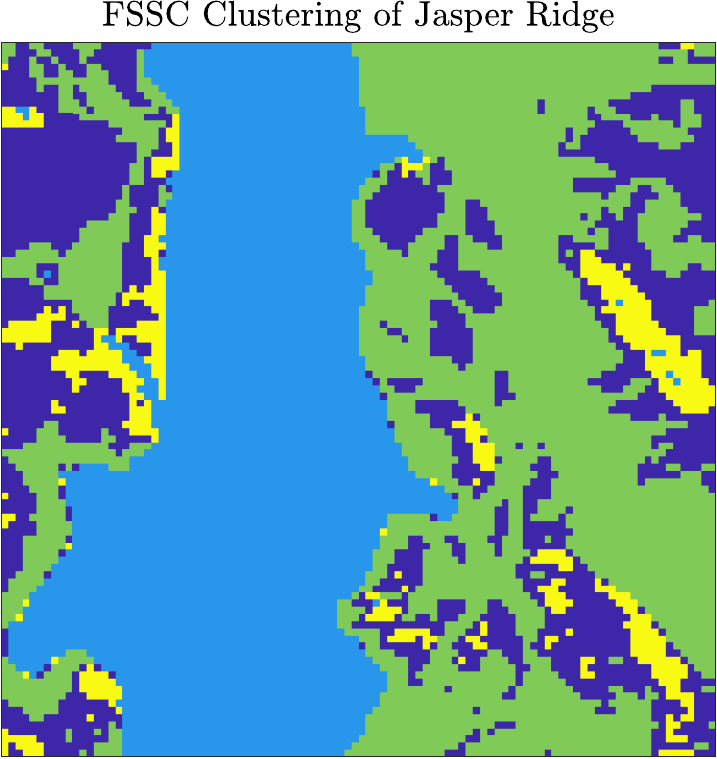}
    \hspace{0.02in}
    \includegraphics[height = 1.45in]{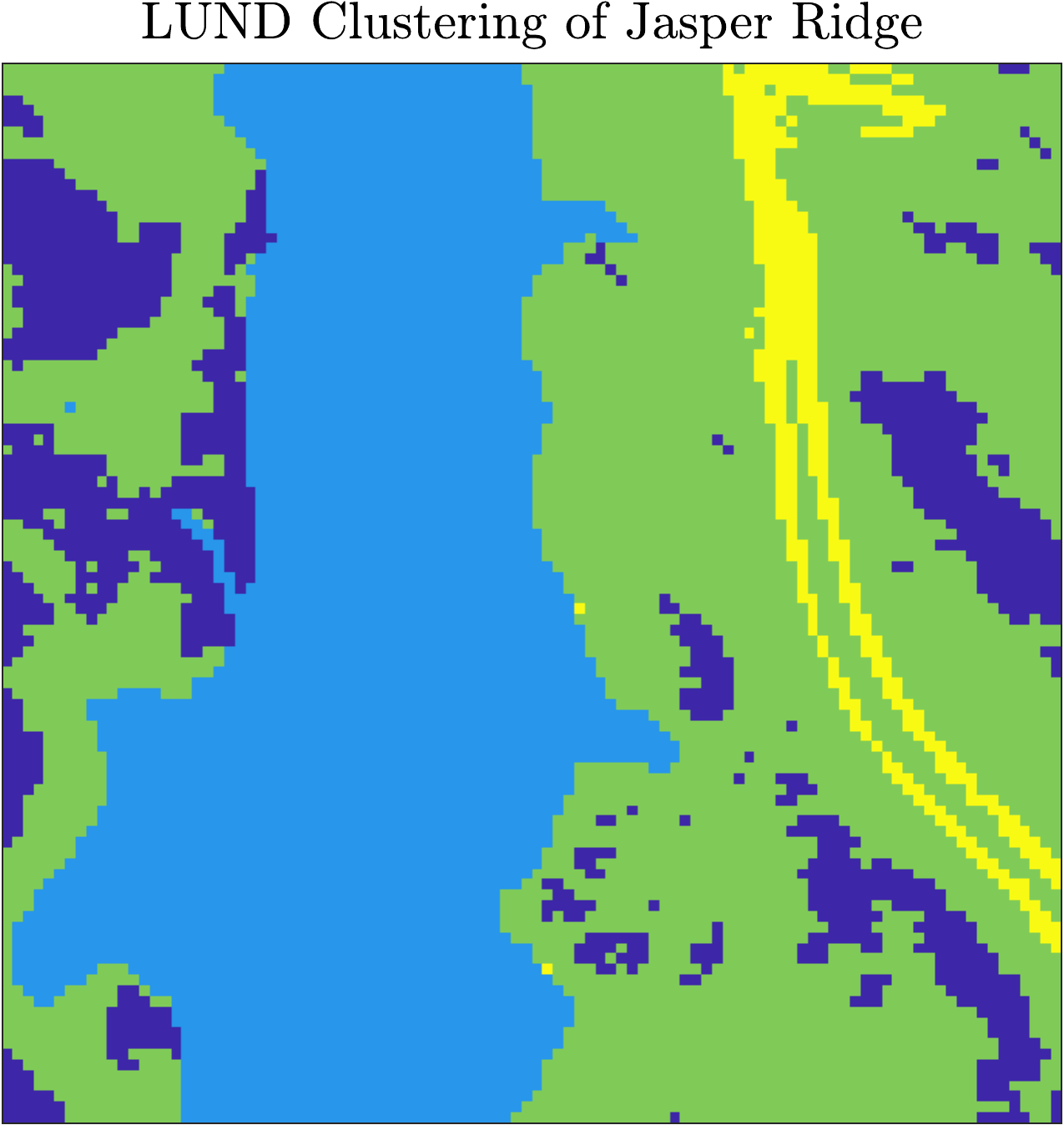}
    \hspace{0.02in}
    \includegraphics[height = 1.45in]{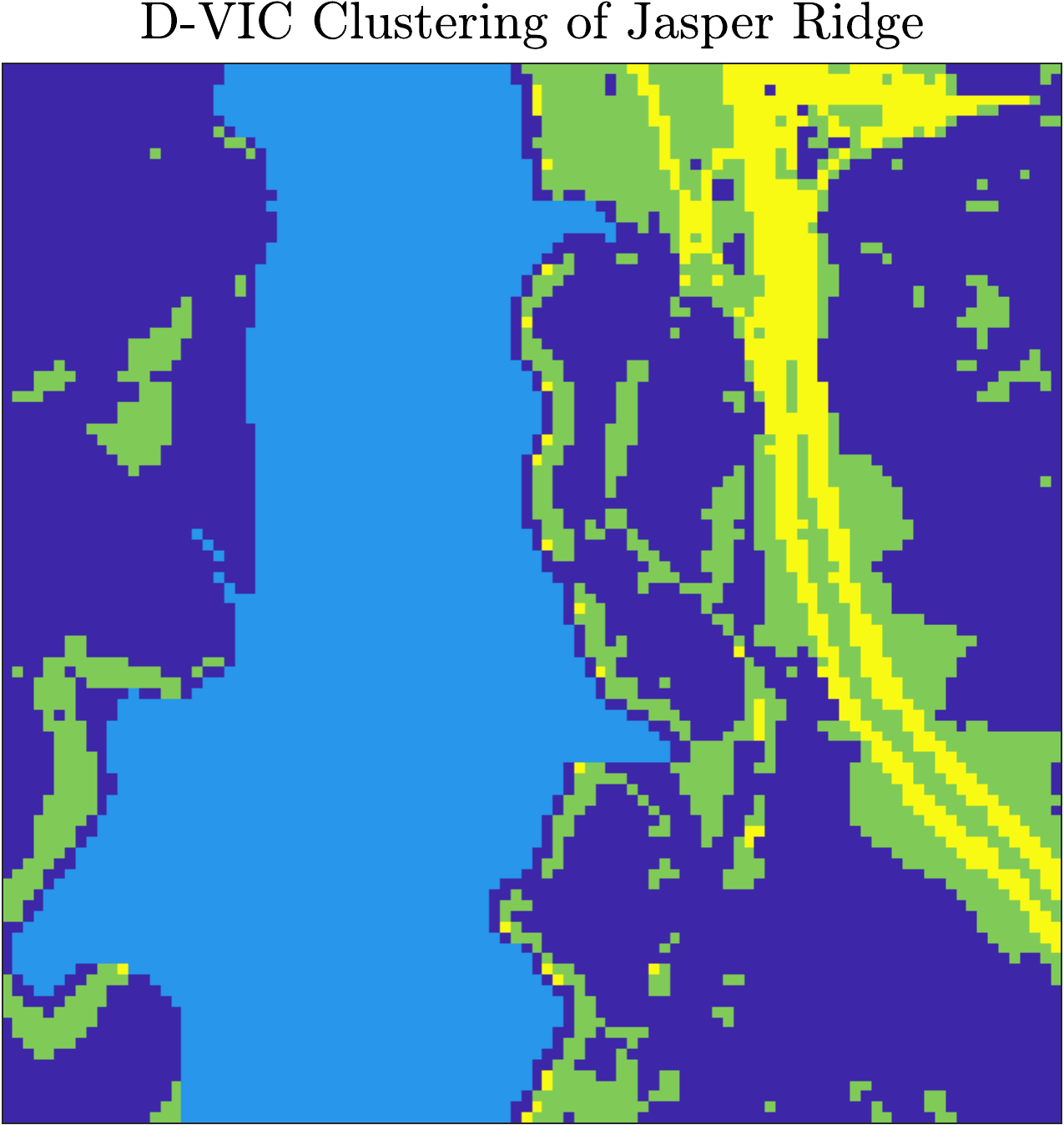}
    
\caption{Comparison of clusterings produced by D-VIC and related algorithms on the Jasper Ridge HSI. D-VIC outperforms all other algorithms, largely due to superior performance among pixels corresponding to tree and soil classes (indicated in dark blue and green, respectively).  \label{fig:jasperRidge} } 

\end{figure}

\begin{table}[t]
\small
\caption{Runtimes (seconds) of D-VIC and related algorithms. D-VIC achieves runtimes comparable to state-of-the-art algorithms and scales well to the larger Indian Pines dataset. \label{tab: runtime}}
\begin{tabular}{p{2.5cm}|c|c|c}
\toprule
\textbf{} & Salinas A	& Jasper Ridge	& Indian Pines\\
\midrule
$K$-Means     &     0.04   & 0.10 &   1.04\\
$K$-Means+PCA &     0.10&    0.14  &  0.58 \\
GMM+PCA       &     0.13 &   0.23   & 2.19\\
DPC           &     3.20 &  6.41    &  25.77 \\
SC            &     1.82   & 3.15  & 14.54\\
SymNMF        &     3.50&    4.42 &  48.29\\
KNN-SSC       &     4.11 &   7.91  & 103.05\\
FSSC          &     13.53 &  30.40  & 130.72\\
LUND          &     2.35  &  4.14 & 14.74\\
D-VIC         &     4.95   & 7.64  & 23.70\\
\bottomrule
\end{tabular}
\end{table}

As visualized in Fig. \ref{fig:SalinasA}, D-VIC achieved nearly perfect recovery of the ground truth labels for Salinas A. Most notably, though all comparison methods erroneously separate the ground truth cluster indicated in yellow in Fig. \ref{fig:salinasAGT} (corresponding to 8-week maturity romaine), D-VIC correctly groups the pixels in this cluster, resulting in performance that was 0.089 higher in OA and 0.110 in $\kappa$ than the that of LUND, its closest competitor in Table \ref{tab: performance}. As such, downweighting high-density points that are not also exemplary of the latent material structure improves not only modal, but also non-modal labeling. Moreover, what error does exist in the D-VIC clustering of Salinas A could likely be remedied through spatial regularization or smoothing post-processing~\cite{murphy2019spectral, murphy2020spatially}. 

D-VIC similarly achieved much higher performance than related state-of-the-art graph-based clustering algorithms on Jasper Ridge (as visualized in Fig. \ref{fig:jasperRidge}). This difference in performance was substantially driven by superior separation of the classes indicated in dark blue (corresponding to tree cover) and green (corresponding to soil) in Fig. \ref{fig:jasperGT}. Indeed, though LUND groups most tree cover pixels with soil pixels in Fig. \ref{fig:jasperRidge}, D-VIC correctly separates much of the latent structure for this class. The difference between LUND's and D-VIC's clusterings indicates that the pixels corresponding to the tree cover class, though lower density than pixels corresponding to the soil class, have relatively high pixel purity.

\subsubsection{Runtime Analysis} \label{sec: runtime}
\noindent
This section compares runtimes of the algorithms implemented in Section \ref{sec: performance}, where hyperparameters were set to be those which produced the results in Table \ref{tab: performance}. All experiments were run in MATLAB on the same environment: a macOS Big Sur system with an 8-core Apple\textsuperscript{\textregistered} M1\textsuperscript{\texttrademark} Processor and 8 GB of RAM. Each core had a processor base frequency of 3.20 GHz. Runtimes are provided in Table \ref{tab: runtime}. All classical algorithms have smaller runtimes than D-VIC, but the performances reported in Table \ref{tab: performance} for these algorithms are substantially less than those reported for D-VIC. On the other hand, though KNN-SSC and SymNMF achieve performances competitive to D-VIC, unlike D-VIC, these algorithms appear to scale poorly to large datasets. DPC, which relies on Euclidean distances between high-dimensional pixel spectra, has lower runtimes on Salinas A than D-VIC but scales poorly to the larger Indian Pines image. In addition, D-VIC outperforms FSSC and operates at lower runtimes across HSI datasets. Finally, D-VIC outperforms LUND at the cost of only a small increase in runtime (associated with the spectral unmixing step).

\begin{figure}[b]
    \centering
    \includegraphics[height = 2in]{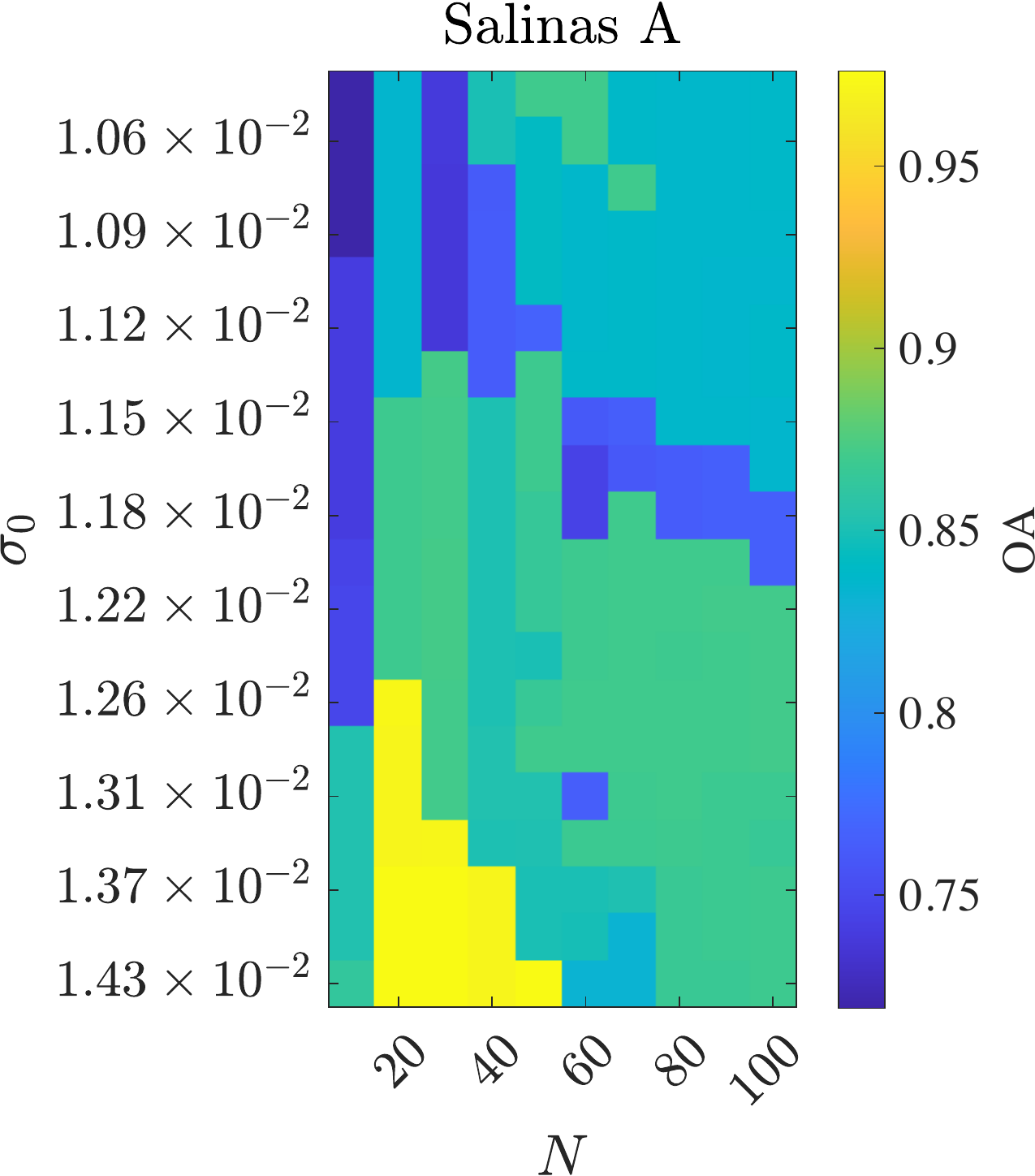}
    \includegraphics[height = 2in]{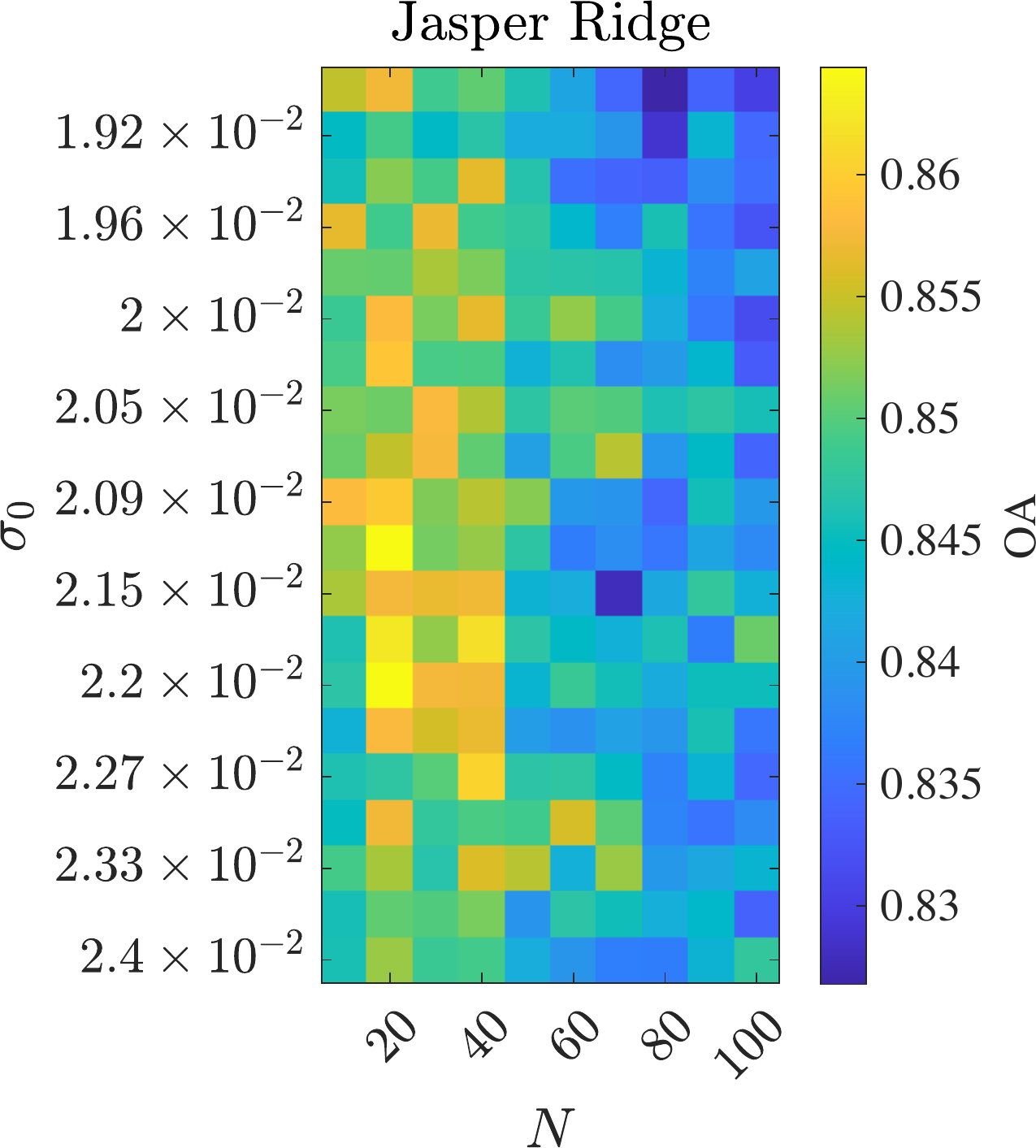} 
    \includegraphics[height = 2in]{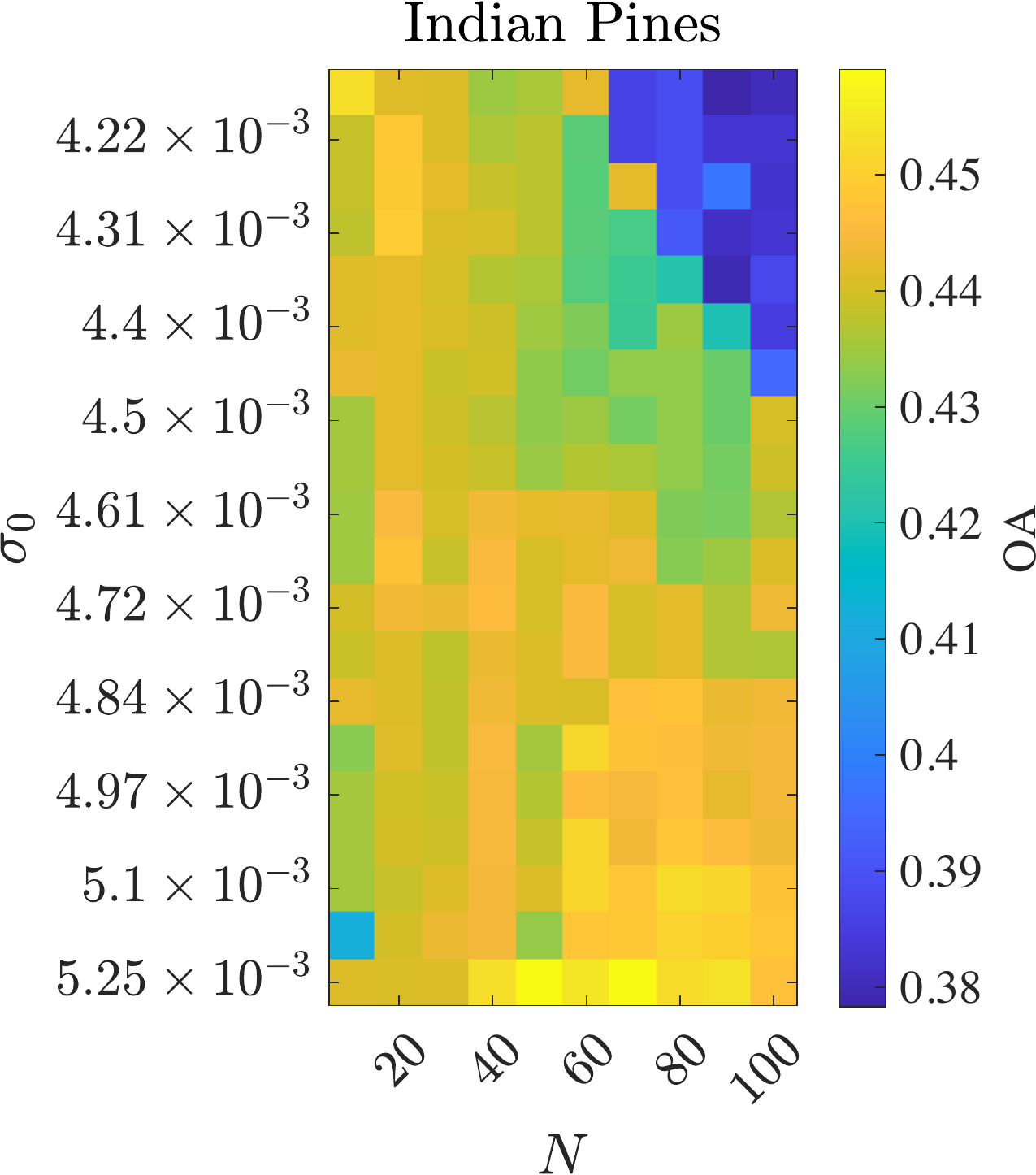} 
    \caption{Visualization of D-VIC's median OA across 50 trials as hyperparameters $N$ and $\sigma_0$ are varied. D-VIC achieves high performance across a large set of hyperparameters. 
    }
    \label{fig: hyperparameters}
\end{figure}

\subsubsection{Robustness to Hyperparameter Selection}\label{sec: HyperParameterRobustness}
\noindent
This section analyzes the robustness of D-VIC's performance to hyperparameter selection. For each node in a grid of $(N,\sigma_0)$, D-VIC was implemented 50 times, and the median OA value across these 50 trials was stored. Performance degraded as $N$ increased substantially past 100, and such a choice is not advised. In Fig. \ref{fig: hyperparameters}, we visualize how the performance of D-VIC varies with $N$ and $\sigma_0$. The relatively small range in nominal values of $\sigma_0$ in our grid reflects that pixels from the HSIs analyzed in this article are relatively close to their $\ell^2$-nearest neighbors on average. As is described in Appendix \ref{app: grid search}---where our hyperparameter optimization is discussed in greater detail---the range of $\sigma_0$ used for each grid search is data-dependent, ranging the distribution of $\ell^2$-distances between pixel spectra and their 1000 $\ell^2$-nearest neighbors.

It is clear from Fig. \ref{fig: hyperparameters} that D-VIC is capable of achieving high performance across a broad range of hyperparameters on each HSI. Thus, given little hyperparameter tuning, D-VIC is likely to output a partition that is competitive with clusterings reported in Table \ref{tab: performance}. Fig. \ref{fig: hyperparameters} also motivates recommendations for hyperparameter selection to optimize the OA of D-VIC.  Larger datasets (e.g., Indian Pines) tend to require larger values of $N$ for D-VIC to achieve high OA, corresponding with recommendations in the literature that $N$ should grow logarithmically with $n$~\cite{maggioni2019learning}.  Additionally, D-VIC achieves the highest OA for datasets with high-purity material classes (e.g., Salinas A) using large $\sigma_0$. This reflects that, as $\sigma_0$ increases, the KDE $p(x)$ becomes more constant across the HSI and $\zeta(x) \approx \eta(x)$. Since purity is an excellent indicator of material class structure for Salinas A, D-VIC becomes better able to recover the latent material structure with larger $\sigma_0$.

\begin{figure}[t]
    \centering
    \includegraphics[width = 0.6\textwidth]{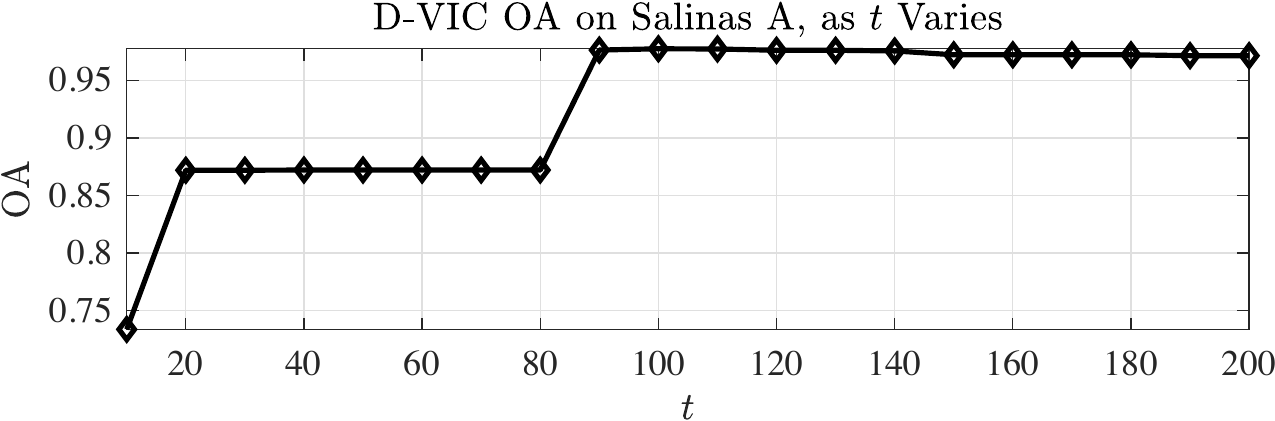}
    
    \includegraphics[width = 0.6\textwidth]{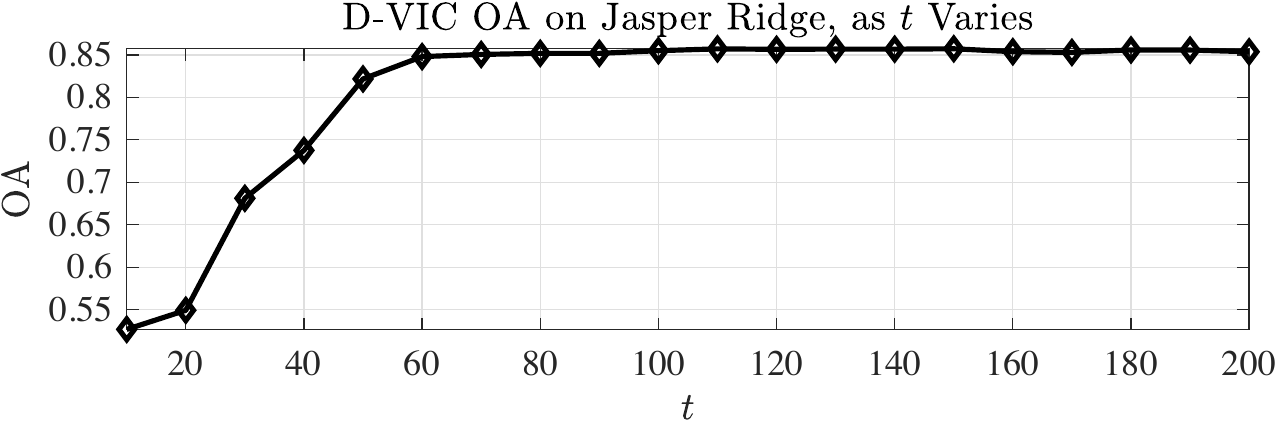}

    \includegraphics[width = 0.6\textwidth]{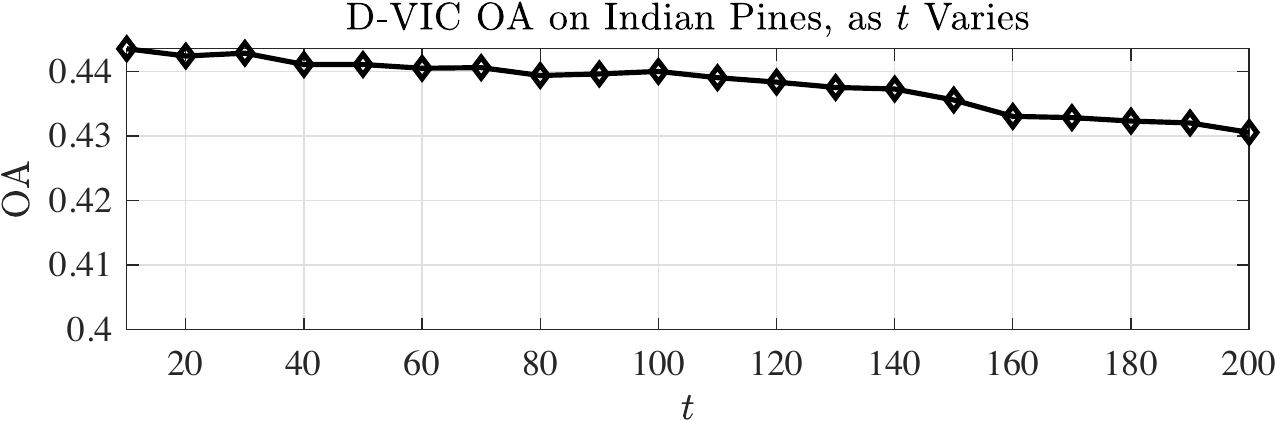}
    
    \caption{Analysis of D-VIC's performance as $t$ varies across $[10,200]$. Values are the median OA across 100 implementations of D-VIC with the optimal $N$ and $\sigma_0$ values. Generally, $t$ appears to have little impact on the OA of D-VIC, and D-VIC achieves performances comparable to those reported in Table \ref{tab: performance} across $t\in [90,200]$, uniformly across all data sets. }
    \label{fig:tAnalysis}
\end{figure}

We also analyze the robustness of D-VIC's performance to the selection of the diffusion time parameter $t$. Using the optimal values of $N$ and $\sigma_0$, D-VIC was evaluated 100 times at $t$-values ranging $\{10,20,\dots, 200\}$. Fig. \ref{fig:tAnalysis}, which visualizes the results of this analysis, indicates D-VIC achieves high OA values across a broad range of $t$; for each $t\in [90,200]$, D-VIC outputs a clustering with OA equal to or very close to those reported in Table \ref{tab: performance}. These results indicate that D-VIC is well-equipped to provide high-quality clusterings given little or no tuning of $t$. Indeed, a simple choice of $t=100$ works exceptionally well across all datasets.

\subsection{Analysis of the Madingley HSI Dataset}\label{sec: Madingley Data Analysis}

This section presents implementations of D-VIC and other clustering algorithms on real HSI data to illustrate that unsupervised clustering algorithms may be used to generate ash dieback disease mappings from remotely-sensed HSI data, even when no ground truth labels are available. Algorithms were evaluated on the Madingley HSI dataset, which was collected by a manned aircraft in August 2018 over a $512~\text{m} \times 356 ~\text{m}$ region of temperate deciduous forest in Madingley Village near Cambridge, United Kingdom~\cite{chan2021monitoring}. This HSI was recorded by a Norsk Elektro Optikk hyperspectral camera (Hyspex VNIR 1800) at a high spatial resolution of 0.32~m. Spectral signatures, ranging in recorded wavelength from 410~nm to 1001~nm across 186 spectral bands, were recorded across $1601\times 1113$ pixels ($n=1816835$). The Madingley HSI was preprocessed using QUick Atmospheric Correction~\cite{eismann2012hyperspectral} (to remove atmospheric effects on pixel spectra) and standardization of spectral signatures (to mitigate differences in illumination across pixels)~\cite{chan2021monitoring}. 

\begin{figure}[t]
 \captionsetup[subfigure]{justification=centering}
    \centering
    
    \begin{subfigure}[t]{0.42\textwidth}
        \centering
        \includegraphics[height = 2.1in]{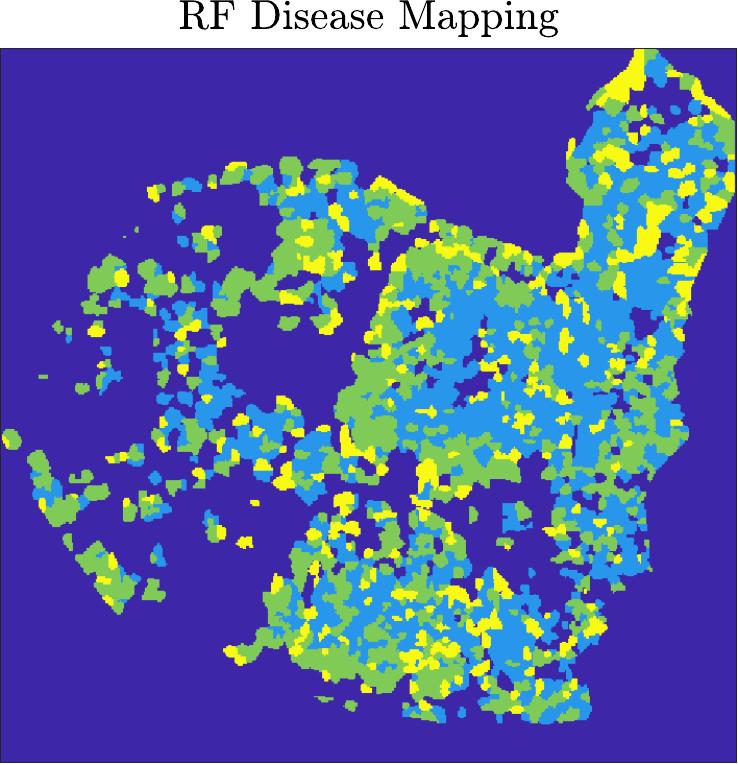}
            \caption{RF Disease Mapping}
        \label{fig:RF}
    \end{subfigure}
    \begin{subfigure}[t]{0.42\textwidth}
        \centering
        \includegraphics[height = 2.1in]{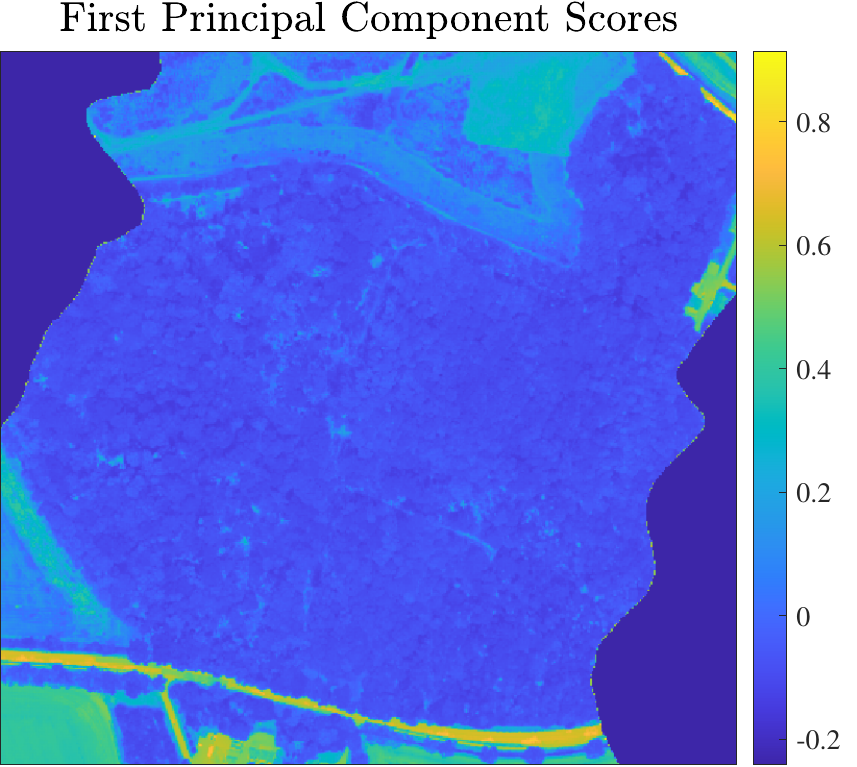}
            \caption{Principal Component Scores}
        \label{fig:MadingleyPCA}
    \end{subfigure}

    \caption{Visualizations of the Madingley HSI.     The RF disease mapping is visualized in  (\ref{fig:RF}) and the Madingley HSI's first principal component scores are visualized in  (\ref{fig:MadingleyPCA}). In (\ref{fig:RF}), yellow indicates severely-infected ash, green indicates infected ash, and light blue indicates healthy ash. 
    }
    \label{fig: Madingley GT}
\end{figure}

Healthy and dieback-infected ash trees were identified in the Madingley scene using a pair of supervised classifiers~\cite{chan2021monitoring}. First, to isolate ash tree crowns in the scene, a supervised Partial Least Squares Discriminant Analysis (PLSDA) classifier was trained to predict tree species using manually-collected ground truth labels for 166 tree crowns in the Madingley scene and 256 tree crowns in three other forest regions near Cambridge~\cite{chan2021monitoring}. Labeled tree crowns were split into training (70\%) and validation (30\%) sets. The trained PLSDA classifier generalized well to the validation set, achieving an OA of 85.3\% on those data~\cite{chan2021monitoring}. Next, a supervised ash dieback disease map was generated for trees in the Madingley scene classified as ash by the PLSDA~\cite{chan2021monitoring}. Specifically, a supervised random forest (RF) classifier was trained to classify a tree crown as one of three disease classes---healthy, infected, and severely infected---using the average pixel spectra from pixels corresponding to that tree crown. The RF was trained using manually-labeled tree crowns across the four aforementioned scenes and evaluated on a validation set consisting of 16 tree crowns from each disease class. The trained RF classifier was highly successful at identifying ash dieback disease, with an OA of 77.1\% on its validation set~\cite{chan2021monitoring}. Visualizations of the Madingley HSI and the RF disease mapping are provided in Fig. \ref{fig: Madingley GT}.

Dieback-infected ash trees tend to have a mosaic of healthy and dead branches, so bicubic interpolation~\cite{keys1981bicubic} was implemented on the Madingley HSI before cluster analysis to downsample pixels to a 1.28~m spatial resolution~\cite{keys1981bicubic}. Thus, each pixel covered a spatial region containing multiple branches, leading to a more holistic characterization of tree health (rather than of individual branches)~\cite{polk2022unsupervised}. Unsupervised clustering algorithms were evaluated on the $n=72775$ pixels in the resulting $401\times 279$ scene that corresponded to ash trees in the down-sampled PLSDA species mapping~\cite{chan2021monitoring}. For each clustering algorithm, we set $K=2$ so that clusters of pixels corresponded to healthy and dieback-infected trees. Unsupervised clusterings were evaluated by comparing against the supervised RF disease mapping after combining the ``infected'' and ``severely infected'' classes and aligning labels using the Hungarian algorithm.

\subsubsection{Discussion of Madingley Experiments}
Table \ref{tab: Madingley performance} summarizes the overlap of D-VIC's and other algorithms' clusterings of the Madingley HSI with the RF disease mapping. Notably, four algorithms---SC, KNN-SSC, LUND, and D-VIC---achieved comparably high OA and $\kappa$ values. We remark that the RF disease mapping used for validation is the result of a supervised learning algorithm trained on a small set of labels.  Because it is an imperfect labeling of the Madingley HSI, small differences in OA or $\kappa$ values between SC, KNN-SSC, LUND, and D-VIC should not be taken as an indication that one of these clustering methods is better or worse than another. Nevertheless, these algorithms' high levels of overlap with the RF disease mapping indicate that graph-based unsupervised clustering algorithms like D-VIC may be applied to remotely-sensed HSI data to assess forest health even when ground truth labels are unavailable.

\begin{figure}[t]
\centering
    \centering
    \includegraphics[height = 1.45in]{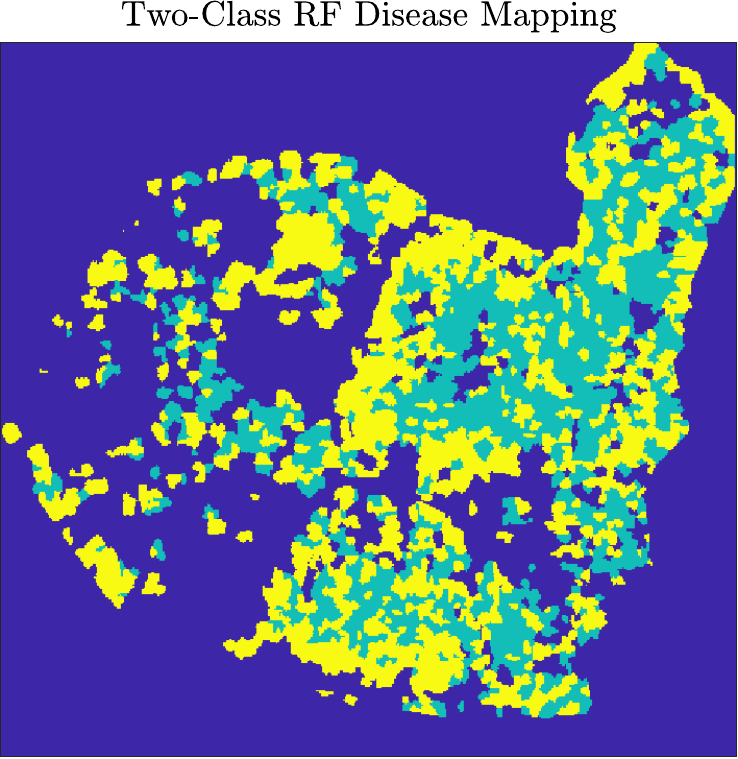}
    \hspace{0.02in}
    \includegraphics[height = 1.45in]{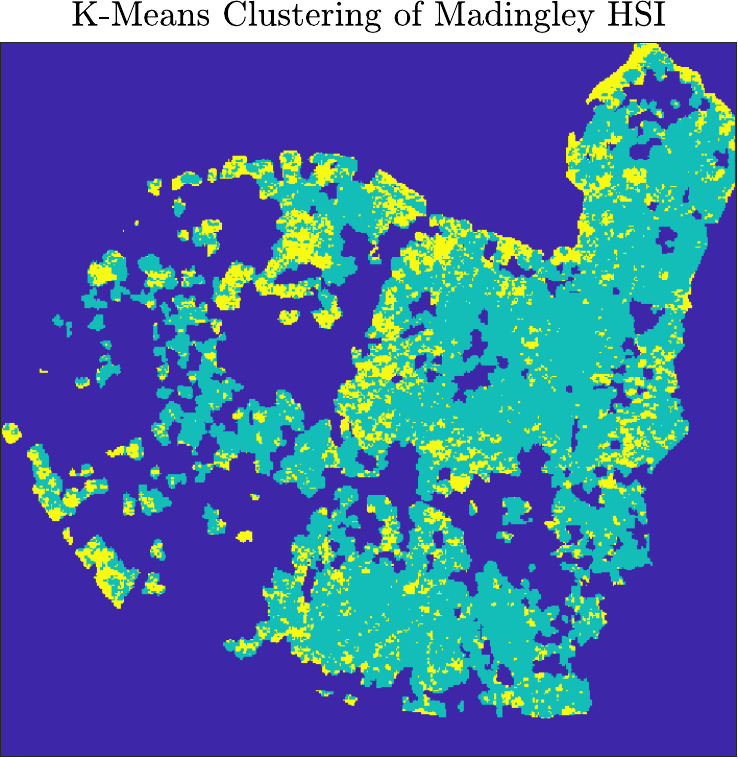}
    \hspace{0.02in}
    \includegraphics[height = 1.45in]{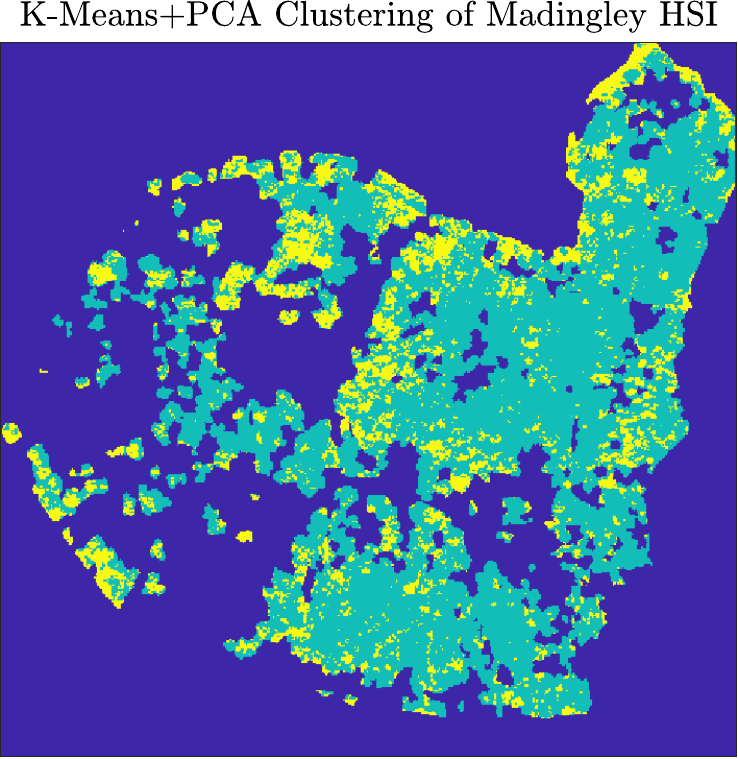}
    \hspace{0.02in}
    \includegraphics[height = 1.45in]{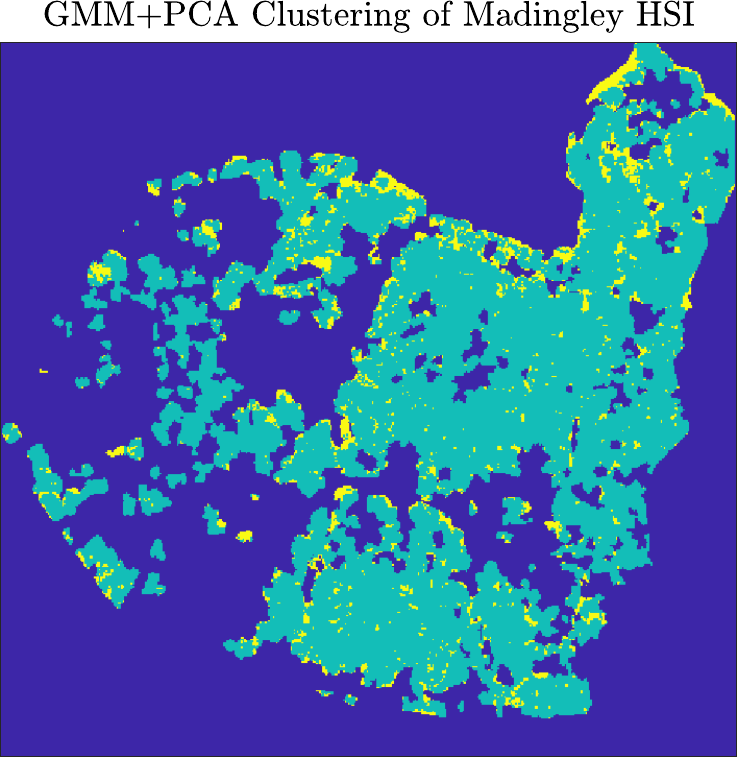}
    
    \vspace{0.1in}
    \includegraphics[height = 1.45in]{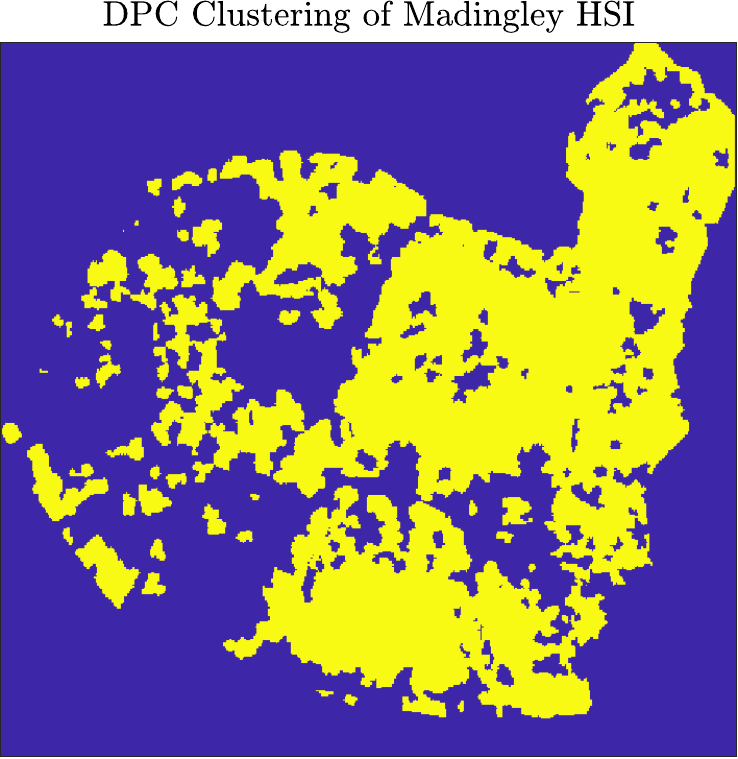}
    \hspace{0.02in}
    \includegraphics[height = 1.45in]{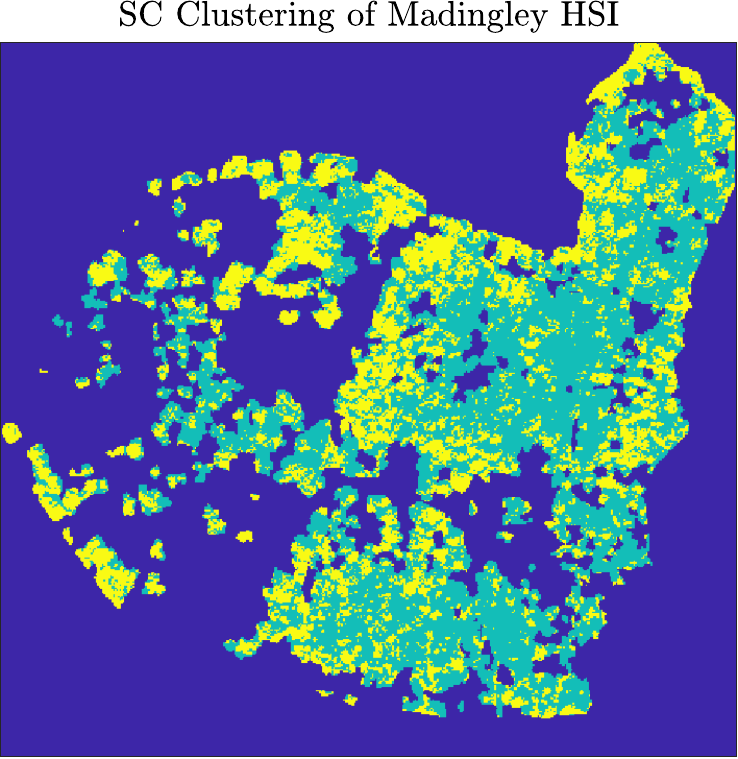}
    \hspace{0.02in}
    \includegraphics[height = 1.45in]{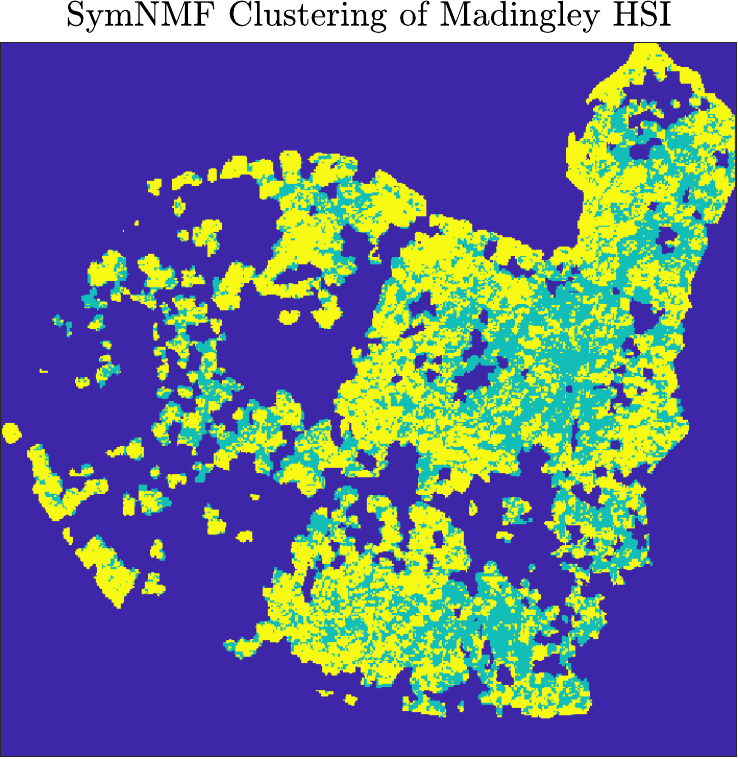}
    \hspace{0.02in}
    \includegraphics[height = 1.45in]{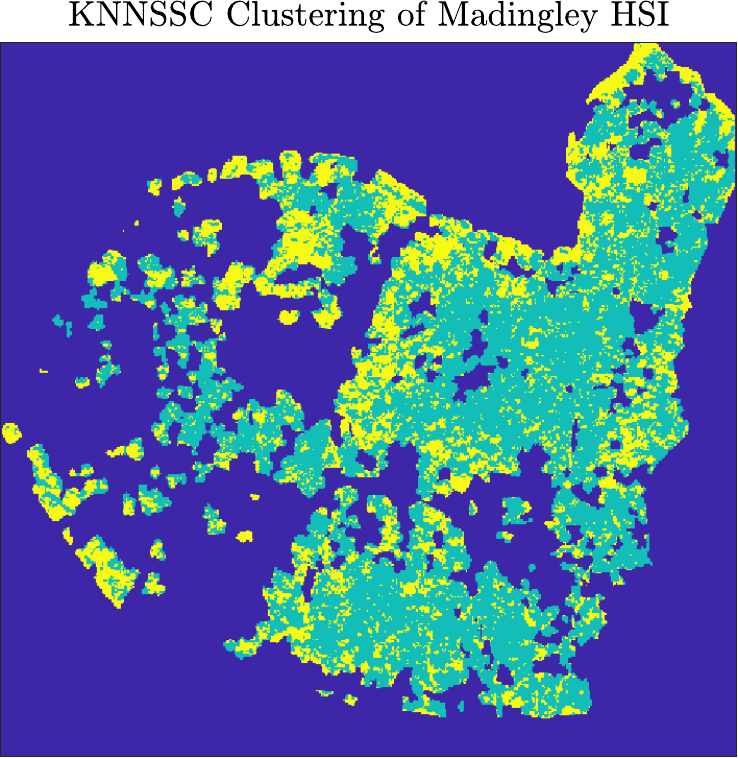}

    \vspace{0.1in}
    \includegraphics[height = 1.45in]{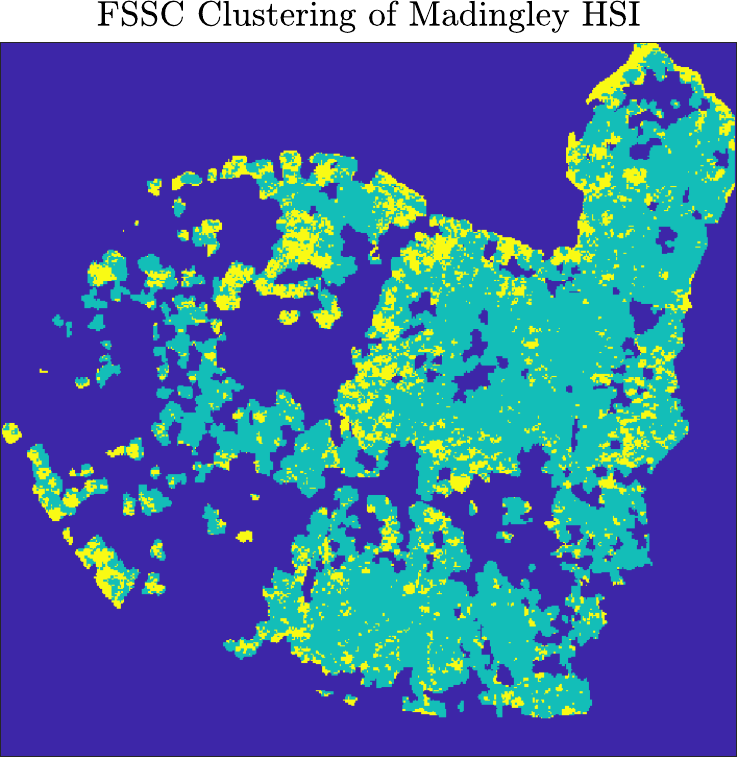}
    \hspace{0.02in}
    \includegraphics[height = 1.45in]{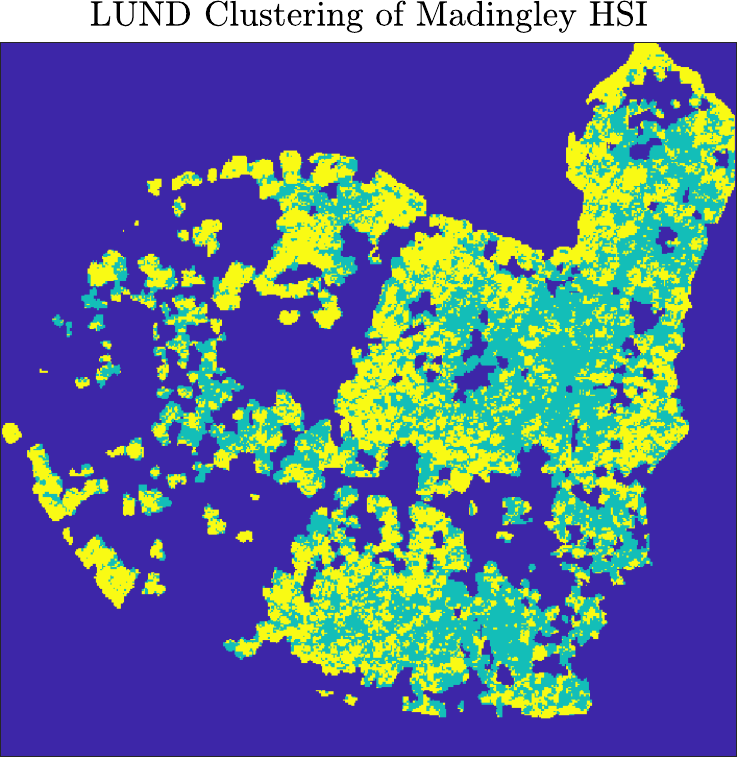} \hspace{0.02in}
    \includegraphics[height = 1.45in]{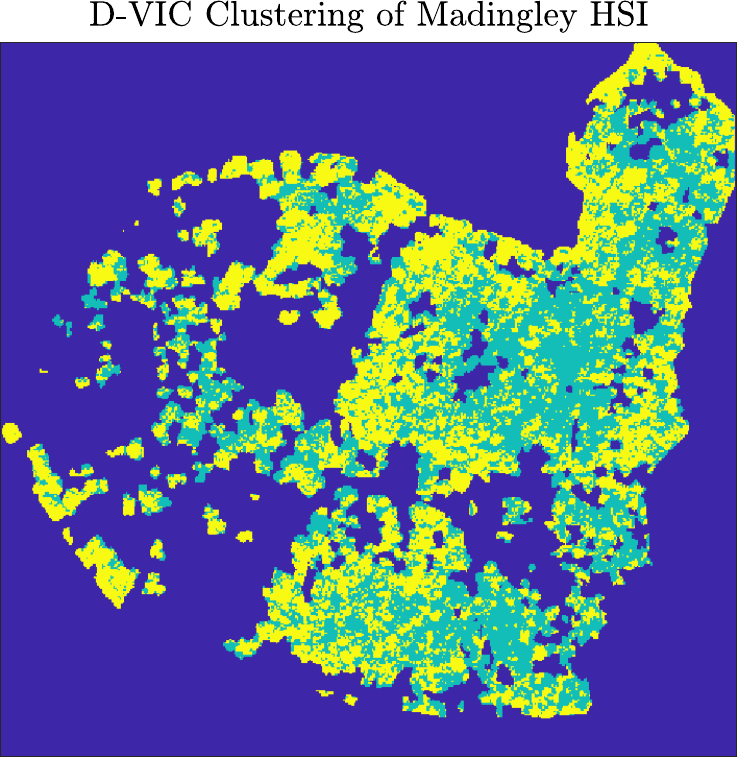}
    
\caption{Comparison of clusterings produced by D-VIC and related algorithms on the Madingley HSI. Labels were aligned so that yellow indicates dieback-infected ash and teal indicates healthy ash. Though the performance of many graph-based algorithms (SC, KNN-SSC, LUND, and D-VIC) was similar in Table \ref{tab: Madingley performance}, qualitative differences exist between these algorithms' clusterings.  \label{fig:Madingley} } 

\end{figure}

\begin{table}[b]
\caption{\label{tab: Madingley performance} 
Performances of D-VIC and related algorithms on the Madingley HSI. Highest and second-highest performances are bolded and underlined, respectively. Many graph-based algorithms---SC, KNN-SSC, LUND, and D-VIC---achieved approximately the same high performance on the Madingley HSI, indicating that graph-based HSI clustering algorithms may be used for unsupervised ash dieback disease mapping, even when no ground truth labels exist. 
} 
\resizebox{1\textwidth}{!}{%
\begin{tabular}{c|c|c|c|c|c|c|c|c|c|c}
\toprule
 & KM & KM+PCA & GMM+PCA & DPC & SC & SymNMF & KNN-SSC & FSSC & LUND & D-VIC \\ 
 \midrule
 OA & 0.570 & 0.570 & 0.477 & 0.555  & 0.595	& 0.630 & \textbf{0.651} & 0.608 & \underline{0.648} & 0.645\\
 $\kappa$ & 0.245 & 0.245 & 0.099 &   0.000  & \underline{0.300} & 0.243 & \textbf{0.328} & 0.262 & 0.296 & 0.287\\ \bottomrule
\end{tabular}%
} 
\end{table}

Though many graph-based HSI clustering algorithms exhibited similar levels of overlap with the supervised RF disease mapping, substantial differences exist between the unsupervised disease mappings obtained by different clustering algorithms, as can be observed in Fig. \ref{fig:Madingley}. Indeed, LUND and D-VIC tended to predict ash dieback disease in regions considered healthy according to the RF disease map~\cite{chan2021monitoring}. On the other hand, other similarly-performing graph-based clustering algorithms (SC and KNN-SSC) tended to label trees as healthy even in regions where the RF disease map indicates substantial dieback disease infection~\cite{chan2021monitoring}. All clusterings exhibit salt-and-pepper error, indicating that spatial regularization~\cite{murphy2019spectral, murphy2020spatially} or majority voting within tree crowns~\cite{chan2021monitoring, polk2022unsupervised} may improve overlap between unsupervised clusterings and the RF labeling even further.

\section{Conclusion} \label{sec: conclusion}
\noindent
This article introduces the \emph{D-VIC clustering algorithm} for unsupervised material classification in HSIs. D-VIC assigns modal labels to high-density, high-purity pixels within the HSI that are far in diffusion distance from other high-density, high-purity pixels~\cite{chan2011simplex, maggioni2019learning, cui2021unsupervised}. We have argued that these \emph{cluster modes} are highly indicative of underlying material structure, leading to more interpretable and accurate clusterings than those produced by related algorithms~\cite{murphy2018unsupervised, maggioni2019learning}. Indeed, experiments presented in Section \ref{sec: numerical experiments} show that incorporating pixel purity into D-VIC results in clusterings closer to the ground truth labels on three benchmark real HSI datasets of varying sizes and complexities and enables high-fidelity unsupervised ash dieback disease detection on remotely-sensed HSI data. As such, D-VIC is equipped to perform efficient material clustering on broad ranges of spectrally mixed HSI datasets.

Future work includes modifying the spectral unmixing step in D-VIC. To demonstrate the effect of including a pixel purity estimate in a diffusion-based HSI clustering algorithm, we have chosen a simple, standard linear unmixing procedure to generate the pixel purity estimate in D-VIC: using HySime to estimate the number of endmembers in an HSI~\cite{bioucas2008HySime}, AVMAX to estimate those endmembers~\cite{chan2011simplex}, and a nonlinear least squares solver~\cite{bro1997fast} to calculate abundances. The spectral unmixing procedure in D-VIC is quite modular, however, and improvements to D-VIC's clustering performance may be gained through improvements to this procedure; for example, by explicitly constraining abundances to sum to one~\cite{heinz2001fully} or accounting for nonlinear mixing of endmembers~\cite{chen2012nonlinear, halimi2011nonlinear, yokoya2013nonlinear}. Linear endmember extraction is computationally inexpensive~\cite{bioucas2008HySime, chan2011simplex, bro1997fast}, and it results in strong performance in D-VIC, but recent years have brought significant advances in algorithms for the nonlinear spectral unmixing of HSIs~\cite{chen2012nonlinear, halimi2011nonlinear, yokoya2013nonlinear}. Modifying the spectral unmixing step in D-VIC may improve performance, especially for HSIs in which assumptions on linear mixing do not hold~\cite{chen2012nonlinear, halimi2011nonlinear, yokoya2013nonlinear}. 

Additionally, much of the error in D-VIC's clusterings may be corrected by incorporating spatial information into its labeling. Such a modification of D-VIC may improve performance on datasets with spatially homogeneous clusters~\cite{murphy2019spectral, murphy2020patch, polk2021multiscale, fauvel2012advances,ghamisi2013spectral,fang2015classification, tarabalka2009spectral, mohan2007spatially}. Moreover, it is likely that varying the diffusion time parameter $t$ in D-VIC may enable the detection of multiple scales of latent cluster structure, a problem we would like to consider further in future work~\cite{murphy2021multiscale, polk2021multiscale, polk2022diffusion, gillis2014hierarchical}. Additionally,  we expect D-VIC may be modified for \emph{active learning}, wherein ground truth labels for a small number of carefully selected pixels are queried and propagated across the image~\cite{maggioni2019LAND, murphy2020spatially}.  Finally, we expect that D-VIC (or one of the extensions described above) may be modified for change detection in remotely-sensed scenes~\cite{camalan2022change}.

\section*{Acknowledgments}

The US National Science Foundation partially supported this research through grants NSF-DMS 1912737, NSF-DMS 1924513, and NSF-CCF 1934553. We thank C. Sch\"{o}nlieb and M. S. Kotzagiannidis for conversations that aided in the development of D-VIC. We acknowledge C. Barnes and 2Excel Geo for collecting the Madingley HSI used in this study. We thank the University of Cambridge for access to the Madingley field site and the Wildlife Trust for Bedfordshire, Cambridgeshire \& Nottinghamshire for access to other forest field sites. Finally, we thank N. Gillis, D. Kuang, H. Park, C. Ding, M. Abdolali, D. Kun, I. Gerg, and J. Wang for making code for HySime~\cite{bioucas2008HySime}, AVMAX~\cite{matlab_hsi_toolbox, chan2011simplex}, SymNMF~\cite{kuang2012symmetric}, KNN-SSC~\cite{abdolali2021beyond}, and FSSC~\cite{wang2021fast} publicly available.  We also thank the academic editor and the three reviewers for their helpful comments, which improved the presentation of this paper.

\section*{Data Availability}

Real benchmark hyperspectral image data used in this study can be found at the following links: \url{http://www.ehu.eus/ccwintco/index.php?title=Hyperspectral_Remote_Sensing_Scenes} and \url{https://rslab.ut.ac.ir/data}. Experiments for benchmark data can be replicated at \url{https://github.com/sampolk/D-VIC}. Software and data required to replicate experiments on the Madingley HSI shall be made available upon reasonable request.

\appendix

\section[\appendixname~\thesection]{Hyperparameter Optimization} \label{app: grid search}
\noindent
This appendix describes the hyperparameter optimization performed to generate numerical results. The parameter grids used for each algorithm are summarized in Table \ref{tab:hyperparameters}. For $K$-Means+PCA and GMM+PCA, we clustered the first $z$ principal components of the HSI, where $z$ was chosen so that 99\% of the variation in the HSI was maintained after PCA dimensionality reduction. Thus, $K$-Means, $K$-Means+PCA, and GMM+PCA required no hyperparameter inputs. For stochastic algorithms with hyperparameter inputs (SC, SymNMF, FSSC, and D-VIC), we optimized for the median OA across 100 trials at each node in the hyperparameter grids described below.

\begin{table}[t]
\caption{Hyperparameter grids for algorithms. The number of nearest neighbors $N$ took values in $\mathcal{N}$: an exponential sampling of the set $[10,900]$. The set $\mathscr{A}$ is a grid of values ranging from 0 to 1 used as FSSC regularization parameters. The set $\mathscr{D}$ contains $\ell^2$-distances between data points and their 1000 $\ell^2$-nearest neighbors. The set $\mathscr{T}$ is an exponential sampling of the diffusion process: $\mathscr{T}=\{0,1,2, \dots, 2^2, \dots, 2^T\}$.  A ``---'' indicates a lack of a hyperparameter input. 
    \label{tab:hyperparameters}}
\begin{tabular}{c|c|c|c}
\toprule
\textbf{} & \textbf{Parameter 1 Grid}	& \textbf{Parameter 2 Grid}	& \textbf{Parameter 3 Grid}\\
\midrule
$K$-Means & --- & ---  & ---  \\ 
$K$-Means+PCA                       & --- & ---  & ---   \\
GMM+PCA                             & --- & ---  & ---  \\
DPC~\cite{rodriguez2014clustering}       &  $N\in \mathcal{N}$ & $\sigma_0 \in \mathscr{D}$ & ---  \\
SC~\cite{ng2002spectral}            & $N\in \mathcal{N}$ & ---  & ---  \\
SymNMF~\cite{kuang2012symmetric} & $N\in \mathcal{N}$ & --- & --- \\ 
KNN-SSC~\cite{abdolali2021beyond,zhuang2016locality}            &  $N\in\mathcal{N}$ &  $\lambda=10$   & ---    \\
FSSC~\cite{wang2021fast}            &  $N\in\mathcal{N}$ &  $\alpha_u\in\mathscr{A}$   & $\ell=2^{11}$    \\
LUND~\cite{maggioni2019learning}       &  $N\in \mathcal{N}$ & $\sigma_0 \in \mathscr{D}$ & $t \in \mathscr{T}$  \\
D-VIC &  $N\in \mathcal{N}$ & $\sigma_0 \in \mathscr{D}$   & $t\in \mathscr{T}$ \\
\bottomrule
\end{tabular}
\small
\end{table}

All graph-based algorithms relied on adjacency matrices built from sparse KNN graphs. The number of nearest neighbors was optimized for each algorithm across $\mathcal{N}$: an exponential sampling of the set $[10,900]$. KNN-SSC's regularization parameter was set to $\lambda = 10$, motivated by prior work with this parameter~\cite{abdolali2021beyond}. FSSC was evaluated using regularization parameters $\alpha_u\in\mathscr{A}=\{0, 10^{-5}, 10^{-3}, 10^{-1}, 0.5, 0.99, 0.999, 0.9999\}$, as was suggested in~\cite{wang2021fast}. FSSC, as an anchor-based clustering algorithm, requires the number of anchor pixels $m$ as input. We set $\ell= 2^{11}$, as this $\ell$-value is greater than all $N\in \mathcal{N}$~\cite{wang2021fast}. We used the same KDE and hyperparameter ranges of $\sigma_0$ for DPC, LUND, and D-VIC. In our grid searches, $\sigma_0$ ranged $\mathscr{D}$: a sampling of the distribution of $\ell^2$-distances between data points and their 1000 $\ell^2$-nearest neighbors. Both LUND and D-VIC were implemented at time steps $t\in \mathscr{T}= \{0,1,2,2^2, \dots, 2^T\}$, where  $T = \lceil \log_2 [\log_{\lambda_2(\mathbf{P})}(\frac{2\times 10^{-5}}{\min(\pi)})]\rceil$. Searches end at time $t=2^T$ because $\max_{x,y\in X}D_t(x,y) \leq 10^{-5}$ for $t\geq~2^T$~\cite{polk2021multiscale, murphy2021multiscale}. For each dataset, we chose the $t\in \{0,1,2,2^2, \dots, 2^T\}$ resulting in maximal OA. As is described in Section \ref{sec: HyperParameterRobustness}, D-VIC is quite robust to this choice of parameter.
\bibliography{ref} 

\begin{thebibliography}{157}
\providecommand{\natexlab}[1]{#1}
\providecommand{\url}[1]{\texttt{#1}}
\expandafter\ifx\csname urlstyle\endcsname\relax
  \providecommand{\doi}[1]{doi: #1}\else
  \providecommand{\doi}{doi: \begingroup \urlstyle{rm}\Url}\fi

\bibitem[Eismann(2012)]{eismann2012hyperspectral}
Michael~Theodore Eismann.
\newblock \emph{Hyperspectral remote sensing}.
\newblock SPIE, 2012.

\bibitem[Ghamisi et~al.(2017)Ghamisi, Plaza, Chen, Li, and
  Plaza]{ghamisi2017advanced}
Pedram Ghamisi, Javier Plaza, Yushi Chen, Jun Li, and Antonio~J Plaza.
\newblock Advanced spectral classifiers for hyperspectral images: A review.
\newblock \emph{IEEE Geosci Remote Sens Mag}, 5\penalty0 (1):\penalty0 8--32,
  2017.

\bibitem[Plaza et~al.(2011)Plaza, Mart{\'i}n, Plaza, Zortea, and
  S{\'a}nchez]{plaza2011recent}
Antonio Plaza, Gabriel Mart{\'i}n, Javier Plaza, Maciel Zortea, and Sergio
  S{\'a}nchez.
\newblock Recent developments in endmember extraction and spectral unmixing.
\newblock In \emph{Opt Remote Sens}, pages 235--267. Springer, 2011.

\bibitem[Edelman et~al.(2012)Edelman, Gaston, Van~Leeuwen, Cullen, and
  Aalders]{edelman2012hyperspectral}
Gerda~J Edelman, Edurne Gaston, Ton~G Van~Leeuwen, PJ~Cullen, and Maurice~CG
  Aalders.
\newblock Hyperspectral imaging for non-contact analysis of forensic traces.
\newblock \emph{Forensic Sci Int}, 223\penalty0 (1-3):\penalty0 28--39, 2012.

\bibitem[Adam et~al.(2010)Adam, Mutanga, and Rugege]{adam2010multispectral}
E~Adam, O~Mutanga, and D~Rugege.
\newblock Multispectral and hyperspectral remote sensing for identification and
  mapping of wetland vegetation: a review.
\newblock \emph{Wetl Ecol Manag}, 18\penalty0 (3):\penalty0 281--296, 2010.

\bibitem[Hirano et~al.(2003)Hirano, Madden, and Welch]{hirano2003hyperspectral}
A~Hirano, M~Madden, and R~Welch.
\newblock Hyperspectral image data for mapping wetland vegetation.
\newblock \emph{Wetl}, 23\penalty0 (2):\penalty0 436--448, 2003.

\bibitem[Clevers et~al.(2010)Clevers, Kooistra, and
  Schaepman]{clevers2010estimating}
J~G P~W Clevers, L~Kooistra, and M~E Schaepman.
\newblock Estimating canopy water content using hyperspectral remote sensing
  data.
\newblock \emph{Int J Appl Earth Obs Geoinf}, 12\penalty0 (2):\penalty0
  119--125, 2010.

\bibitem[Dalponte et~al.(2008)Dalponte, Bruzzone, and
  Gianelle]{dalponte2008fusion}
M~Dalponte, L~Bruzzone, and D~Gianelle.
\newblock Fusion of hyperspectral and lidar remote sensing data for
  classification of complex forest areas.
\newblock \emph{IEEE Trans Geosci Remote Sens}, 46\penalty0 (5):\penalty0
  1416--1427, 2008.

\bibitem[Wang et~al.(2022)Wang, Guan, Zhang, Lee, Margenot, Ge, Peng, Zhou,
  Zhou, and Huang]{wang2022using}
Sheng Wang, Kaiyu Guan, Chenhui Zhang, DoKyoung Lee, Andrew~J Margenot, Yufeng
  Ge, Jian Peng, Wang Zhou, Qu~Zhou, and Yizhi Huang.
\newblock Using soil library hyperspectral reflectance and machine learning to
  predict soil organic carbon: Assessing potential of airborne and spaceborne
  optical soil sensing.
\newblock \emph{Remote Sens Env}, 271:\penalty0 112914, 2022.

\bibitem[Jia et~al.(2020)Jia, Wang, Chen, Guo, Shu, and Wang]{jia2020status}
Jianxin Jia, Yueming Wang, Jinsong Chen, Ran Guo, Rong Shu, and Jianyu Wang.
\newblock Status and application of advanced airborne hyperspectral imaging
  technology: A review.
\newblock \emph{Infr Phys Technol}, 104:\penalty0 103115, 2020.

\bibitem[Price(1997)]{price1997spectral}
John~C Price.
\newblock Spectral band selection for visible-near infrared remote sensing:
  spectral-spatial resolution tradeoffs.
\newblock \emph{IEEE Trans Geosci Remote Sens}, 35\penalty0 (5):\penalty0
  1277--1285, 1997.

\bibitem[Bioucas-Dias et~al.(2013)Bioucas-Dias, Plaza, Camps-Valls, Scheunders,
  Nasrabadi, and Chanussot]{bioucas2013hyperspectral}
Jos{\'e}~M Bioucas-Dias, Antonio Plaza, Gustavo Camps-Valls, Paul Scheunders,
  Nasser Nasrabadi, and Jocelyn Chanussot.
\newblock Hyperspectral remote sensing data analysis and future challenges.
\newblock \emph{IEEE Geosci Remote Sens Mag}, 1\penalty0 (2):\penalty0 6--36,
  2013.

\bibitem[Laparrcr and Santos-Rodriguez(2015)]{laparrcr2015spatial}
Valero Laparrcr and Ra{\'u}l Santos-Rodriguez.
\newblock Spatial/spectral information trade-off in hyperspectral images.
\newblock In \emph{International Geosci Remote Sens Symposium}, pages
  1124--1127. IEEE, 2015.

\bibitem[Miao and Qi(2007)]{miao2007endmember}
L.~Miao and H.~Qi.
\newblock Endmember extraction from highly mixed data using minimum volume
  constrained nonnegative matrix factorization.
\newblock \emph{IEEE Trans Geosci Remote Sens}, 45\penalty0 (3):\penalty0
  765--777, 2007.

\bibitem[Pacheco-Labrador et~al.(2022)Pacheco-Labrador, Migliavacca, Ma,
  Mahecha, Carvalhais, Weber, Benavides, Bouriaud, Barnoaiea, and
  Coomes]{pacheco2022challenging}
Javier Pacheco-Labrador, Mirco Migliavacca, Xuanlong Ma, Miguel Mahecha, Nuno
  Carvalhais, Ulrich Weber, Raquel Benavides, Olivier Bouriaud, Ionut
  Barnoaiea, and David~A Coomes.
\newblock Challenging the link between functional and spectral diversity with
  radiative transfer modeling and data.
\newblock \emph{Remote Sens Env}, 280:\penalty0 113170, 2022.

\bibitem[Jia et~al.(2017)Jia, Wang, Zhuang, Yao, Wang, Zhao, Shu, and
  Wang]{jia2017high}
J~Jia, Y~Wang, X~Zhuang, Y~Yao, S~Wang, D~Zhao, R~Shu, and J~Wang.
\newblock High spatial resolution shortwave infrared imaging technology based
  on time delay and digital accumulation method.
\newblock \emph{Infr Phys Technol}, 81:\penalty0 305--312, 2017.

\bibitem[Friedman et~al.(2001)Friedman, Hastie, and
  Tibshirani]{friedman2001elements}
Jerome Friedman, Trevor Hastie, and Robert Tibshirani.
\newblock \emph{The elements of statistical learning}, volume~1.
\newblock Springer Series in Statistics, 2001.

\bibitem[Murphy and Maggioni(2018)]{murphy2018unsupervised}
James~M Murphy and Mauro Maggioni.
\newblock Unsupervised clustering and active learning of hyperspectral images
  with nonlinear diffusion.
\newblock \emph{IEEE Trans Geosci Remote Sens}, 57\penalty0 (3):\penalty0
  1829--1845, 2018.

\bibitem[Abdolali and Gillis(2021)]{abdolali2021beyond}
Maryam Abdolali and Nicolas Gillis.
\newblock Beyond linear subspace clustering: A comparative study of nonlinear
  manifold clustering algorithms.
\newblock \emph{Comput Sci Rev}, 42:\penalty0 1574--0137, 2021.

\bibitem[Zhuang et~al.(2016)Zhuang, Wang, Lin, Yang, Ma, and
  Yu]{zhuang2016locality}
Liansheng Zhuang, Jingjing Wang, Zhouchen Lin, Allen~Y Yang, Yi~Ma, and Nenghai
  Yu.
\newblock Locality-preserving low-rank representation for graph construction
  from nonlinear manifolds.
\newblock \emph{Neurocomputing}, 175:\penalty0 715--722, 2016.

\bibitem[Kuang et~al.(2012)Kuang, Ding, and Park]{kuang2012symmetric}
Da~Kuang, Chris Ding, and Haesun Park.
\newblock Symmetric nonnegative matrix factorization for graph clustering.
\newblock In \emph{SIAM International Conference Data Min}, pages 106--117.
  SIAM, 2012.

\bibitem[Wang et~al.(2019{\natexlab{a}})Wang, Nie, Wang, He, and
  Li]{wang2019scalable}
R~Wang, N~Nie, Z~Wang, F~He, and X~Li.
\newblock Scalable graph-based clustering with nonnegative relaxation for large
  hyperspectral image.
\newblock \emph{IEEE Trans Geosci Remote Sens}, 57\penalty0 (10):\penalty0
  7352--7364, 2019{\natexlab{a}}.

\bibitem[Camps-Valls et~al.(2007)Camps-Valls, Marsheva, and
  Zhou]{camps2007semi}
G~Camps-Valls, T~V~B Marsheva, and D~Zhou.
\newblock Semi-supervised graph-based hyperspectral image classification.
\newblock \emph{IEEE Trans Geosci Remote Sens}, 45\penalty0 (10):\penalty0
  3044--3054, 2007.

\bibitem[Gao et~al.(2014)Gao, Ji, Cui, Dai, and Hua]{gao2014hyperspectral}
Y~Gao, R~Ji, P~Cui, Q~Dai, and G~Hua.
\newblock Hyperspectral image classification through bilayer graph-based
  learning.
\newblock \emph{IEEE Trans Image Process}, 23\penalty0 (7):\penalty0
  2769--2778, 2014.

\bibitem[Wu and Prasad(2017)]{wu2017semi}
H~Wu and Saurabh Prasad.
\newblock Semi-supervised deep learning using pseudo labels for hyperspectral
  image classification.
\newblock \emph{IEEE Trans Image Process}, 27\penalty0 (3):\penalty0
  1259--1270, 2017.

\bibitem[Yang et~al.(2018)Yang, Ye, Li, Lau, Zhang, and
  Huang]{yang2018hyperspectral}
Xiaofei Yang, Yunming Ye, Xutao Li, Raymond~YK Lau, Xiaofeng Zhang, and Xiaohui
  Huang.
\newblock Hyperspectral image classification with deep learning models.
\newblock \emph{IEEE Trans Geosci Remote Sens}, 56\penalty0 (9):\penalty0
  5408--5423, 2018.

\bibitem[Nalepa et~al.(2020)Nalepa, Myller, Imai, Honda, Takeda, and
  Antoniak]{nalepa2020unsupervised}
Jakub Nalepa, Michal Myller, Yasuteru Imai, Ken-ichi Honda, Tomomi Takeda, and
  Marek Antoniak.
\newblock Unsupervised segmentation of hyperspectral images using 3-d
  convolutional autoencoders.
\newblock \emph{IEEE Geosci Remote Sens Lett}, 17\penalty0 (11):\penalty0
  1948--1952, 2020.

\bibitem[Gillis et~al.(2014)Gillis, Kuang, and Park]{gillis2014hierarchical}
Nicolas Gillis, Da~Kuang, and Haesun Park.
\newblock Hierarchical clustering of hyperspectral images using rank-two
  nonnegative matrix factorization.
\newblock \emph{IEEE Trans Geosci Remote Sens}, 53\penalty0 (4):\penalty0
  2066--2078, 2014.

\bibitem[Li et~al.(2021)Li, Qin, Ling, Wang, Lin, and An]{li2021self}
K~Li, Y~Qin, Q~Ling, Y~Wang, Z~Lin, and W~An.
\newblock Self-supervised deep subspace clustering for hyperspectral images
  with adaptive self-expressive coefficient matrix initialization.
\newblock \emph{IEEE J Sel Top Appl Earth Obs Remote Sens}, 14:\penalty0
  3215--3227, 2021.

\bibitem[Sun et~al.(2020)Sun, Wang, Wei, Fang, Tang, Xu, Yu, and
  Yao]{sun2020deep}
J~Sun, W~Wang, X~Wei, L~Fang, X~Tang, Y~Xu, H~Yu, and W~Yao.
\newblock Deep clustering with intraclass distance constraint for hyperspectral
  images.
\newblock \emph{IEEE Trans Geosci Remote Sens}, 59\penalty0 (5):\penalty0
  4135--4149, 2020.

\bibitem[Zhou et~al.(2016)Zhou, Kwan, Ayhan, and Eismann]{zhou2016novel}
J.~Zhou, C.~Kwan, B.~Ayhan, and M.~T. Eismann.
\newblock A novel cluster kernel \uppercase{RX} algorithm for anomaly and
  change detection using hyperspectral images.
\newblock \emph{IEEE Trans Geosci Remote Sens}, 54\penalty0 (11):\penalty0
  6497--6504, 2016.

\bibitem[Cui and Plemmons(2021)]{cui2021unsupervised}
Kangning Cui and Robert~J Plemmons.
\newblock Unsupervised classification of \uppercase{AVIRIS}-\uppercase{NG}
  hyperspectral images.
\newblock In \emph{Workshop Hyperspectral Image Signal Process Evol Remote
  Sens}, pages 1--5. IEEE, 2021.

\bibitem[Cui et~al.(2022)Cui, Li, Polk, Murphy, Plemmons, and
  Chan]{cui2022unsupervised}
K.~Cui, R.~Li, S.~L. Polk, J.~M. Murphy, R.~J. Plemmons, and R.~H. Chan.
\newblock Unsupervised spatial-spectral hyperspectral image reconstruction and
  clustering with diffusion geometry.
\newblock In \emph{Workshop Hyperspectral Image Signal Process Evol Remote
  Sens}, pages 1--5. IEEE, 2022.

\bibitem[Bachmann et~al.(2005)Bachmann, Ainsworth, and
  Fusina]{bachmann2005exploiting}
Charles~M Bachmann, Thomas~L Ainsworth, and Robert~A Fusina.
\newblock Exploiting manifold geometry in hyperspectral imagery.
\newblock \emph{IEEE Trans Geosci Remote Sens}, 43\penalty0 (3):\penalty0
  441--454, 2005.

\bibitem[Coifman and Lafon(2006)]{coifman2006diffusion}
Ronald~R Coifman and St{\'e}phane Lafon.
\newblock Diffusion maps.
\newblock \emph{Appl and Comput Harm Anal}, 21\penalty0 (1):\penalty0 5--30,
  2006.

\bibitem[Baral et~al.(2014)Baral, Queloz, and Hosoya]{baral2014hymenoscyphus}
H~Baral, V~Queloz, and T~Hosoya.
\newblock Hymenoscyphus fraxineus, the correct scientific name for the fungus
  causing ash dieback in europe.
\newblock \emph{IMA Fungus}, 5\penalty0 (1):\penalty0 79--80, 2014.

\bibitem[McKinney et~al.(2014)McKinney, Nielsen, Collinge, Thomsen, Hansen, and
  Kj{\ae}r]{mckinney2014ash}
LV~McKinney, LR~Nielsen, DB~Collinge, IM~Thomsen, JK~Hansen, and ED~Kj{\ae}r.
\newblock The ash dieback crisis: genetic variation in resistance can prove a
  long-term solution.
\newblock \emph{Plant Pathol}, 63\penalty0 (3):\penalty0 485--499, 2014.

\bibitem[Stone and Mohammed(2017)]{stone2017application}
C.~Stone and C.~Mohammed.
\newblock Application of remote sensing technologies for assessing planted
  forests damaged by insect pests and fungal pathogens: a review.
\newblock \emph{Curr For Rep}, 3\penalty0 (2):\penalty0 75--92, 2017.

\bibitem[Waser et~al.(2014)Waser, K{\"u}chler, J{\"u}tte, and
  Stampfer]{waser2014evaluating}
L.~T. Waser, M.~K{\"u}chler, K.~J{\"u}tte, and T.~Stampfer.
\newblock Evaluating the potential of \uppercase{W}orldview-2 data to classify
  tree species and different levels of ash mortality.
\newblock \emph{Remote Sens}, 6\penalty0 (5):\penalty0 4515--4545, 2014.

\bibitem[Chan et~al.(2021)Chan, Barnes, Swinfield, and
  Coomes]{chan2021monitoring}
A.~H.~Y. Chan, C.~Barnes, T.~Swinfield, and D.~A. Coomes.
\newblock Monitoring ash dieback (\textit{\uppercase{h}ymenoscyphus fraxineus})
  in \uppercase{B}ritish forests using hyperspectral remote sensing.
\newblock \emph{Remote Sens Ecol Conserv}, 7\penalty0 (2):\penalty0 306--320,
  2021.

\bibitem[Ng et~al.(2002)Ng, Jordan, and Weiss]{ng2002spectral}
Andrew~Y Ng, Michael~I Jordan, and Yair Weiss.
\newblock On spectral clustering: Analysis and an algorithm.
\newblock In \emph{Adv Neural Inf Process Syst}, pages 849--856, 2002.

\bibitem[Maggioni and Murphy(2019{\natexlab{a}})]{maggioni2019learning}
Mauro Maggioni and James~M Murphy.
\newblock Learning by unsupervised nonlinear diffusion.
\newblock \emph{J Mach Learn Res}, 20\penalty0 (160):\penalty0 1--56,
  2019{\natexlab{a}}.

\bibitem[Cahill et~al.(2014)Cahill, Czaja, and
  Messinger]{cahill2014schroedinger}
N~D Cahill, W~Czaja, and D~W Messinger.
\newblock Schroedinger eigenmaps with nondiagonal potentials for
  spatial-spectral clustering of hyperspectral imagery.
\newblock In \emph{SPIE}, volume 9088, pages 27--39. SPIE, 2014.

\bibitem[Theodoridis and Koutroumbas(2006)]{theodoridis2006pattern}
S~Theodoridis and K~Koutroumbas.
\newblock \emph{Pattern recognition}.
\newblock Elsevier, 2006.

\bibitem[Zhu et~al.(2017)Zhu, Chayes, Tiard, Sanchez, Dahlberg, Bertozzi,
  Osher, Zosso, and Kuang]{zhu2017unsupervised}
Wei Zhu, Victoria Chayes, Alexandre Tiard, Stephanie Sanchez, Devin Dahlberg,
  Andrea~L Bertozzi, Stanley Osher, Dominique Zosso, and Da~Kuang.
\newblock Unsupervised classification in hyperspectral imagery with nonlocal
  total variation and primal-dual hybrid gradient algorithm.
\newblock \emph{IEEE Trans Geosci Remote Sens}, 55\penalty0 (5):\penalty0
  2786--2798, 2017.

\bibitem[Wang et~al.(2021)Wang, Ma, Nie, and Li]{wang2021fast}
J~Wang, Z~Ma, F~Nie, and X~Li.
\newblock Fast self-supervised clustering with anchor graph.
\newblock \emph{IEEE Trans Neural Netw Learn Syst}, 2021.

\bibitem[Bandyopadhyay and Mukherjee(2022)]{bandyopadhyay2022tree}
D~Bandyopadhyay and S~Mukherjee.
\newblock Tree species classification from hyperspectral data using
  graph-regularized neural networks.
\newblock \emph{arXiv preprint arXiv:2208.08675}, 2022.

\bibitem[Tenenbaum et~al.(2000)Tenenbaum, de~Silva, and
  Langford]{tenenbaum2000global}
J~B Tenenbaum, V~de~Silva, and J~C Langford.
\newblock A global geometric framework for nonlinear dimensionality reduction.
\newblock \emph{Science}, 290\penalty0 (5500):\penalty0 2319--2323, 2000.

\bibitem[Roweis and Saul(2000)]{roweis2000nonlinear}
S~T Roweis and L~K Saul.
\newblock Nonlinear dimensionality reduction by locally linear embedding.
\newblock \emph{Science}, 290\penalty0 (5500):\penalty0 2323--2326, 2000.

\bibitem[Belkin and Niyogi(2001)]{belkin2001laplacian}
M~Belkin and Partha Niyogi.
\newblock Laplacian eigenmaps and spectral techniques for embedding and
  clustering.
\newblock In \emph{Adv Neural Inf Process Syst}, volume~14, 2001.

\bibitem[Rohe et~al.(2011)Rohe, Chatterjee, and Yu]{rohe2011spectral}
K~Rohe, S~Chatterjee, and B~Yu.
\newblock Spectral clustering and the high-dimensional stochastic blockmodel.
\newblock \emph{Ann Stat}, 39\penalty0 (4):\penalty0 1878--1915, 2011.

\bibitem[Murphy and Polk(2022)]{murphy2021multiscale}
James~M. Murphy and Sam~L. Polk.
\newblock A multiscale environment for learning by diffusion.
\newblock \emph{Appl Comput Harm Anal}, 57:\penalty0 58--100, 2022.

\bibitem[Nadler and Galun(2007)]{nadler2007fundamental}
Boaz Nadler and Meirav Galun.
\newblock Fundamental limitations of spectral clustering.
\newblock In \emph{Adv Neural Inf Process Syst}, pages 1017--1024, 2007.

\bibitem[Dilokthanakul et~al.(2016)Dilokthanakul, Mediano, Garnelo, Lee,
  Salimbeni, Arulkumaran, and Shanahan]{dilokthanakul2016deep}
N~Dilokthanakul, P~A~M Mediano, M~Garnelo, M~C~H Lee, H~Salimbeni,
  K~Arulkumaran, and M~Shanahan.
\newblock Deep unsupervised clustering with \uppercase{g}aussian mixture
  variational autoencoders.
\newblock \emph{arXiv preprint arXiv:1611.02648}, 2016.

\bibitem[Min et~al.(2018)Min, Guo, Liu, Zhang, Cui, and Long]{min2018survey}
E~Min, X~Guo, Q~Liu, G~Zhang, J~Cui, and J~Long.
\newblock A survey of clustering with deep learning: From the perspective of
  network architecture.
\newblock \emph{IEEE Access}, 6:\penalty0 39501--39514, 2018.

\bibitem[Tasissa et~al.(2021)Tasissa, Nguyen, and Murphy]{tasissa2021deep}
Abiy Tasissa, Duc Nguyen, and James~M Murphy.
\newblock Deep diffusion processes for active learning of hyperspectral images.
\newblock In \emph{International Geosci Remote Sens Symposium}, pages
  3665--3668. IEEE, 2021.

\bibitem[Nguyen et~al.(2015)Nguyen, Yosinski, and Clune]{nguyen2015deep}
A~Nguyen, J~Yosinski, and J~Clune.
\newblock Deep neural networks are easily fooled: High confidence predictions
  for unrecognizable images.
\newblock In \emph{Comput Vis Pattern Recognit}, pages 427--436, 2015.

\bibitem[Szegedy et~al.(2014)Szegedy, Zaremba, Sutskever, Bruna, Erhan,
  Goodfellow, and Fergus]{szegedy2013intriguing}
C~Szegedy, W~Zaremba, I~Sutskever, J~Bruna, D~Erhan, I~Goodfellow, and
  R~Fergus.
\newblock Intriguing properties of neural networks.
\newblock In \emph{International Conference Learn Represent}, 2014.

\bibitem[Haeffele et~al.(2020)Haeffele, You, and Vidal]{haeffele2020critique}
B~D Haeffele, C~You, and R~Vidal.
\newblock A critique of self-expressive deep subspace clustering.
\newblock In \emph{International Conference Learn Represent}, 2020.

\bibitem[Polk and Murphy(2021)]{polk2021multiscale}
Sam~L. Polk and James~M. Murphy.
\newblock Multiscale clustering of hyperspectral images through
  spectral-spatial diffusion geometry.
\newblock In \emph{International Geosci Remote Sens Symposium}, pages
  4688--4691, 2021.

\bibitem[Haghverdi et~al.(2016)Haghverdi, B{\"u}ttner, Wolf, Buettner, and
  Theis]{haghverdi2016diffusion}
L~Haghverdi, M~B{\"u}ttner, F~A Wolf, F~Buettner, and F~J Theis.
\newblock Diffusion pseudotime robustly reconstructs lineage branching.
\newblock \emph{Nat methods}, 13\penalty0 (10):\penalty0 845--848, 2016.

\bibitem[Van~Dijk et~al.(2018)Van~Dijk, Sharma, Nainys, Yim, Kathail, Carr,
  Burdziak, Moon, Chaffer, and Pattabiraman]{van2018recovering}
D~Van~Dijk, R~Sharma, J~Nainys, K~Yim, P~Kathail, A~J Carr, C~Burdziak, K~R
  Moon, C~L Chaffer, and D~Pattabiraman.
\newblock Recovering gene interactions from single-cell data using data
  diffusion.
\newblock \emph{Cell}, 174\penalty0 (3):\penalty0 716--729, 2018.

\bibitem[Zhao and Singer(2014)]{zhao2014rotationally}
Z~Zhao and A~Singer.
\newblock Rotationally invariant image representation for viewing direction
  classification in cryo-\uppercase{EM}.
\newblock \emph{J Struct Biol}, 186\penalty0 (1):\penalty0 153--166, 2014.

\bibitem[Moon et~al.(2019)Moon, van Dijk, Wang, Gigante, Burkhardt, Chen, Yim,
  van~den Elzen, Hirn, and Coifman]{moon2019visualizing}
K~R Moon, D~van Dijk, Z~Wang, S~Gigante, D~B Burkhardt, W~S Chen, K~Yim,
  A~van~den Elzen, M~J Hirn, and R~R Coifman.
\newblock Visualizing structure and transitions in high-dimensional biological
  data.
\newblock \emph{Nat Biotechnol}, 37\penalty0 (12):\penalty0 1482--1492, 2019.

\bibitem[Rohrdanz et~al.(2011)Rohrdanz, Zheng, Maggioni, and
  Clementi]{rohrdanz2011determination}
M~A Rohrdanz, W~Zheng, M~Maggioni, and C~Clementi.
\newblock Determination of reaction coordinates via locally scaled diffusion
  map.
\newblock \emph{J Chem Phys}, 134\penalty0 (12):\penalty0 03B624, 2011.

\bibitem[Zheng et~al.(2011)Zheng, Rohrdanz, Maggioni, and
  Clementi]{zheng2011polymer}
W~Zheng, M~A Rohrdanz, M~Maggioni, and C~Clementi.
\newblock Polymer reversal rate calculated via locally scaled diffusion map.
\newblock \emph{J Chem Phys}, 134\penalty0 (14):\penalty0 144109, 2011.

\bibitem[Chen and Ferguson(2018)]{chen2018molecular}
W~Chen and A~L Ferguson.
\newblock Molecular enhanced sampling with autoencoders: \uppercase{O}n-the-fly
  collective variable discovery and accelerated free energy landscape
  exploration.
\newblock \emph{J of Comput Chem}, 39\penalty0 (25):\penalty0 2079--2102, 2018.

\bibitem[Coifman et~al.(2005)Coifman, Lafon, Lee, Maggioni, Nadler, Warner, and
  Zucker]{coifman2005geometric}
Ronald~R Coifman, Stephane Lafon, Ann~B Lee, Mauro Maggioni, Boaz Nadler,
  Frederick Warner, and Steven~W Zucker.
\newblock Geometric diffusions as a tool for harmonic analysis and structure
  definition of data: Diffusion maps.
\newblock \emph{Natl Acad Sci USA}, 102\penalty0 (21):\penalty0 7426--7431,
  2005.

\bibitem[Nadler et~al.(2006)Nadler, Lafon, Coifman, and
  Kevrekidis]{nadler2006diffusion}
Boaz Nadler, St{\'e}phane Lafon, Ronald~R Coifman, and Ioannis~G Kevrekidis.
\newblock Diffusion maps, spectral clustering and reaction coordinates of
  dynamical systems.
\newblock \emph{Appl Comput Harmon Anal}, 21\penalty0 (1):\penalty0 113--127,
  2006.

\bibitem[Chan et~al.(2011)Chan, Ma, Ambikapathi, and Chi]{chan2011simplex}
Tsung Chan, Wing Ma, ArulMurugan Ambikapathi, and Chong Chi.
\newblock A simplex volume maximization framework for hyperspectral endmember
  extraction.
\newblock \emph{IEEE Trans Geosci Remote Sens}, 49\penalty0 (11):\penalty0
  4177--4193, 2011.

\bibitem[Winter(1999)]{winter1999nfindr}
Michael~E Winter.
\newblock \uppercase{N}-\uppercase{FINDR}: An algorithm for fast autonomous
  spectral end-member determination in hyperspectral data.
\newblock In \emph{Imaging Spectr V}, volume 3753, pages 266--275. SPIE, 1999.

\bibitem[Manolakis et~al.(2001)Manolakis, Siracusa, and
  Shaw]{manolakis2001hyperspectral}
Dimitris Manolakis, Christina Siracusa, and Gary Shaw.
\newblock Hyperspectral subpixel target detection using the linear mixing
  model.
\newblock \emph{IEEE Trans Geosci Remote Sens}, 39\penalty0 (7):\penalty0
  1392--1409, 2001.

\bibitem[Zhao et~al.(2013)Zhao, Wang, Huang, Ng, and
  Plemmons]{zhao2013deblurring}
Xi-Le Zhao, Fan Wang, Ting-Zhu Huang, Michael~K Ng, and Robert~J Plemmons.
\newblock Deblurring and sparse unmixing for hyperspectral images.
\newblock \emph{IEEE Trans Geosci Remote Sens}, 51\penalty0 (7):\penalty0
  4045--4058, 2013.

\bibitem[Berisha et~al.(2015)Berisha, Nagy, and
  Plemmons]{berisha2015deblurring}
Sebastian Berisha, James~G Nagy, and Robert~J Plemmons.
\newblock Deblurring and sparse unmixing of hyperspectral images using multiple
  point spread functions.
\newblock \emph{SIAM J Sci Comput}, 37\penalty0 (5):\penalty0 S389--S406, 2015.

\bibitem[Wang et~al.(2018)Wang, Feng, Gao, Wang, and He]{wang2018compressed}
Li~Wang, Yan Feng, Yanlong Gao, Zhongliang Wang, and Mingyi He.
\newblock Compressed sensing reconstruction of hyperspectral images based on
  spectral unmixing.
\newblock \emph{IEEE J Sel Top Appl Earth Obs Remote Sens}, 11\penalty0
  (4):\penalty0 1266--1284, 2018.

\bibitem[Cerra et~al.(2013)Cerra, M{\"u}ller, and Reinartz]{cerra2013noise}
Daniele Cerra, Rupert M{\"u}ller, and Peter Reinartz.
\newblock Noise reduction in hyperspectral images through spectral unmixing.
\newblock \emph{IEEE Geosci Remote Sens Lett}, 11\penalty0 (1):\penalty0
  109--113, 2013.

\bibitem[Rasti et~al.(2018)Rasti, Scheunders, Ghamisi, Licciardi, and
  Chanussot]{rasti2018noise}
B~Rasti, P~Scheunders, P~Ghamisi, G~Licciardi, and J~Chanussot.
\newblock Noise reduction in hyperspectral imagery: Overview and application.
\newblock \emph{Remote Sens}, 10\penalty0 (3):\penalty0 482, 2018.

\bibitem[Rasti et~al.(2020)Rasti, Koirala, Scheunders, and
  Ghamisi]{rasti2020hyperspectral}
Behnood Rasti, Bikram Koirala, Paul Scheunders, and Pedram Ghamisi.
\newblock How hyperspectral image unmixing and denoising can boost each other.
\newblock \emph{Remote Sens}, 12\penalty0 (11):\penalty0 1728, 2020.

\bibitem[Ert{\"u}rk et~al.(2014)Ert{\"u}rk, G{\"u}ll{\"u}, {\c{C}}e{\c{s}}meci,
  Ger{\c{c}}ek, and Ert{\"u}rk]{erturk2014spatial}
Alp Ert{\"u}rk, Mehmet~Kemal G{\"u}ll{\"u}, Davut {\c{C}}e{\c{s}}meci, Deniz
  Ger{\c{c}}ek, and Sarp Ert{\"u}rk.
\newblock Spatial resolution enhancement of hyperspectral images using unmixing
  and binary particle swarm optimization.
\newblock \emph{IEEE Geosci Remote Sens Lett}, 11\penalty0 (12):\penalty0
  2100--2104, 2014.

\bibitem[Bendoumi et~al.(2014)Bendoumi, He, and Mei]{bendoumi2014hyperspectral}
Mohamed~Amine Bendoumi, Mingyi He, and Shaohui Mei.
\newblock Hyperspectral image resolution enhancement using high-resolution
  multispectral image based on spectral unmixing.
\newblock \emph{IEEE Trans Geosci Remote Sens}, 52\penalty0 (10):\penalty0
  6574--6583, 2014.

\bibitem[Kordi~Ghasrodashti et~al.(2017)Kordi~Ghasrodashti, Karami, Heylen, and
  Scheunders]{kordi2017spatial}
E~Kordi~Ghasrodashti, A~Karami, R~Heylen, and P~Scheunders.
\newblock Spatial resolution enhancement of hyperspectral images using spectral
  unmixing and \uppercase{B}ayesian sparse representation.
\newblock \emph{Remote Sens}, 9\penalty0 (6):\penalty0 541, 2017.

\bibitem[Villa et~al.(2010)Villa, Chanussot, Benediktsson, and
  Jutten]{villa2010spectral}
Alberto Villa, Jocelyn Chanussot, Jon~Atli Benediktsson, and Christian Jutten.
\newblock Spectral unmixing for the classification of hyperspectral images at a
  finer spatial resolution.
\newblock \emph{IEEE J Sel Top Appl Earth Obs Remote Sens}, 5\penalty0
  (3):\penalty0 521--533, 2010.

\bibitem[D{\'o}pido et~al.(2012)D{\'o}pido, Villa, Plaza, and
  Gamba]{dopido2012quantitative}
Inmaculada D{\'o}pido, Alberto Villa, Antonio Plaza, and Paolo Gamba.
\newblock A quantitative and comparative assessment of unmixing-based feature
  extraction techniques for hyperspectral image classification.
\newblock \emph{IEEE J Sel Top Appl Earth Obs Remote Sens}, 5\penalty0
  (2):\penalty0 421--435, 2012.

\bibitem[Ert{\"u}rk and Plaza(2015)]{erturk2015informative}
Alp Ert{\"u}rk and Antonio Plaza.
\newblock Informative change detection by unmixing for hyperspectral images.
\newblock \emph{IEEE Geosci Remote Sens Lett}, 12\penalty0 (6):\penalty0
  1252--1256, 2015.

\bibitem[Liu et~al.(2016)Liu, Bruzzone, Bovolo, and Du]{liu2016unsupervised}
Sicong Liu, Lorenzo Bruzzone, Francesca Bovolo, and Peijun Du.
\newblock Unsupervised multitemporal spectral unmixing for detecting multiple
  changes in hyperspectral images.
\newblock \emph{IEEE Trans Geosci Remote Sens}, 54\penalty0 (5):\penalty0
  2733--2748, 2016.

\bibitem[Camalan et~al.(2022)Camalan, Cui, Pauca, Alqahtani, Silman, Chan,
  Plemmons, Dethier, Fernandez, and Lutz]{camalan2022change}
S.~Camalan, K.~Cui, V.~P. Pauca, S.~Alqahtani, M.~Silman, R.~Chan, R.~J.
  Plemmons, E.~N. Dethier, L.~E. Fernandez, and D.~A. Lutz.
\newblock Change detection of \uppercase{a}mazonian alluvial gold mining using
  deep learning and \uppercase{S}entinel-2 imagery.
\newblock \emph{Remote Sens}, 14\penalty0 (7):\penalty0 1746, 2022.

\bibitem[Li et~al.(2022)Li, Wu, and Xu]{li2022integrated}
Haishan Li, Ke~Wu, and Ying Xu.
\newblock An integrated change detection method based on spectral unmixing and
  the \uppercase{CNN} for hyperspectral imagery.
\newblock \emph{Remote Sens}, 14\penalty0 (11):\penalty0 2523, 2022.

\bibitem[Qu et~al.(2018)Qu, Wang, Guo, Ayhan, Kwan, Vance, and
  Qi]{qu2018hyperspectral}
Ying Qu, Wei Wang, Rui Guo, Bulent Ayhan, Chiman Kwan, Steven Vance, and
  Hairong Qi.
\newblock Hyperspectral anomaly detection through spectral unmixing and
  dictionary-based low-rank decomposition.
\newblock \emph{IEEE Trans Geosci Remote Sens}, 56\penalty0 (8):\penalty0
  4391--4405, 2018.

\bibitem[Ma et~al.(2018)Ma, Yuan, and Wang]{ma2018hyperspectral}
Dandan Ma, Yuan Yuan, and Qi~Wang.
\newblock Hyperspectral anomaly detection via discriminative feature learning
  with multiple-dictionary sparse representation.
\newblock \emph{Remote Sens}, 10\penalty0 (5):\penalty0 745, 2018.

\bibitem[Somers et~al.(2011)Somers, Asner, Tits, and
  Coppin]{somers2011endmember}
Ben Somers, Gregory~P Asner, Laurent Tits, and Pol Coppin.
\newblock Endmember variability in spectral mixture analysis: A review.
\newblock \emph{Remote Sens Environ}, 115\penalty0 (7):\penalty0 1603--1616,
  2011.

\bibitem[Quintano et~al.(2012)Quintano, Fern{\'a}ndez-Manso, Shimabukuro, and
  Pereira]{quintano2012spectral}
C~Quintano, A~Fern{\'a}ndez-Manso, Y~E Shimabukuro, and G~Pereira.
\newblock Spectral unmixing.
\newblock \emph{Int J Remote Sens}, 33\penalty0 (17):\penalty0 5307--5340,
  2012.

\bibitem[Bioucas-Dias et~al.(2012)Bioucas-Dias, Plaza, Dobigeon, Parente, Du,
  Gader, and Chanussot]{bioucas2012hyperspectral}
Jos{\'e}~M Bioucas-Dias, Antonio Plaza, Nicolas Dobigeon, Mario Parente, Qian
  Du, Paul Gader, and Jocelyn Chanussot.
\newblock Hyperspectral unmixing overview: Geometrical, statistical, and sparse
  regression-based approaches.
\newblock \emph{IEEE J Sel Top Appl Earth Obs Remote Sens}, 5\penalty0
  (2):\penalty0 354--379, 2012.

\bibitem[Heylen et~al.(2014)Heylen, Parente, and Gader]{heylen2014review}
Rob Heylen, Mario Parente, and Paul Gader.
\newblock A review of nonlinear hyperspectral unmixing methods.
\newblock \emph{IEEE J Sel Top Appl Earth Obs Remote Sens}, 7\penalty0
  (6):\penalty0 1844--1868, 2014.

\bibitem[Borsoi et~al.(2021)Borsoi, Imbiriba, Bermudez, Richard, Chanussot,
  Drumetz, Tourneret, Zare, and Jutten]{borsoi2021spectral}
Ricardo Borsoi, Tales Imbiriba, Jose~Carlos Bermudez, Cedric Richard, Jocelyn
  Chanussot, Lucas Drumetz, Jean-Yves Tourneret, Alina Zare, and Christian
  Jutten.
\newblock Spectral variability in hyperspectral data unmixing: A comprehensive
  review.
\newblock \emph{IEEE Geosci Remote Sens Mag}, 2021.

\bibitem[Chang et~al.(2006)Chang, Wu, Liu, and Ouyang]{chang2006new}
C~Chang, C~Wu, W~Liu, and Y~Ouyang.
\newblock A new growing method for simplex-based endmember extraction
  algorithm.
\newblock \emph{IEEE Trans Geosci Remote Sens}, 44\penalty0 (10):\penalty0
  2804--2819, 2006.

\bibitem[Neville(1999)]{neville1999automatic}
R~Neville.
\newblock Automatic endmember extraction from hyperspectral data for mineral
  exploration.
\newblock In \emph{Canadian Symposium Remote Sens}, 1999.

\bibitem[Boardman et~al.(1995)Boardman, Kruse, and Green]{boardman1995mapping}
J~W Boardman, F~A Kruse, and R~O Green.
\newblock Mapping target signatures via partial unmixing of \uppercase{AVIRIS}
  data.
\newblock Technical report, Jet Propulsion Laboratory, 1995.

\bibitem[Boardman(1993)]{boardman1993automating}
Joseph~W Boardman.
\newblock Automating spectral unmixing of \uppercase{AVIRIS} data using convex
  geometry concepts.
\newblock In \emph{Annu JPL Airborne Geosci Workshop}, volume~1, pages 11--14,
  1993.

\bibitem[Chan et~al.(2009)Chan, Chi, Huang, and Ma]{chan2009convex}
T~Chan, C~Chi, Y~Huang, and W~Ma.
\newblock A convex analysis-based minimum-volume enclosing simplex algorithm
  for hyperspectral unmixing.
\newblock \emph{IEEE Trans Signal Process}, 57\penalty0 (11):\penalty0
  4418--4432, 2009.

\bibitem[Nascimento and Dias(2005)]{nascimento2005VCA}
Jos{\'e}~MP Nascimento and Jos{\'e}~MB Dias.
\newblock Vertex component analysis: A fast algorithm to unmix hyperspectral
  data.
\newblock \emph{IEEE Trans Geosci Remote Sens}, 43\penalty0 (4):\penalty0
  898--910, 2005.

\bibitem[Clasen et~al.(2015)Clasen, Somers, Pipkins, Tits, Segl, Brell,
  Kleinschmit, Spengler, Lausch, and F{\"o}rster]{clasen2015spectral}
Anne Clasen, Ben Somers, Kyle Pipkins, Laurent Tits, Karl Segl, Max Brell,
  Birgit Kleinschmit, Daniel Spengler, Angela Lausch, and Michael F{\"o}rster.
\newblock Spectral unmixing of forest crown components at close range, airborne
  and simulated \uppercase{S}entinel-2 and \uppercase{E}n\uppercase{MAP}
  spectral imaging scale.
\newblock \emph{Remote Sens}, 7\penalty0 (11):\penalty0 15361--15387, 2015.

\bibitem[Heylen et~al.(2011)Heylen, Burazerovic, and
  Scheunders]{heylen2011fully}
R~Heylen, D~Burazerovic, and P~Scheunders.
\newblock Fully constrained least squares spectral unmixing by simplex
  projection.
\newblock \emph{IEEE Trans Geosci Remote Sens}, 49\penalty0 (11):\penalty0
  4112--4122, 2011.

\bibitem[Hendrix et~al.(2011)Hendrix, Garcia, Plaza, Martin, and
  Plaza]{hendrix2011new}
Eligius~MT Hendrix, Inmaculada Garcia, Javier Plaza, Gabriel Martin, and
  Antonio Plaza.
\newblock A new minimum-volume enclosing algorithm for endmember identification
  and abundance estimation in hyperspectral data.
\newblock \emph{IEEE Trans Geosci Remote Sens}, 50\penalty0 (7):\penalty0
  2744--2757, 2011.

\bibitem[Iordache et~al.(2011)Iordache, Bioucas-Dias, and
  Plaza]{iordache2011sparse}
Marian-Daniel Iordache, Jos{\'e}~M Bioucas-Dias, and Antonio Plaza.
\newblock Sparse unmixing of hyperspectral data.
\newblock \emph{IEEE Trans Geosci Remote Sens}, 49\penalty0 (6):\penalty0
  2014--2039, 2011.

\bibitem[Berman et~al.(2004)Berman, Kiiveri, Lagerstrom, Ernst, Dunne, and
  Huntington]{berman2004ice}
M~Berman, H~Kiiveri, R~Lagerstrom, A~Ernst, R~Dunne, and J~F Huntington.
\newblock \uppercase{ICE}: A statistical approach to identifying endmembers in
  hyperspectral images.
\newblock \emph{IEEE Trans Signal Process}, 42\penalty0 (10):\penalty0
  2085--2095, 2004.

\bibitem[Zare and Gader(2007)]{zare2007sparsity}
A~Zare and P~Gader.
\newblock Sparsity promoting iterated constrained endmember detection in
  hyperspectral imagery.
\newblock \emph{IEEE Geosci Remote Sens Lett}, 4\penalty0 (3):\penalty0
  446--450, 2007.

\bibitem[Dobigeon et~al.(2009)Dobigeon, Moussaoui, Coulon, Tourneret, and
  Hero]{dobigeon2009joint}
N~Dobigeon, S~Moussaoui, M~Coulon, J~Tourneret, and A~O Hero.
\newblock Joint \uppercase{B}ayesian endmember extraction and linear unmixing
  for hyperspectral imagery.
\newblock \emph{IEEE Trans Signal Process}, 57\penalty0 (11):\penalty0
  4355--4368, 2009.

\bibitem[Moussaoui et~al.(2006)Moussaoui, Brie, Mohammad-Djafari, and
  Carteret]{moussaoui2006separation}
S~Moussaoui, D~Brie, A~Mohammad-Djafari, and C~Carteret.
\newblock Separation of non-negative mixture of non-negative sources using a
  bayesian approach and \uppercase{MCMC} sampling.
\newblock \emph{IEEE Trans Signal Process}, 54\penalty0 (11):\penalty0
  4133--4145, 2006.

\bibitem[Themelis et~al.(2011)Themelis, Rontogiannis, and
  Koutroumbas]{themelis2011novel}
Konstantinos~E Themelis, Athanasios~A Rontogiannis, and Konstantinos~D
  Koutroumbas.
\newblock A novel hierarchical \uppercase{B}ayesian approach for sparse
  semisupervised hyperspectral unmixing.
\newblock \emph{IEEE Trans Signal Process}, 60\penalty0 (2):\penalty0 585--599,
  2011.

\bibitem[Palsson et~al.(2020)Palsson, Ulfarsson, and
  Sveinsson]{palsson2020convolutional}
B~Palsson, M~O Ulfarsson, and J~R Sveinsson.
\newblock Convolutional autoencoder for spectral--spatial hyperspectral
  unmixing.
\newblock \emph{IEEE Trans Geosci Remote Sens}, 59\penalty0 (1):\penalty0
  535--549, 2020.

\bibitem[Su et~al.(2019)Su, Li, Plaza, Marinoni, Gamba, and
  Chakravortty]{su2019daen}
Y~Su, J~Li, A~Plaza, A~Marinoni, P~Gamba, and S~Chakravortty.
\newblock \uppercase{DAEN}: \uppercase{D}eep autoencoder networks for
  hyperspectral unmixing.
\newblock \emph{IEEE Trans Geosci Remote Sens}, 57\penalty0 (7):\penalty0
  4309--4321, 2019.

\bibitem[Palsson et~al.(2018)Palsson, Sigurdsson, Sveinsson, and
  Ulfarsson]{palsson2018hyperspectral}
B~Palsson, J~Sigurdsson, J~R Sveinsson, and M~O Ulfarsson.
\newblock Hyperspectral unmixing using a neural network autoencoder.
\newblock \emph{IEEE Access}, 6:\penalty0 25646--25656, 2018.

\bibitem[Qu and Qi(2018)]{qu2018udas}
Y~Qu and H~Qi.
\newblock u\uppercase{DAS}: An untied denoising autoencoder with sparsity for
  spectral unmixing.
\newblock \emph{IEEE Trans Geosci Remote Sens}, 57\penalty0 (3):\penalty0
  1698--1712, 2018.

\bibitem[Ozkan et~al.(2018)Ozkan, Kaya, and Akar]{ozkan2018endnet}
Savas Ozkan, Berk Kaya, and Gozde~Bozdagi Akar.
\newblock Endnet: \uppercase{S}parse autoencoder network for endmember
  extraction and hyperspectral unmixing.
\newblock \emph{IEEE Trans Geosci Remote Sens}, 57\penalty0 (1):\penalty0
  482--496, 2018.

\bibitem[Zhang et~al.(2018)Zhang, Sun, Zhang, Wu, and
  Jiao]{zhang2018hyperspectral}
X~Zhang, Y~Sun, J~Zhang, P~Wu, and L~Jiao.
\newblock Hyperspectral unmixing via deep convolutional neural networks.
\newblock \emph{IEEE Geosci Remote Sens Lett}, 15\penalty0 (11):\penalty0
  1755--1759, 2018.

\bibitem[Su et~al.(2018)Su, Marinoni, Li, Plaza, and Gamba]{su2018stacked}
Y~Su, A~Marinoni, J~Li, J~Plaza, and P~Gamba.
\newblock Stacked nonnegative sparse autoencoders for robust hyperspectral
  unmixing.
\newblock \emph{IEEE Geosci Remote Sens Lett}, 15\penalty0 (9):\penalty0
  1427--1431, 2018.

\bibitem[Khajehrayeni and Ghassemian(2020)]{khajehrayeni2020hyperspectral}
F~Khajehrayeni and H~Ghassemian.
\newblock Hyperspectral unmixing using deep convolutional autoencoders in a
  supervised scenario.
\newblock \emph{IEEE J Sel Top Appl Earth Obs Remote Sens}, 13:\penalty0
  567--576, 2020.

\bibitem[Feng et~al.(2018)Feng, Li, Li, Du, Plaza, and
  Emery]{feng2018hyperspectral}
X~Feng, H~Li, J~Li, Q~Du, A~Plaza, and W~J Emery.
\newblock Hyperspectral unmixing using sparsity-constrained deep nonnegative
  matrix factorization with total variation.
\newblock \emph{IEEE Trans Geosci Remote Sens}, 56\penalty0 (10):\penalty0
  6245--6257, 2018.

\bibitem[Guilfoyle et~al.(2001)Guilfoyle, Althouse, and
  Chang]{guilfoyle2001quantitative}
K~J Guilfoyle, M~L Althouse, and C~Chang.
\newblock A quantitative and comparative analysis of linear and nonlinear
  spectral mixture models using radial basis function neural networks.
\newblock \emph{IEEE Trans Geosci Remote Sens}, 39\penalty0 (10):\penalty0
  2314--2318, 2001.

\bibitem[Licciardi and Del~Frate(2011)]{licciardi2011pixel}
G~A Licciardi and F~Del~Frate.
\newblock Pixel unmixing in hyperspectral data by means of neural networks.
\newblock \emph{IEEE Trans Geosci Remote Sens}, 49\penalty0 (11):\penalty0
  4163--4172, 2011.

\bibitem[Charles et~al.(2011)Charles, Olshausen, and
  Rozell]{charles2011learning}
A~S Charles, B~A Olshausen, and C~J Rozell.
\newblock Learning sparse codes for hyperspectral imagery.
\newblock \emph{IEEE J Sel Top Appl Earth Obs Remote Sens}, 5\penalty0
  (5):\penalty0 963--978, 2011.

\bibitem[Wang et~al.(2019{\natexlab{b}})Wang, Zhao, Chen, and
  Rahardja]{wang2019nonlinear}
M~Wang, M~Zhao, J~Chen, and S~Rahardja.
\newblock Nonlinear unmixing of hyperspectral data via deep autoencoder
  networks.
\newblock \emph{IEEE Geosci Remote Sens Lett}, 16\penalty0 (9):\penalty0
  1467--1471, 2019{\natexlab{b}}.

\bibitem[Yokoya et~al.(2013)Yokoya, Chanussot, and
  Iwasaki]{yokoya2013nonlinear}
Naoto Yokoya, Jocelyn Chanussot, and Akira Iwasaki.
\newblock Nonlinear unmixing of hyperspectral data using semi-nonnegative
  matrix factorization.
\newblock \emph{IEEE Trans Geosci Remote Sens}, 52\penalty0 (2):\penalty0
  1430--1437, 2013.

\bibitem[Halimi et~al.(2011)Halimi, Altmann, Dobigeon, and
  Tourneret]{halimi2011nonlinear}
Abderrahim Halimi, Yoann Altmann, Nicolas Dobigeon, and Jean-Yves Tourneret.
\newblock Nonlinear unmixing of hyperspectral images using a generalized
  bilinear model.
\newblock \emph{IEEE Trans Geosci Remote Sens}, 49\penalty0 (11):\penalty0
  4153--4162, 2011.

\bibitem[Chen et~al.(2012)Chen, Richard, and Honeine]{chen2012nonlinear}
J~Chen, C~Richard, and Paul Honeine.
\newblock Nonlinear unmixing of hyperspectral data based on a
  linear-mixture/nonlinear-fluctuation model.
\newblock \emph{IEEE Trans Signal Process}, 61\penalty0 (2):\penalty0 480--492,
  2012.

\bibitem[Heylen and Scheunders(2015)]{heylen2015multilinear}
R~Heylen and P~Scheunders.
\newblock A multilinear mixing model for nonlinear spectral unmixing.
\newblock \emph{IEEE Trans Geosci Remote Sens}, 54\penalty0 (1):\penalty0
  240--251, 2015.

\bibitem[Heylen et~al.(2010)Heylen, Burazerovic, and Scheunders]{heylen2010non}
R~Heylen, D~Burazerovic, and P~Scheunders.
\newblock Non-linear spectral unmixing by geodesic simplex volume maximization.
\newblock \emph{IEEE J Sel Top Appl Earth Obs Remote Sens}, 5\penalty0
  (3):\penalty0 534--542, 2010.

\bibitem[Bioucas-Dias and Nascimento(2008)]{bioucas2008HySime}
Jos{\'e}~M Bioucas-Dias and Jos{\'e}~MP Nascimento.
\newblock Hyperspectral subspace identification.
\newblock \emph{IEEE Trans Geosci Remote Sens}, 46\penalty0 (8):\penalty0
  2435--2445, 2008.

\bibitem[Chang(2018)]{chang2018review}
Chein Chang.
\newblock A review of virtual dimensionality for hyperspectral imagery.
\newblock \emph{IEEE J Sel Top Appl Earth Obs Remote Sens}, 11\penalty0
  (4):\penalty0 1285--1305, 2018.

\bibitem[Chang and Du(2004)]{chang2004estimation}
Chein Chang and Qian Du.
\newblock Estimation of number of spectrally distinct signal sources in
  hyperspectral imagery.
\newblock \emph{IEEE Trans Geosci Remote Sens}, 42\penalty0 (3):\penalty0
  608--619, 2004.

\bibitem[Chan et~al.(2008)Chan, Ma, Chi, and Wang]{chan2008convex}
T-H Chan, W-K Ma, C-Y Chi, and Y~Wang.
\newblock A convex analysis framework for blind separation of non-negative
  sources.
\newblock \emph{IEEE Trans Signal Process}, 56\penalty0 (10):\penalty0
  5120--5134, 2008.

\bibitem[Bro and De~Jong(1997)]{bro1997fast}
Rasmus Bro and Sijmen De~Jong.
\newblock A fast non-negativity-constrained least squares algorithm.
\newblock \emph{J Chemom}, 11\penalty0 (5):\penalty0 393--401, 1997.

\bibitem[Chen et~al.(2013)Chen, Richard, and Honeine]{chen2013nonlinear}
J~Chen, C~Richard, and P~Honeine.
\newblock Nonlinear estimation of material abundances in hyperspectral images
  with $\ell^1$-norm spatial regularization.
\newblock \emph{IEEE Trans Geosci Remote Sens}, 52\penalty0 (5):\penalty0
  2654--2665, 2013.

\bibitem[Heinz and Chang(2001)]{heinz2001fully}
D~C Heinz and C-I Chang.
\newblock Fully constrained least squares linear spectral mixture analysis
  method for material quantification in hyperspectral imagery.
\newblock \emph{IEEE Trans Geosci Remote Sens}, 39\penalty0 (3):\penalty0
  529--545, 2001.

\bibitem[Rodriguez and Laio(2014)]{rodriguez2014clustering}
Alex Rodriguez and Alessandro Laio.
\newblock Clustering by fast search and find of density peaks.
\newblock \emph{Science}, 344\penalty0 (6191):\penalty0 1492--1496, 2014.

\bibitem[Beygelzimer et~al.(2006)Beygelzimer, Kakade, and
  Langford]{beygelzimer2006cover}
Alina Beygelzimer, Sham Kakade, and John Langford.
\newblock Cover trees for nearest neighbor.
\newblock In \emph{International Conference Mach Learn}, pages 97--104, 2006.

\bibitem[Shi and Malik(2000)]{shi2000normalized}
Jianbo Shi and Jitendra Malik.
\newblock Normalized cuts and image segmentation.
\newblock \emph{IEEE Trans Pattern Anal Mach Intell}, 22\penalty0 (8):\penalty0
  888--905, 2000.

\bibitem[Polk et~al.(2022)Polk, Chan, Cui, Plemmons, Coomes, and
  Murphy]{polk2022unsupervised}
S.~L. Polk, A.~H.~Y. Chan, K.~Cui, R.~J. Plemmons, D.~A. Coomes, and J.~M.
  Murphy.
\newblock Unsupervised detection of ash dieback disease (\textit{Hymenoscyphus
  fraxineus}) using diffusion-based hyperspectral image clustering.
\newblock In \emph{International Geosci Remote Sens Symposium}, pages
  2287--2290, 2022.

\bibitem[Cohen(1960)]{cohen1960kappa}
Jacob Cohen.
\newblock A coefficient of agreement for nominal scales.
\newblock \emph{Educ Psychol Meas}, 20\penalty0 (1):\penalty0 37--46, 1960.

\bibitem[Swinfield et~al.(2020)Swinfield, Both, Riutta, Bongalov, Elias,
  Majalap-Lee, Ostle, Sv{\'a}tek, Kvasnica, and
  Milodowski]{swinfield2020imaging}
Tom Swinfield, Sabine Both, Terhi Riutta, Boris Bongalov, Dafydd Elias, Noreen
  Majalap-Lee, Nicholas Ostle, Martin Sv{\'a}tek, Jakub Kvasnica, and David
  Milodowski.
\newblock Imaging spectroscopy reveals the effects of topography and logging on
  the leaf chemistry of tropical forest canopy trees.
\newblock \emph{Glob Chang Biol}, 26\penalty0 (2):\penalty0 989--1002, 2020.

\bibitem[Kotzagiannidis and Sch{\"o}nlieb(2021)]{kotzagiannidis2021semi}
M~S Kotzagiannidis and C-B Sch{\"o}nlieb.
\newblock Semi-supervised superpixel-based multi-feature graph learning for
  hyperspectral image data.
\newblock \emph{IEEE Trans Geosci Remote Sens}, 60:\penalty0 1--12, 2021.

\bibitem[Qin et~al.(2018)Qin, Shang, Tian, Wang, Zhang, and
  Tang]{qin2018spectral}
A~Qin, Z~Shang, J~Tian, Y~Wang, T~Zhang, and Y~Y Tang.
\newblock Spectral--spatial graph convolutional networks for semisupervised
  hyperspectral image classification.
\newblock \emph{IEEE Geosci Remote Sens Lett}, 16\penalty0 (2):\penalty0
  241--245, 2018.

\bibitem[Hong et~al.(2020)Hong, Gao, Yao, Zhang, Plaza, and
  Chanussot]{hong2020graph}
D~Hong, L~Gao, J~Yao, B~Zhang, A~Plaza, and J~Chanussot.
\newblock Graph convolutional networks for hyperspectral image classification.
\newblock \emph{IEEE Trans Geosci Remote Sens}, 59\penalty0 (7):\penalty0
  5966--5978, 2020.

\bibitem[Sun et~al.(2021)Sun, Zheng, and Lu]{sun2021supervised}
H~Sun, X~Zheng, and X~Lu.
\newblock A supervised segmentation network for hyperspectral image
  classification.
\newblock \emph{IEEE Trans Image Process}, 30:\penalty0 2810--2825, 2021.

\bibitem[Kavalerov et~al.(2020)Kavalerov, Li, Czaja, and
  Chellappa]{kavalerov20203}
I~Kavalerov, Weilin Li, W~Czaja, and R~Chellappa.
\newblock 3-\uppercase{D} \uppercase{F}ourier scattering transform and
  classification of hyperspectral images.
\newblock \emph{IEEE Trans Geosci Remote Sens}, 59\penalty0 (12):\penalty0
  10312--10327, 2020.

\bibitem[Murphy and Maggioni(2019)]{murphy2019spectral}
James~M Murphy and Mauro Maggioni.
\newblock Spectral--spatial diffusion geometry for hyperspectral image
  clustering.
\newblock \emph{IEEE Geosci Remote Sens Lett}, 17\penalty0 (7):\penalty0
  1243--1247, 2019.

\bibitem[Murphy(2020{\natexlab{a}})]{murphy2020spatially}
James~M Murphy.
\newblock Spatially regularized active diffusion learning for high-dimensional
  images.
\newblock \emph{Pattern Recognit Lett}, 135:\penalty0 213--220,
  2020{\natexlab{a}}.

\bibitem[Keys(1981)]{keys1981bicubic}
R.~Keys.
\newblock Cubic convolution interpolation for digital image processing.
\newblock \emph{IEEE Trans Signal Process}, 29\penalty0 (6):\penalty0
  1153--1160, 1981.

\bibitem[Murphy(2020{\natexlab{b}})]{murphy2020patch}
James~M Murphy.
\newblock Patch-based diffusion learning for hyperspectral image clustering.
\newblock In \emph{International Geosci Remote Sens Symposium}, pages
  1042--1045. IEEE, 2020{\natexlab{b}}.

\bibitem[Fauvel et~al.(2012)Fauvel, Tarabalka, Benediktsson, Chanussot, and
  Tilton]{fauvel2012advances}
Mathieu Fauvel, Yuliya Tarabalka, Jon~Atli Benediktsson, Jocelyn Chanussot, and
  James~C Tilton.
\newblock Advances in spectral-spatial classification of hyperspectral images.
\newblock \emph{Proceedings IEEE}, 101\penalty0 (3):\penalty0 652--675, 2012.

\bibitem[Ghamisi et~al.(2013)Ghamisi, Benediktsson, and
  Ulfarsson]{ghamisi2013spectral}
Pedram Ghamisi, Jon~Atli Benediktsson, and Magnus~Orn Ulfarsson.
\newblock Spectral--spatial classification of hyperspectral images based on
  hidden \uppercase{M}arkov random fields.
\newblock \emph{IEEE Trans Geosci Remote Sens}, 52\penalty0 (5):\penalty0
  2565--2574, 2013.

\bibitem[Fang et~al.(2015)Fang, Li, Duan, Ren, and
  Benediktsson]{fang2015classification}
Leyuan Fang, Shutao Li, Wuhui Duan, Jinchang Ren, and J{\'o}n~Atli
  Benediktsson.
\newblock Classification of hyperspectral images by exploiting
  spectral--spatial information of superpixel via multiple kernels.
\newblock \emph{IEEE Trans Geosci Remote Sens}, 53\penalty0 (12):\penalty0
  6663--6674, 2015.

\bibitem[Tarabalka et~al.(2009)Tarabalka, Benediktsson, and
  Chanussot]{tarabalka2009spectral}
Yuliya Tarabalka, J{\'o}n~Atli Benediktsson, and Jocelyn Chanussot.
\newblock Spectral--spatial classification of hyperspectral imagery based on
  partitional clustering techniques.
\newblock \emph{IEEE Trans Geosci Remote Sens}, 47\penalty0 (8):\penalty0
  2973--2987, 2009.

\bibitem[Mohan et~al.(2007)Mohan, Sapiro, and Bosch]{mohan2007spatially}
A~Mohan, G~Sapiro, and E~Bosch.
\newblock Spatially coherent nonlinear dimensionality reduction and
  segmentation of hyperspectral images.
\newblock \emph{IEEE Geosci Remote Sens Lett}, 4\penalty0 (2):\penalty0
  206--210, 2007.

\bibitem[Polk(2022)]{polk2022diffusion}
Sam~L Polk.
\newblock \emph{Diffusion-Based Clustering of High-Dimensional Datasets}.
\newblock PhD thesis, Tufts University, 2022.

\bibitem[Maggioni and Murphy(2019{\natexlab{b}})]{maggioni2019LAND}
Mauro Maggioni and James~M Murphy.
\newblock Learning by active nonlinear diffusion.
\newblock \emph{Found Data Sci}, 1\penalty0 (3):\penalty0 271,
  2019{\natexlab{b}}.

\bibitem[Gerg and Kun(2022)]{matlab_hsi_toolbox}
Isaac Gerg and David Kun.
\newblock Hyperspectral toolbox.
\newblock \url{https://github.com/davidkun/HyperSpectralToolbox}, 2022.

\end{thebibliography}
\bibliographystyle{unsrtnat}

\end{document}